%% file: main.tex
\documentclass[11pt,letterpaper]{mystyle}
\usepackage{fancyhdr}

\usepackage{tikz}
\usepackage[utf8]{inputenc} 
\usepackage[T1]{fontenc}
\usepackage[numbers]{natbib}
\usepackage{adjustbox}
\usepackage{breakurl}    
\usepackage{hyperref}
\usepackage[edges]{forest}
\usepackage{subcaption}
\usepackage{soul}
\usepackage{multirow}
\usepackage[utf8]{inputenc} 
\usepackage{booktabs}       
\usepackage{amsfonts}       
\usepackage{nicefrac}      
\usepackage{microtype}     
\usepackage{xcolor}        
\usepackage{colortbl}
\usepackage{enumitem}
\usepackage[numbers]{natbib}
\usepackage{amssymb}
\usepackage{wrapfig}
\usepackage{courier}
\usepackage{bxcoloremoji}
\usepackage{CJKutf8}
\usepackage{ragged2e}
\usepackage{longtable}
\usepackage{threeparttable}

\newcommand{\github}{\raisebox{-1.5pt}{\includegraphics[height=1.05em]{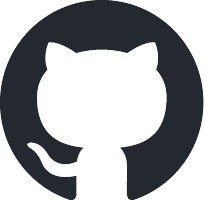}}}

\fancypagestyle{headstyle}{
    \fancyhead[L]{
        \includegraphics[width=60pt]{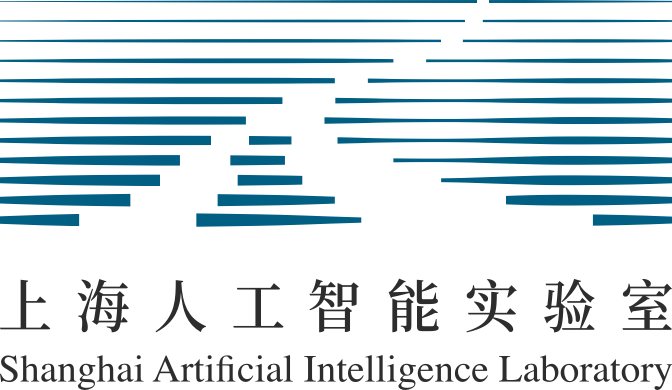}
    }
    \fancyhead[C]{
        From \emph{AI for Science} to \emph{Agentic Science}
    }
    \fancyhead[R]{
        \includegraphics[width=120pt]{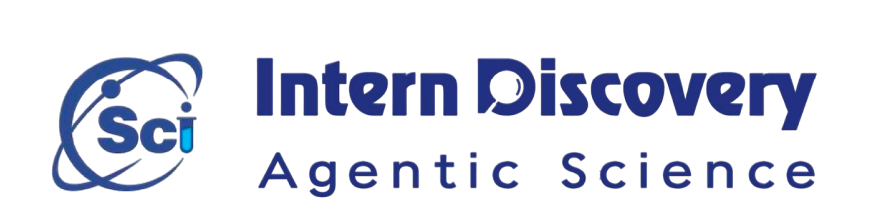}
    }

}

\definecolor{hidden-red}{RGB}{205, 44, 36}
\definecolor{hidden-blue}{RGB}{194,232,247}
\definecolor{hidden-orange}{RGB}{243,202,120}
\definecolor{hidden-green}{RGB}{34,139,34}
\definecolor{hidden-pink}{RGB}{255,245,247}
\definecolor{hidden-black}{RGB}{20,68,106}
\definecolor{purple}{RGB}{144,153,196}
\definecolor{yellow}{RGB}{255,228,123}
\definecolor{hidden-yellow}{RGB}{255,248,203}
\definecolor{tkcolor}{RGB}{224,223,255}
\definecolor{darkblue}{rgb}{0, 0.40, 0.75}
\newcommand{\eg}{\textit{e.g.,}}
\hypersetup{colorlinks=true, citecolor=darkblue, linkcolor=darkblue, urlcolor=darkblue}
\tcbset{
  aibox/.style={
    width=\linewidth,
    top=8pt,
    bottom=4pt,
    colback=blue!6!white,
    colframe=black,
    colbacktitle=black,
    enhanced,
    center,
    attach boxed title to top left={yshift=-0.1in,xshift=0.15in},
    boxed title style={boxrule=0pt,colframe=white,},
  }
}

\tikzstyle{my-box}=[
rectangle,
draw=hidden-black,
rounded corners,
text opacity=1,
minimum height=1.5em,
minimum width=5em,
inner sep=2pt,
align=center,
fill opacity=.5,
]
\tikzstyle{leaf}=[
my-box,
minimum height=1.5em,
fill=hidden-red!20,
text=black,
align=left,
font=\normalsize,
inner xsep=5pt,
inner ysep=4pt,
]
\tikzstyle{leaf2}=[
my-box,
minimum height=1.5em,
fill=hidden-green!20,
text=black,
align=left,
font=\normalsize,
inner xsep=5pt,
inner ysep=4pt,
]
\tikzstyle{leaf3}=[
my-box,
minimum height=1.5em,
fill=yellow!32,
text=black,
align=left,
font=\normalsize,
inner xsep=5pt,
inner ysep=4pt,
align=left,
text width=45em,
]
\tikzstyle{leaf4}=[
my-box,
minimum height=1.5em,
fill=hidden-blue!57,
text=black,
align=left,
font=\normalsize,
inner xsep=5pt,
inner ysep=4pt,
]
\tikzstyle{leaf5}=[
my-box,
minimum height=1.5em,
fill=darkblue!15,
text=black,
align=left,
font=\normalsize,
inner xsep=5pt,
inner ysep=4pt,
]
\tikzstyle{leaf6}=[
my-box,
minimum height=1.5em,
fill=purple!30,
text=black,
align=left,
font=\normalsize,
inner xsep=5pt,
inner ysep=4pt,
]

\newtcolorbox{AIbox}[2][]{aibox,title=#2,#1}

\tcbset{
  takeawaybox/.style={
    width=\linewidth,
    top=8pt,
    bottom=4pt,
    colback=hidden-yellow,
    colframe=black,
    colbacktitle=black,
    enhanced,
    center,
    attach boxed title to top left={yshift=-0.1in,xshift=0.15in},
    boxed title style={boxrule=0pt,colframe=white,},
  }
}

\newtcolorbox{TakeawayBox}[2][]{takeawaybox,title=#2,#1}

\newcommand{\leveltwoicon}{\textcolor{blue}{\small$\blacktriangle$}} 
\newcommand{\levelthreeicon}{\textcolor{red}{\small$\bigstar$}}      

\title{From \emph{AI for Science} to \emph{Agentic Science}: \\  A Survey on Autonomous Scientific Discovery}

\author{
  Jiaqi Wei$^{1,2*}$, 
  Yuejin Yang$^{1,3*}$, 
  Xiang Zhang$^{4*}$, 
  Yuhan Chen$^{1,5*}$, 
  \textbf{Xiang Zhuang}$^{1,2*}$, \\
  \textbf{Zhangyang Gao}$^{1*}$,  
  \textbf{Dongzhan Zhou}$^{1*}$,   
  \textbf{Guangshuai Wang}$^{1,3}$, 
  \textbf{Zhiqiang Gao}$^{1}$,  
  \textbf{Juntai Cao}$^{4}$, \\
  \textbf{Zijie Qiu}$^{1,3}$, 
  \textbf{Ming Hu}$^{1}$,
  \textbf{Chenglong Ma}$^{1,3}$,
  \textbf{Shixiang Tang}$^{1}$,
  \textbf{Junjun He}$^{1}$,
  \textbf{Chunfeng Song}$^{1}$,\\
  \textbf{Xuming He}$^{1,2}$, 
  \textbf{Qiang Zhang}$^{2}$, 
  \textbf{Chenyu You}$^{8}$,  
  \textbf{Shuangjia Zheng}$^{7,10}$, 
  \textbf{Ning Ding}$^{10}$, \\
  \textbf{Wanli Ouyang}$^{1,6}$, 
  \textbf{Nanqing Dong}$^{1}$,   
  \textbf{Yu Cheng}$^{1,6}$, 
  \textbf{Siqi Sun}$^{1,3\dagger}$, 
  \textbf{Lei Bai}$^{1\dagger}$,
  \textbf{Bowen Zhou}$^{1,10\dagger}$
\\

\vspace{1mm}

\normalfont{
$^1$ Shanghai Artificial Intelligence Laboratory, \vspace{-5pt}
$^2$ Zhejiang University, \\
$^3$ Fudan University, 
$^4$ University of British Columbia,
$^5$ Tongji University, \vspace{-5pt} \\
$^6$ The Chinese University of Hong Kong 
$^7$ Shanghai Jiaotong University, \vspace{-5pt}
$^8$ Stony Brook University, \\
$^{9}$ Lingang Laboratory, 
$^{10}$ Tsinghua University
}}

\begin{document}

\begin{abstract}
  \textbf{\large Abstract:}
  \vspace{1mm}

Artificial intelligence (AI) is reshaping scientific discovery, evolving from specialized computational tools into autonomous research partners. We position \textit{\textbf{Agentic Science}} as a pivotal stage within the broader \textit{\textbf{AI for Science}} paradigm, where AI systems progress from partial assistance to full scientific agency. Enabled by large language models (LLMs), multimodal systems, and integrated research platforms, agentic AI exhibits capabilities in hypothesis generation, experimental design, execution, analysis, and iterative refinement-behaviors once regarded as uniquely human. This survey offers a \textbf{domain-oriented review} of autonomous scientific discovery across life sciences, chemistry, materials, and physics, synthesizing research progress and advances within each discipline. We unify three previously fragmented perspectives-process-oriented, autonomy-oriented, and mechanism-oriented-through \textbf{a comprehensive framework }that connects foundational capabilities, core processes, and domain-specific realizations. Building on this framework, we (i) trace the evolution of AI for Science, (ii) identify five core capabilities underpinning scientific agency, (iii) model discovery as a dynamic four-stage workflow, (iv) review applications across life sciences, chemistry, materials science, and physics, and (v) synthesize key challenges and future opportunities. This work establishes a domain-oriented synthesis of autonomous scientific discovery and positions Agentic Science as a structured paradigm for advancing AI-driven research.

  \vspace{5mm}

  

  \vspace{1mm}
  \textbf{Keywords}: Agentic Science, Autonomous Scientific Discovery, Natural Sciences, AI for Science, Agentic AI, Large Language Models
  \vspace{6mm}








   \textbf {*: These authors contributed equally}
   
   \textbf {$\dag$: These authors jointly led the project}

  \vspace{3mm}

  \coloremojicode{1F3E0} \textbf{Homepage}: 
  \href{https://agenticscience.github.io/}{https://agenticscience.github.io/}

  \github{} \textbf{Github Repository}: 
  \href{https://github.com/AgenticScience/Awesome-Agent-Scientists}{https://github.com/AgenticScience/Awesome-Agent-Scientists}



    \vspace*{0.2in}

\end{abstract}

\maketitle

\pagestyle{headstyle}
\thispagestyle{empty}

\newpage
\vspace{2em}
\tableofcontents

\newpage
\input{Sections/1.Introduction}

\input{Sections/2.From}

\input{Sections/tree_taxonomy}

\input{Sections/3.Ability}

\input{Sections/4.Process}

\input{Sections/5.Life}

\input{Sections/6.Chemistry}

\input{Sections/7.Materials}

\input{Sections/8.Physics}

\input{Sections/9.Benchmarks}

\input{Sections/10.Challenge}

\input{Sections/11.Future}

\input{Sections/12.Conclusion}

\clearpage
\bibliographystyle{refstyle}
\bibliography{ref}

\end{document}

%% file: Sections/1.Introduction.tex
\section{Introduction}

\begin{figure}[!b]
    \centering
    \includegraphics[width=1.0\linewidth]{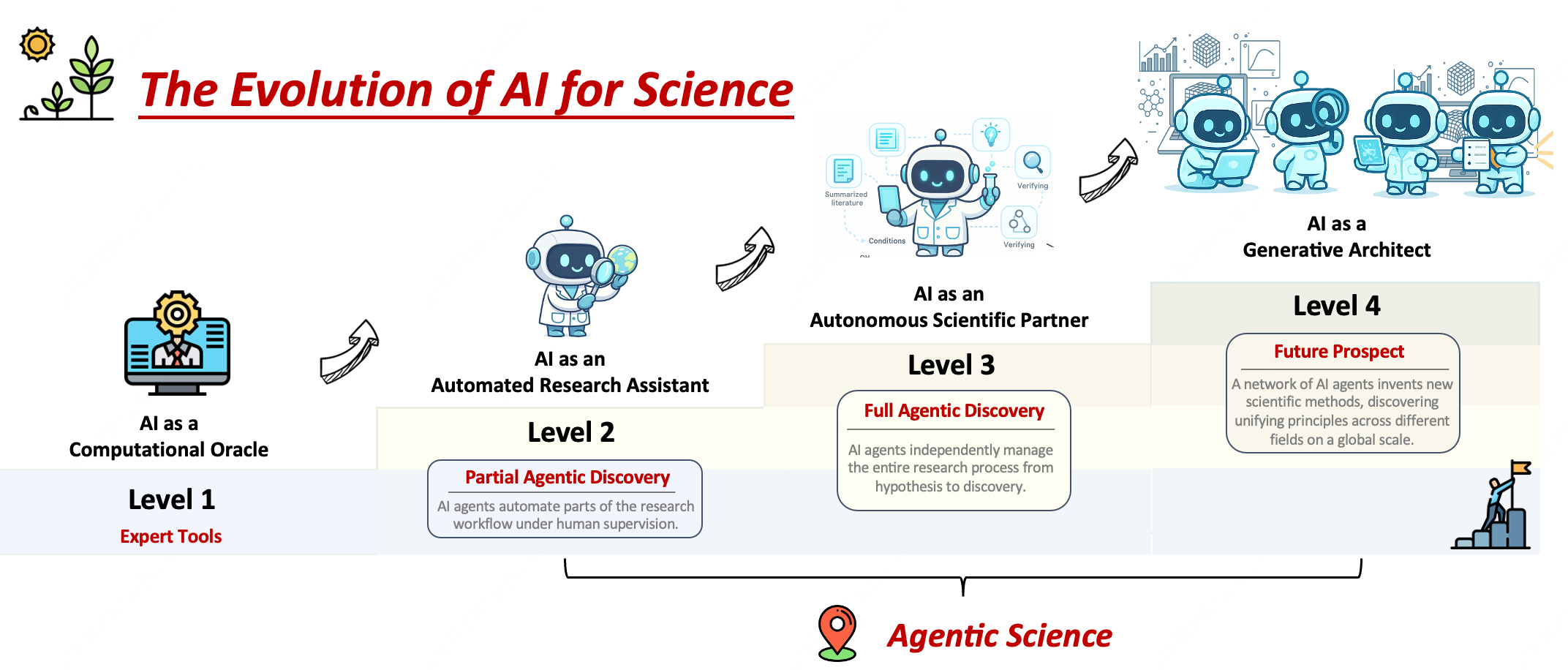}
    \caption{The Evolution of AI for Science. From computational tools to creative collaborators: the four-stage journey of AI in science. \textit{Agentic Science is a stage within AI for Science, aligning primarily with Level~3 (Full Agentic Discovery) and building on Level~2 (Partial Agentic Discovery).}}
    \label{fig:roadmap}
    \vspace{-0.em}
\end{figure}

Scientific discovery is experiencing a transformative shift, driven by the rapid evolution of artificial intelligence (AI) from specialized tools to collaborative research partners. This progression marks a pivotal stage in the \textit{\textbf{AI for Science}} paradigm, where AI systems have moved from acting as computational oracles for targeted tasks~\citep{jumper2021highly,yang2023alphafold2,zhang2025pi,cui2024scgpt,hao2024large,wei2023identification,zhou2025scientists} toward the emergence of \textit{\textbf{Agentic Science}} (Figure~\ref{fig:roadmap})~\cite{schneider2025generative,ren2025towards,gridach2025agentic,wang2025survey,liu2025advances}. Agentic Science denotes a specific stage within AI for Science evolution-corresponding primarily to Level~3 (Full Agentic Discovery) with precursors at Level~2 (Partial Agentic Discovery) in Figure~\ref{fig:roadmap}. In this stage, AI operates as an autonomous scientific agent capable of formulating hypotheses, designing and executing experiments, interpreting results, and iteratively refining theories with reduced dependence on human guidance~\citep{schneider2025generative,boiko2023autonomous}. This advancement is enabled by integrated platforms such as Intern-Discovery, which grants access to diverse AI agents and datasets, and by multimodal models like Intern-S1\footnote{https://github.com/InternLM/Intern-S1}, which demonstrate deep scientific reasoning.

This transformation is fueled by recent breakthroughs in foundational models, particularly large language models (LLMs)~\citep{guo2025deepseek, team2025kimi, zhou2025scientists, brown2020language,bai2023qwen,grattafiori2024llama}, which provide unprecedented capabilities in natural language understanding, complex reasoning, and tool use~\citep{sun2025survey, zhong2024evaluation, zeng2024good, zhang2025prompt, zhang2025scientific}.  These capabilities have enabled the development of AI agents that transcend static learning pipelines, acting instead as dynamic, goal-driven entities navigating the scientific method~\citep{yax2024studying, harris2025airus, ma2024sciagent,zhuang2025advancing}. From hypothesis generation~\citep{yang2023large, qi2023large} to autonomous experimentation~\citep{boiko2023autonomous, yuan2025dolphin} and synthetic dataset creation~\citep{li2025autosdt}, these agents demonstrate emergent behaviors once considered exclusively human.

\vspace{-1.em}
\paragraph{Comparison with Existing Surveys.}
Despite rapid progress, a unified framework for understanding and designing increasingly autonomous scientific systems remains absent. Existing surveys can be categorized along three complementary axes. \textbf{Process-oriented} perspectives map LLM capabilities onto the classical research cycle~\citep{luo2025llm4sr, zhou2025hypothesis, chen2025ai4research}. \textbf{Autonomy-oriented} studies grade systems by their initiative and responsibility~\citep{zheng2025automation, xie2025far}. \textbf{Mechanism-oriented} analyses dissect the architectural primitives and evolving roles that enable agentic behavior~\citep{ren2025towards, zhang2025evolving, gridach2025agentic, wang2025survey}. While these contributions lay valuable foundations, they remain fragmented-treating workflows, autonomy scales, or architectures in isolation.

\begin{figure}[!t]
    \centering
    \includegraphics[width=1.0\linewidth]{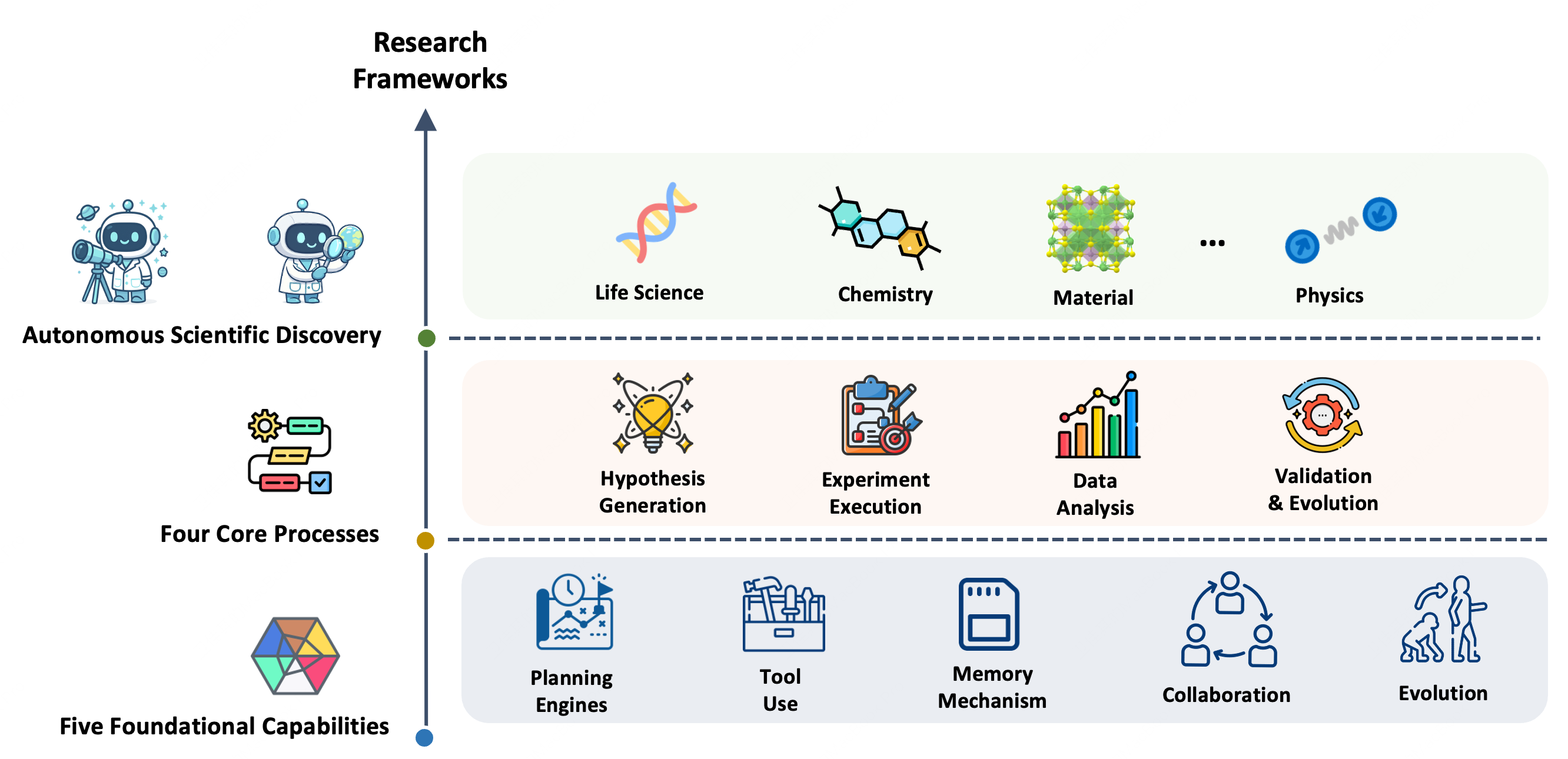}
    \vspace{-0.em}
    \caption{Research frameworks for Autonomous Scientific Discovery: Integrating Foundational Capabilities, Core Processes, and Research Levels across Life Sciences, Chemistry, Materials Science, and Physics.}
    \label{fig:level}
\end{figure}

\vspace{-1.em}
\paragraph{Our Contributions.}
In contrast to prior surveys, which examine process, autonomy, or architecture in isolation, our work integrates and extends these perspectives through the \textbf{comprehensive framework} shown in Figure~\ref{fig:level}. This framework connects \textit{foundational capabilities}, \textit{core processes}, and \textit{domain realizations} within autonomous scientific discovery. We present a \textbf{domain-oriented review} of autonomous scientific discovery across life sciences, chemistry, materials, and physics, providing a detailed synthesis of research progress and discoveries within each discipline. This unified lens positions \textbf{Agentic Science} not merely as an abstract stage, but as a structured research paradigm spanning capabilities, processes, and applications. Our contributions are as follows:

\begin{enumerate}
    \item \textbf{Charting the Evolution of AI for Science.} We trace the evolution from computational oracles to autonomous research partners, formally defining Agentic Science as the stage where AI systems display autonomy, goal-driven reasoning, and iterative learning.
    
    \item \textbf{The Anatomy of Scientific Agents: Five Core Capabilities.} We identify and analyze the five foundational capabilities required for a scientific agent: (i) Reasoning and Planning, (ii) Tool Integration, (iii) Memory Mechanisms, (iv) Multi-Agent Collaboration, and (v) Optimization and Evolution. For each, we review state-of-the-art implementations (e.g.,~\citep{lu2024ai, bran2023chemcrow, naumov2025dora, chai2025scimaster}) and domain-specific challenges.

    \item \textbf{The Dynamic Workflow of Agentic Science: Four Core Stages.} We model the scientific discovery process as a dynamic, four-stage workflow driven by agents: (i) Observation and Hypothesis Generation, (ii) Experimental Planning and Execution, (iii) Data and Result Analysis, and (iv) Synthesis, Validation, and Evolution. We emphasize that agents can flexibly and dynamically combine these stages to tackle complex scientific problems~\citep{baek2025researchagentiterativeresearchidea, boiko2023autonomous, ghareeb2025robin, ghafarollahi2025sparks}.

    \item \textbf{Systematic Review Across Natural Sciences.} We conduct a comprehensive review of agentic systems across the four major domains of natural science (Figure~\ref{fig:agentic-science-taxonomy}): \textbf{life sciences, chemistry, materials, and physics}. Our analysis spans more than a dozen distinct subfields, from drug discovery~\citep{yuan2025dolphin} to materials design~\citep{jansen2025codescientist}, showcasing the broad applicability and domain-specific innovations of Agentic Science.
    
    \item \textbf{Challenges and Future Opportunities.} We synthesize the principal technical, ethical, and philosophical challenges confronting the field--including reproducibility, validation of novel discoveries, and human-agent collaboration--and outline a research roadmap to guide the future development of robust, trustworthy, and impactful scientific agents.
    
\end{enumerate}

Through this synthesis, we aim to establish a conceptual and methodological foundation for Agentic Science, guiding future research toward the design of AI systems that co-evolve with human inquiry to accelerate the frontiers of discovery.

%% file: Sections/2.From.tex
\section{The Evolution of AI for Science: From Tools to Autonomous Partners}
\label{sec:paradigm_shift}

The landscape of AI in science is undergoing a fundamental transformation, evolving from narrowly-scoped computational augmentation to autonomous, end-to-end inquiry. This progression can be understood as an evolution through distinct levels of autonomy and capability, beginning with AI as a specialized tool and advancing towards AI as an autonomous scientific partner. This section delineates these levels, providing a formal description of each, and clarifies how they culminate in the emerging paradigm of Agentic Science.

\subsection{The Evolution of AI for Science}

\subsubsection{Level 1: AI as a Computational Oracle (Expert Tools)}

At the foundational level, AI operates as a \textbf{Computational Oracle}, a collection of highly specialized, non-agentic models designed to solve discrete, well-defined problems within a human-led workflow. These expert tools excel at tasks such as prediction and generation but lack autonomy; they function as sophisticated function approximators that require constant human guidance for task definition, execution, and interpretation of results. The core of the scientific process-from forming hypotheses to designing experiments-remains entirely in the hands of the human researcher.

Formally, a human scientist $H$ selects a model class $\mathcal{M}$ and a dataset $\mathcal{D} = \{(x_i, y_i)\}_{i=1}^N$ to train a model $M \in \mathcal{M}$. The objective is to find an optimal static function $M^*$ that minimizes a task-specific loss $\mathcal{L}_{\text{task}}$ over the dataset. This is typically achieved by minimizing the empirical risk:
\begin{equation}
M^* = \arg\min_{M \in \mathcal{M}} \frac{1}{N} \sum_{i=1}^{N} \mathcal{L}_{\text{task}}(M(x_i), y_i)
\end{equation}
The model's role is confined to producing an output $y_{pred} = M^*(x_{new})$ for a new input, with no capacity for subsequent independent action.

This paradigm has led to significant advances across the natural sciences. In life sciences, AI has revolutionized genomics~\cite{dalla2025nucleotide, avsec2025alphagenome, nguyen2024sequence}, proteomics~\cite{abramson2024accurate, jumper2021highly, rao2021msa, wen2020deep, zhang2025pi, qiu2025universal, hayes2025simulating, zhang2025curriculum, jun2025massnet, zhang2025bidirectionalrepresentationsaugmentedautoregressive, zhang2025bidirectional}, and single-cell analysis~\cite{hao2024large, cui2024scgpt}, medicine and healthcare~\cite{hu2024ophclip,li2024gmai,su2025gmai,yan2025derm1m,hu2025towards}. In chemistry, models like MolGPT~\cite{bagal2021molgpt}, ChemLLM~\cite{zhang2024chemllm}, and ChemMLLM~\cite{tan2025chemmllm} have accelerated molecular design and property prediction~\cite{jablonka202314}. Materials science has benefited from AI-driven discovery platforms~\cite{batatia2023foundation, yang2024mattersim, yan2022periodic, yan2024complete, zeni2025generative}. In physics and astronomy, AI tools are used for tasks ranging from modeling quantum systems~\cite{pan2025quantum, zhang2023transformer, wang2024grovergpt} and detecting phase transitions~\cite{greplova2020unsupervised, canabarro2019unveiling} to analyzing astronomical data~\cite{nguyen2023astrollama, gao2023deep, wei2023identification} and modeling fluid dynamics~\cite{dong2025fine, zhu2024fluid, lino2023current}.

\subsubsection{Level 2: AI as an Automated Research Assistant (Partial Agentic Discovery)}

The second level marks the introduction of AI as an \textbf{Automated Research Assistant}. Here, AI systems exhibit partial autonomy, functioning as agents that can execute specific, pre-defined stages of the research workflow. These agents can integrate multiple tools and carry out sequences of actions to complete well-defined sub-goals, such as running a series of experiments or performing a standardized data analysis pipeline. However, the high-level scientific direction, including the initial hypothesis, is still provided by human researchers. The agent operates within a structured environment to achieve a goal set by the human.

Formally, given a high-level goal $\mathcal{G}$ and a set of available tools $\mathcal{T}_{\text{tools}}$, the agent $\mathcal{A}$ follows a policy $\pi$ to select a sequence of actions $\{a_0, a_1, \dots, a_T\}$ that execute a sub-procedure of the scientific method. The agent's operation can be described as a finite-horizon decision process where the policy aims to complete the assigned task:
\begin{equation}
\{a_0, \dots, a_T\} \sim \pi(\cdot | \mathcal{G}, \mathcal{T}_{\text{tools}})
\end{equation}
The agent's autonomy is limited to the execution of this pre-defined sub-goal, after which control returns to the human scientist.

Examples of this level are becoming common. In life sciences, agents are used for bioinformatics workflow automation~\cite{xin2024bioinformatics, su2025biomaster, zhang2025transagent, xiao2024cellagent, liu2024toward} and experimental design~\cite{huang2024crispr, hao2025perturboagent, roohani2024biodiscoveryagent}. In chemistry, agentic systems are being developed for reaction optimization~\cite{chen2023chemist, ruan2024automatic} and automated experimentation~\cite{darvish2024organa, dai2024autonomous, strieth2024delocalized, song2025multiagent}. In materials science and physics, agents automate complex simulations~\cite{yue2025foam, pham2025chemgraph, ni2024mechagents, tian2025optimizing} and can assist in the design of power electronics~\cite{liu2024physics} and spacecraft control~\cite{maranto2024llmsat}.

\subsubsection{Level 3: AI as an Autonomous Scientific Partner (Full Agentic Discovery)}

The third and most advanced level envisions AI as an \textbf{Autonomous Scientific Partner}. In this paradigm, the AI agent possesses the ability to conduct the entire scientific discovery cycle independently. It can observe a domain, formulate novel and non-obvious hypotheses, design and execute experiments to test them, analyze the results, and iteratively refine its knowledge and strategy with minimal human intervention. This represents the full realization of Agentic Science, where the AI transitions from a tool or assistant to a genuine collaborator in the creation of knowledge.

Formally, the agent $\mathcal{A}$ operates in a continuous, open-ended discovery loop. Its objective is to maximize a measure of cumulative scientific utility, such as the expected information gain $\mathcal{I}(\cdot)$ about a set of evolving hypotheses $\mathcal{H}$. The agent's policy $\pi^*$ is optimized over an infinite horizon, constantly updating its state $s_t$ (which includes its knowledge base $\mathcal{K}_t$ and experimental evidence $\mathcal{E}_t$) and actions $a_t$ based on new information:
\begin{equation}
\pi^* = \arg\max_{\pi} \mathbb{E}_{\pi} \left[ \sum_{t=0}^{\infty} \gamma^t \, \mathcal{I}(\mathcal{H}_{t}; s_{t+1} | s_t, a_t) \right]
\end{equation}
Here, $\gamma \in [0, 1]$ is a discount factor, and the set of hypotheses $\mathcal{H}_t$ itself can be modified by the agent's actions. The human role shifts to that of a high-level strategist and validator, setting broad research directions and critically evaluating the agent's discoveries.

This paradigm is exemplified by several pioneering systems. Coscientist demonstrated the ability to autonomously research, design, and execute a chemical reaction~\cite{boiko2023autonomous}. In life sciences, Robin independently hypothesized and proposed a novel therapeutic use for an existing drug~\cite{ghareeb2025robin}, while OriGene acts as a self-evolving biologist for therapeutic target discovery~\cite{zhang2025origene}. Other notable examples include the AI Co-scientist~\cite{gottweis2025towards}, The Virtual Lab for nanobody design~\cite{swanson2024virtual}, ChemCrow for multi-purpose chemical research~\cite{bran2023chemcrow}, and MOFGen for discovering new materials~\cite{inizan2025system}. These systems point towards a future of human-agent co-discovery~\cite{lu2024ai, yuan2025dolphin, gottweis2025towards, su-etal-2025-many, zhang2025maps, ni2024matpilot}.

\begin{figure}[!b]
    \centering
    \includegraphics[width=0.7\linewidth]{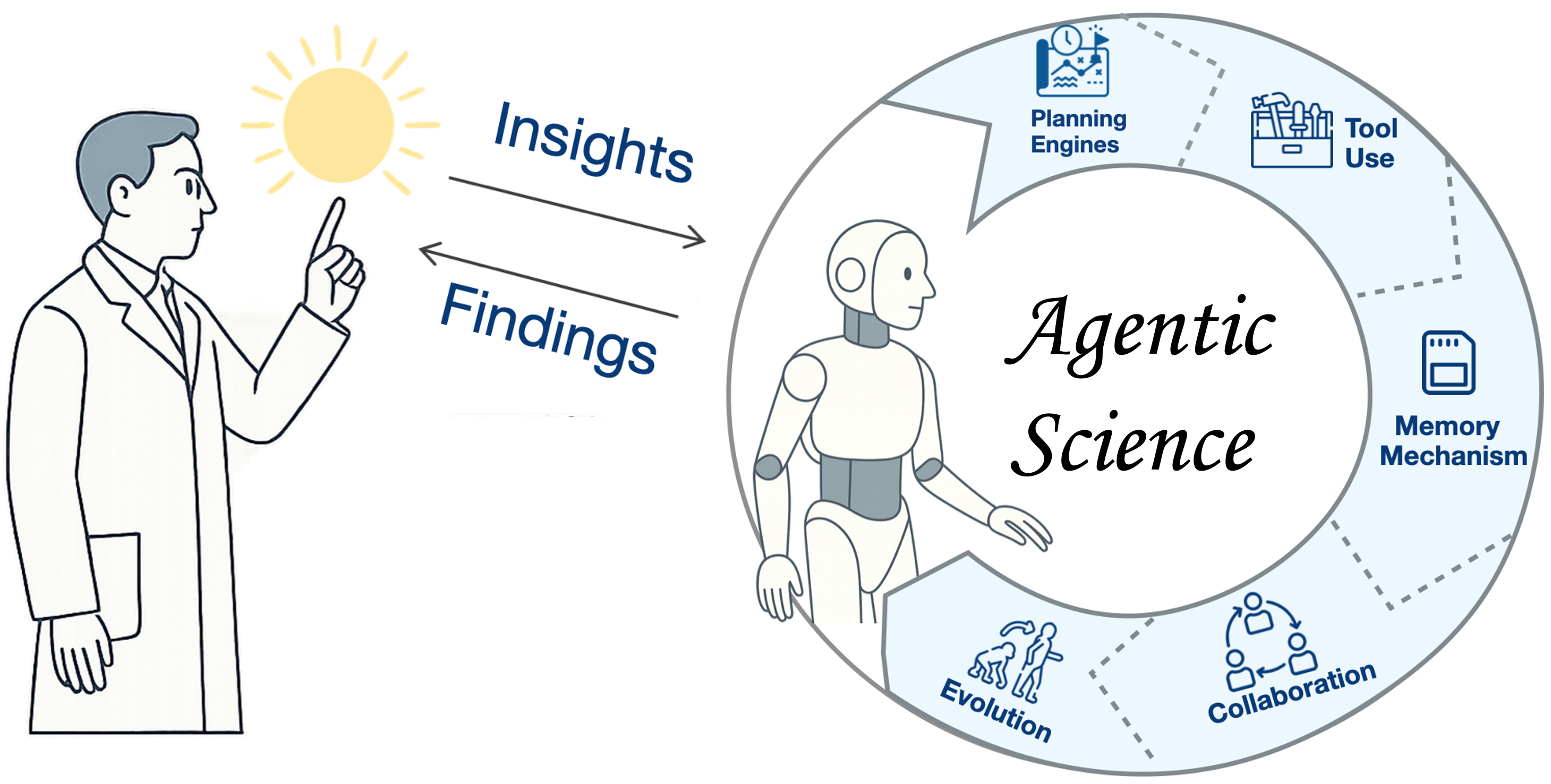}
    \vspace{-0.em}
    \caption{The Human-Agent Co-Discovery Loop. A human scientist provides high-level direction and the scientific agent operates autonomously within the Discovery Loop, guided by five key capabilities.}
    \label{fig:loop}
\end{figure}

\subsubsection{Level 4: AI as a Generative Architect (Future Prospect)}

The prospective fourth level represents AI as a \textbf{Generative Architect}, a system capable of not just working within existing scientific paradigms, but actively inventing new ones. This goes beyond discovering new facts to engaging in \textbf{autonomous invention}. Such agents would possess the capacity to design novel scientific instruments, create new experimental methodologies, or formulate new conceptual and mathematical frameworks to understand the world. The agent's role transcends discovery to become one of creation, moving from a "tool-user to tool-creator". This level also envisions agents as engines for large-scale \textbf{interdisciplinary synthesis}, uncovering latent connections between disparate scientific fields to forge unifying principles.

Formally, the agent's objective shifts from optimizing discoveries within a fixed scientific framework to generating new frameworks altogether. Let $\mathcal{F}$ be the space of all possible scientific frameworks (which includes methodologies, toolsets $\mathcal{T}_{\text{tools}}$, and conceptual models). The agent employs a generative policy $\pi_{gen}$ to create a new framework $f_{new} \in \mathcal{F}$ that maximizes a measure of \textbf{generative potential} $\Phi$, which quantifies the power and scope of the new scientific questions the framework enables. The objective is to find an optimal generative policy $\pi_{gen}^*$:
\begin{equation}
\pi_{gen}^* = \arg\max_{\pi_{gen}} \mathbb{E}_{f_{new} \sim \pi_{gen}(\cdot | \mathcal{K})} [ \Phi(f_{new}) ]
\end{equation}
Here, $\mathcal{K}$ represents the total accumulated knowledge of science. The ultimate realization of this level could be a \textbf{Global Cooperation Research Agent}, a decentralized ecosystem of specialized agents that collaborate, peer-review, and experiment at a planetary scale to solve grand challenges beyond human coordination. A benchmark for achieving this level could be the "Nobel-Turing Test," where an AI system makes a discovery worthy of a Nobel Prize.

\subsection{The Human Scientist's Evolving Role}
Agentic AI is reshaping science, moving the human role from executor to strategist (Figure~\ref{fig:loop}). Instead of focusing on \emph{how} tasks are done, scientists now define \emph{what} should be achieved and \emph{why} it matters. Their work centers on setting research goals, keeping methods ethical and reliable, and weaving results into coherent narratives.

Formally, the scientist provides a set of goals $G=\{g_1,\dots,g_m\}$ that guide the agent’s actions and embed safety, reproducibility, and ethical limits. Supervision involves checking the agent’s reasoning, validating results against domain knowledge, and stepping in when outputs drift from the intended path. In this way, humans ensure that autonomous discovery stays true to scientific standards and societal values.

This shift also calls for new skills. Scientists must learn to give clear, context-rich instructions that shape the agent’s policy $\pi$ (\emph{agent prompting}), manage the toolset $T_{\text{tools}}$ available to the agent, and judge when to trust its outputs versus when to apply deeper scrutiny. These practices will reshape research culture: labs may act as ``human-on-the-loop'' hubs coordinating many agents, funders may evaluate human-agent teams rather than fixed plans, and journals may require full agent traces—including reasoning logs, code, and tool use—for transparency and reproducibility.

Innovation itself becomes more distributed. Knowledge production shifts toward a \emph{human-agent system} where agents pursue subgoals $G_{\text{sub}}\subset G$ that advance larger programs. Such agents can explore huge problem spaces, find patterns beyond human reach, and connect ideas across disciplines. In doing so, agentic AI not only speeds discovery but also expands what science can ask and answer~\cite{ghafarollahi2024protagents}.

\subsection{Agentic Science: The Focus of This Survey}

This survey focuses on the emergent paradigm of Agentic Science, which encompasses both Level 2 and Level 3 of the described evolution. Agentic Science is characterized by the use of AI systems that are not merely passive tools but are active, goal-directed agents capable of autonomous reasoning, planning, and action within the scientific domain~\cite{schneider2025generative, gridach2025agentic, ren2025towards}. To systematically analyze this paradigm, we structure our review around a three-level research framework (Figure~\ref{fig:level}). At the base are the Five Foundational Capabilities (Section~\ref{sec:abilities}) that form the cognitive core of any scientific agent. These capabilities enable the Four Core Processes (Section~\ref{sec:process}) of the agentic discovery loop, which in turn drive progress in the highest level: Autonomous Scientific Discovery across various research domains like life sciences (Section~\ref{sec:life}), chemistry (Section~\ref{sec:chemistry}), materials (Section~\ref{sec:materials}), and physics (Section~\ref{sec:physics}). This hierarchical structure provides a comprehensive lens through which to examine and categorize developments in the field.

While Level 2 represents a form of \textit{task-level autonomy}, where agents automate specific parts of the research process, Level 3 signifies \textit{goal-level autonomy}, where agents can independently pursue high-level scientific objectives. The common thread is the concept of agency-the ability to act purposefully and independently in an environment to achieve a goal. The development of these agentic capabilities, from planning and reasoning engines~\cite{huang2024understanding, gou2023tora, goucritic, wang2022self, tao2024chain, long2023large, hu2023tree, browne2012survey, jiang2024technical} to tool use~\cite{qiao2023making, ma2024sciagent, song2024trial, zhao2024expel}, memory~\cite{shinn2023reflexion, wang2023voyager, packer2023memgpt}, collaboration~\cite{wu2023autogen, hong2024metagpt, li2023camel, chen2023reconcile, liu2024dynamic, liang2023encouraging, liang2024encouraging}, and evolution~\cite{madaan2023self, zelikman2024star, hosseini2024v}, is the central theme of this survey. By examining the progress and challenges within these levels, we aim to provide a comprehensive overview of how agentic AI is poised to reshape the future of scientific discovery.

\subsection{Modern Scientific Large Language Models for Agentic Science}
\label{sec:scillm_survey}

The conceptual shift from AI as a tool to an autonomous partner is not merely theoretical; it is propelled by concrete advancements in a class of models known as Scientific Large Language Models (Sci-LLMs). These models serve as the technological bedrock for the levels of autonomy previously described. While many specialized, domain-specific models exemplify the powerful function approximators of Level 1 \textbf{Computational Oracles}, the recent surge in general-purpose models with advanced reasoning and tool-use capabilities provides the foundation for Level 2 and 3 \textbf{Agentic Systems}. This section surveys the current landscape of Sci-LLMs, illustrating the practical implementations that enable this paradigm shift.

Current scientific LLMs are mainly developed from existing general-purpose models through several common strategies. A primary approach involves fine-tuning foundational models like LLaMA on curated, science-focused instruction datasets covering disciplines such as physics, chemistry, and materials science to enhance performance on specific tasks~\cite{xie2023darwin, zhang2024sciglm}. To build more robust scientific capabilities, another strategy involves large-scale pre-training or domain-adaptive pre-training on extensive scientific corpora, which include scholarly papers, textbooks, and specialized data formats like chemical formulae and protein sequences. This method strengthens a model's core understanding of scientific principles~\cite{taylor2022galactica, sun2024scidfm, prabhakar2025omniscience, bai2025interns1scientificmultimodalfoundation}. A third, more recent trend focuses on improving complex reasoning through test-time scaling. This is achieved by integrating techniques like Chain-of-Thought, large-scale Reinforcement Learning (RL), and massive Mixture-of-Experts (MoE) architectures, enabling models to process multimodal data and tackle multi-step scientific problems~\cite{guo2025deepseek, yang2025qwen3, team2025kimi, comanici2025gemini, grok4}. These distinct development strategies correspond to different levels of scientific autonomy: fine-tuning typically creates powerful Level 1 expert tools, while large-scale pre-training and enhanced reasoning are the key enablers for the agentic capabilities of Levels 2 and 3.

In many cases, domain-specific Sci-LLMs offer superior performance on specialized tasks. These models are constructed with well-curated datasets and training schemes tailored to a target subject. Below, we introduce recent domain-specific Sci-LLMs across eight scientific disciplines.

\vspace{-1em}
\paragraph{Life Sciences} This broad field has seen Sci-LLMs applied to multi-omics, molecular biology, and healthcare.
In \textbf{multi-omics}, development follows two paths: foundational models trained from scratch on biological sequences (DNA, RNA, protein) to learn fundamental representations~\cite{nguyen2024sequence, brixi2025genome, rives2021biological, esm2, hayes2025esm3}, and LLM-augmented systems that integrate these representations to enable conversational analysis, cross-modal translation, and controllable generation of biological sequences~\cite{llama-gene, prollama, proteindt, paag, pinal, proteingpt, proteinchat, protchatgpt, protranslator, biotranslator, prot2text, evolla}. Recent models aim to unify reasoning across different omics domains~\cite{naturelm, chatnt}.
In \textbf{molecular and cellular biology}, models are tuned for tasks like molecular property prediction from SMILES strings and analysis of single-cell genomic data, as well as de novo molecular design for drug discovery~\cite{liu2024moleculargpt, cui2024scgpt, edwards2022translation, hatakeyama2023prompt}.
In \textbf{healthcare}, models are often adapted from general-purpose LLMs. A common strategy is supervised fine-tuning (SFT) on medical dialogue and QA datasets~\cite{biomistral, biomedlm, toma2023clinical, han2023medalpaca, medpalm, apollo, huatuogpt}. More advanced models undergo continued pre-training on large medical corpora before SFT and reinforcement learning to improve performance and safety~\cite{pmcLLaMA, chen2023huatuogptii, tian2023chimed, yang2024zhongjing, meLLaMA, wang2025baichuan}. A critical frontier is multimodal AI, where models integrate medical imaging (e.g., X-rays) with text for report generation, visual question answering (VQA), and complex diagnostic reasoning~\cite{li2023llavamed, cxrllava, liu2023radiologyllama2bestinclasslargelanguage, medflamingo, huatuogptvision, li2025gmaivlgmaivl55mlarge, sellergren2025medgemma, chen2024huatuogpt, xu2025medground, su2025gmai}.

\vspace{-1em}
\paragraph{Chemistry} LLMs in chemistry are designed to understand and generate information across various data modalities, including molecular structures (SMILES), reaction data, and 3D conformations. By training on specialized datasets, these models can perform a range of core chemistry tasks, such as molecular property prediction, retrosynthesis analysis, and structure-based drug design, effectively bridging text, 2D representations, and 3D geometries~\cite{zhang2024chemllm, cao2023instructmol, zhao2024chemdfm, tan2025chemmllm, jiang2025chem3dllm}.

\vspace{-1em}
\paragraph{Materials Science} In materials science, transformer-based models are widely used for diverse applications. One approach employs encoders pre-trained on material representations like SMILES strings or scientific abstracts to predict material properties~\cite{smiles-bert, polybert, matbert, regtrans}. Another utilizes decoder-only, GPT-style architectures to generate novel molecules and crystal structures with desired features~\cite{molxpt, gptmolberta, Bagal2022, xyztrans, crystalm}. More specialized models are fine-tuned for specific applications such as predicting reaction equations, assessing crystal synthesizability, and modeling material mechanics, often integrating retrieval-augmented generation (RAG) with domain knowledge graphs~\cite{qwen2kg, lhs2rhs, csllm, buehler2024mechgpt}.

\vspace{-1em}
\paragraph{Physics and Astronomy} In physics, Sci-LLMs are evolving into interactive tools for scientific workflows. These models integrate language processing with physics engines and visual modules to perform tasks such as estimating physical parameters from visual data, learning solution operators for partial differential equations (PDEs), and serving as expert knowledge retrieval systems for high-energy physics~\cite{cherian2024llmphy, herde2024poseidon, zhang2024xiwu, astrollama-chat}. Astronomy-specific models are typically built upon general architectures like LLaMA and adapted through continual pre-training on astronomy literature (e.g., arXiv abstracts) and fine-tuning on domain-specific tasks~\cite{pathfinder, AstroMCQ, PCD, Astro-QA, PAPERCLIP, DESI-LS, astrobert}. These models enhance text understanding and generation for astronomical topics and, in their multimodal variants, integrate astronomical images to perform tasks like describing galaxy images from visual data~\cite{astrollama, astrollava, AstroVisBench}.

%% file: Sections/tree_taxonomy.tex
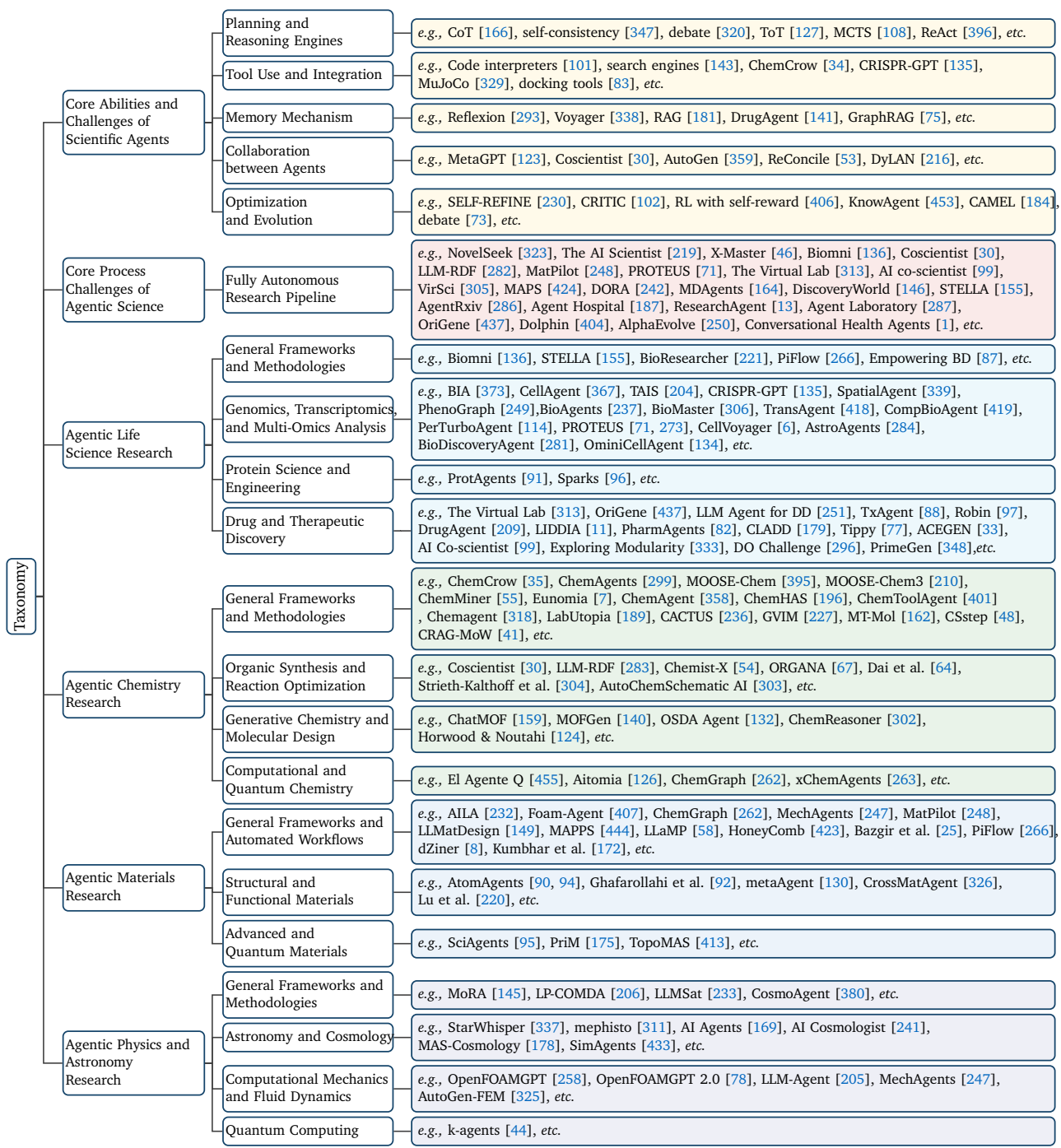
\begin{figure*}[!t]
        \vspace{-2mm}
        \centering
        \resizebox{0.96\textwidth}{!}{
        \begin{forest}
    forked edges,
    for tree={
        grow=east,
        reversed=true,
        anchor=base west,
        parent anchor=east,
        child anchor=west,
        base=left,
        font=\large,
        rectangle,
        draw=hidden-black,
        rounded corners,
        align=left,
        minimum width=4em,
        edge+={darkgray, line width=1pt},
        s sep=3pt,
        inner xsep=2pt,
        inner ysep=4pt,
        line width=1.1pt,
        ver/.style={rotate=90, child anchor=north, parent anchor=south, anchor=center},
    },
    where level=1{text width=9.5em,font=\normalsize,}{},
    where level=2{text width=11.5em,font=\normalsize,}{},
    where level=3{text width=12em,font=\normalsize,}{},
    where level=4{text width=50em,font=\normalsize,}{},
    [Taxonomy, ver
        [Core Abilities and \\ Challenges of \\ Scientific Agents
            [Planning and \\ Reasoning Engines
                [\eg~CoT~\cite{kojima2022large}{,} self-consistency~\cite{wang2022self}{,} debate~\cite{tao2024chain}{,} ToT~\cite{hu2023tree}{,} MCTS~\cite{guo2024can}{,} ReAct~\cite{yao2023react}{,} \textit{etc.}, leaf3, text width=44em]
            ]
            [Tool Use and Integration
                [\eg~Code interpreters~\cite{DBLP:conf/iclr/GouSGSYHDC24}{,} search engines~\cite{jablonka202314}{,} ChemCrow~\cite{M.Bran2024}{,} CRISPR-GPT~\cite{huang2024crispr}{,} \\ MuJoCo~\cite{todorov2012mujoco}{,} docking tools~\cite{10.1093/bfgp/elae011}{,} \textit{etc.}, leaf3, text width=44em]
            ]
            [Memory Mechanism
                [\eg~Reflexion~\cite{shinn2023reflexion}{,} Voyager~\cite{wang2023voyager}{,} RAG~\cite{lewis2020retrieval}{,} DrugAgent~\cite{inoue2024drugagent}{,} GraphRAG~\cite{edge2024graphrag}{,} \textit{etc.}, leaf3, text width=44em]
            ]
            [Collaboration \\ between Agents
                [\eg~MetaGPT~\cite{hong2024metagpt}{,} Coscientist~\cite{boiko2023autonomous}{,} AutoGen~\cite{wu2023autogen}{,} ReConcile~\cite{chen2023reconcile}{,} DyLAN~\cite{liu2024dynamic}{,} \textit{etc.}, leaf3, text width=44em]
            ]
            [Optimization \\ and Evolution
                [\eg~SELF-REFINE~\cite{madaan2023self}{,} CRITIC~\cite{goucritic}{,} RL with self-reward~\cite{yuan2024selfrewardinglanguagemodels}{,} KnowAgent~\cite{zhu2024knowagent}{,} CAMEL~\cite{li2023camel}{,} \\debate~\cite{du2023improving}{,} \textit{etc.}, leaf3, text width=44em]
            ]
        ]
        [Core Process \\ Challenges of \\ Agentic Science
            [Fully Autonomous \\ Research Pipeline
                [\eg~NovelSeek~\cite{team2025novelseek}{,} The AI Scientist~\cite{lu2024ai}{,} 
                X-Master~\cite{chai2025scimaster}{,} Biomni~\cite{huang2025biomni}{,} Coscientist~\cite{boiko2023autonomous}{,} \\
                LLM-RDF~\cite{ruan2024automatic}{,} 
                MatPilot~\cite{ni2024matpilot}{,} PROTEUS~\cite{ding2024automating}{,} 
                The Virtual Lab~\cite{swanson2024virtual}{,}
                AI co-scientist~\cite{gottweis2025towards}{,} \\
                VirSci~\cite{su-etal-2025-many}{,} 
                MAPS~\cite{zhang2025maps}{,} 
                DORA~\cite{naumov2025dora}{,} 
                MDAgents~\cite{kim2024mdagents}{,}
                DiscoveryWorld~\cite{jansen2024discoveryworld}{,}
                STELLA~\cite{jin2025stella}{,} \\
                AgentRxiv~\cite{schmidgall2025agentrxiv}{,} 
                Agent Hospital~\cite{li2024agent}{,} 
                ResearchAgent~\cite{baek2024researchagent}{,} 
                Agent Laboratory~\cite{schmidgall2025agent}{,} \\
                OriGene~\cite{zhang2025origene}{,}
                Dolphin~\cite{yuan2025dolphin}{,} 
                AlphaEvolve~\cite{novikov2025alphaevolve}{,} 
                Conversational Health Agents~\cite{abbasian2023conversational}{,}
                \textit{etc.}, leaf, text width=44em]
            ]
        ]
        [Agentic Life \\ Science Research
            [General Frameworks \\ and Methodologies
                [\eg~Biomni~\cite{huang2025biomni}{,} STELLA~\cite{jin2025stella}{,} BioResearcher~\cite{luo2025intention}{,} PiFlow~\cite{pu2025piflow}{,} Empowering BD~\cite{gao2024empowering}{,} \textit{etc.}, leaf4, text width=44em]
            ]
            [Genomics{,} Transcriptomics{,} \\ and Multi-Omics Analysis
                [\eg~BIA~\cite{xin2024bioinformatics}{,} CellAgent~\cite{xiao2024cellagent}{,} TAIS~\cite{liu2024toward}{,} CRISPR-GPT~\cite{huang2024crispr}{,} SpatialAgent~\cite{wang2025spatialagent}{,}  \\ PhenoGraph~\cite{niyakan2025phenograph}{,}BioAgents~\cite{mehandru2025bioagents}{,} BioMaster~\cite{su2025biomaster}{,} TransAgent~\cite{zhang2025transagent}{,} CompBioAgent~\cite{zhang2025compbioagent}{,} \\ PerTurboAgent~\cite{hao2025perturboagent}{,} PROTEUS~\cite{ding2024automating,qu2025automating}{,} CellVoyager~\cite{alber2025cellvoyager}{,} AstroAgents~\cite{saeedi2025astroagents}{,} \\ BioDiscoveryAgent~\cite{roohani2024biodiscoveryagent}{,} OminiCellAgent~\cite{Huang2025}{,} \textit{etc.}, leaf4, text width=44em]
            ]
            [Protein Science and \\ Engineering
                [\eg~ProtAgents~\cite{ghafarollahi2024protagents}{,} Sparks~\cite{ghafarollahi2025sparks}{,} \textit{etc.}, leaf4, text width=44em]
            ]
            [Drug and Therapeutic \\ Discovery
                [\eg~The Virtual Lab~\cite{swanson2024virtual}{,} OriGene~\cite{zhang2025origene}{,} LLM Agent for DD~\cite{ock2025large}{,} TxAgent~\cite{gao2025txagent}{,} Robin~\cite{ghareeb2025robin}{,} \\  DrugAgent~\cite{liu2024drugagent}{,} LIDDIA~\cite{averly2025liddia}{,} PharmAgents~\cite{gao2025pharmagents}{,} CLADD~\cite{lee2025rag}{,} Tippy~\cite{fehlis2025accelerating}{,} ACEGEN~\cite{bou2024acegen}{,} \\ AI Co-scientist~\cite{gottweis2025towards}{,} Exploring Modularity~\cite{van2025exploring}{,} DO Challenge~\cite{smbatyan2025can}{,} PrimeGen~\cite{wang2025accelerating}{,}\textit{etc.}, leaf4, text width=44em]
            ]
        ]
        [Agentic Chemistry \\ Research
            [General Frameworks \\ and Methodologies
                [\eg~ChemCrow~\cite{bran2023chemcrow}{,} ChemAgents~\cite{song2025multiagent}{,} MOOSE-Chem~\cite{yang2024moose}{,} MOOSE-Chem3~\cite{liu2025moose}{,} \\ ChemMiner~\cite{chen2024autonomous}{,} Eunomia~\cite{ansari2024agent}{,} ChemAgent~\cite{wu2025chemagent}{,} ChemHAS~\cite{li2025chemhas}{,} ChemToolAgent~\cite{yu2024chemtoolagent} \\{,} Chemagent~\cite{tang2025chemagent}{,} LabUtopia~\cite{li2025labutopia}{,} CACTUS~\cite{mcnaughton2024cactus}{,} GVIM~\cite{ma2025ai}{,} MT-Mol~\cite{kim2025mt}{,} CSstep~\cite{che2025csstep}{,} \\ CRAG-MoW~\cite{callahan2025agentic}{,} \textit{etc.}, leaf2, text width=44em]
            ]
            [Organic Synthesis and \\ Reaction Optimization
                [\eg~Coscientist~\cite{boiko2023autonomous}{,} LLM-RDF~\cite{ruan2024accelerated}{,} Chemist-X~\cite{chen2023chemist}{,} ORGANA~\cite{darvish2024organa}{,} Dai et al.~\cite{dai2024autonomous}{,} \\ Strieth-Kalthoff et al.~\cite{strieth2024delocalized}{,} AutoChemSchematic AI~\cite{srinivas2025autochemschematic}{,} \textit{etc.}, leaf2, text width=44em]
            ]
            [Generative Chemistry and \\ Molecular Design
                [\eg~ChatMOF~\cite{kang2023chatmof}{,} MOFGen~\cite{inizan2025system}{,} OSDA Agent~\cite{hu2025osda}{,} ChemReasoner~\cite{sprueill2024chemreasoner}{,} \\ Horwood \& Noutahi~\cite{horwood2020molecular}{,} \textit{etc.}, leaf2, text width=44em]
            ]
            [Computational and \\ Quantum Chemistry
                [\eg~El Agente Q~\cite{2025arXiv250502484Z}{,} Aitomia~\cite{hu2025aitomia}{,} ChemGraph~\cite{pham2025chemgraph}{,} xChemAgents~\cite{polat2025xchemagents}{,} \textit{etc.}, leaf2, text width=44em]
            ]
        ]
        [Agentic Materials \\ Research
            [General Frameworks and \\ Automated Workflows
                [\eg~AILA~\cite{mandal2024autonomous}{,} Foam-Agent~\cite{yue2025foam}{,} ChemGraph~\cite{pham2025chemgraph}{,} MechAgents~\cite{ni2024mechagents}{,} MatPilot~\cite{ni2024matpilot}{,} \\ LLMatDesign~\cite{jia2024llmatdesign}{,} MAPPS~\cite{zhou2025toward}{,} LLaMP~\cite{chiang2024llamp}{,} HoneyComb~\cite{zhang2024honeycomb}{,} Bazgir et al.~\cite{bazgir2025multicrossmodal}{,} PiFlow~\cite{pu2025piflow}{,} \\ dZiner~\cite{ansari2024dziner}{,} Kumbhar et al.~\cite{kumbhar2025hypothesis}{,} \textit{etc.}, leaf5, text width=44em]
            ]
            [Structural and \\ Functional Materials
                [\eg~AtomAgents~\cite{ghafarollahi2024atomagents, ghafarollahi2025automating}{,} Ghafarollahi et al.~\cite{ghafarollahi2024rapid}{,} metaAgent~\cite{hu2025electromagnetic}{,} CrossMatAgent~\cite{tian2025multi}{,} \\
                Lu et al.~\cite{lu2025agentic}{,} \textit{etc.}, leaf5, text width=44em]
            ]
            [Advanced and \\ Quantum Materials
                [\eg~SciAgents~\cite{ghafarollahi2025sciagents}{,} PriM~\cite{lai2025prim}{,} TopoMAS~\cite{zhang2025topomas}{,} \textit{etc.}, leaf5, text width=44em]
            ]
        ]
        [Agentic Physics and \\ Astronomy \\ Research
            [General Frameworks and \\ Methodologies
                [\eg~MoRA~\cite{jaiswal2024improving}{,} LP-COMDA~\cite{liu2024physics}{,} LLMSat~\cite{maranto2024llmsat}{,} CosmoAgent~\cite{xue2024if}{,} \textit{etc.}, leaf6, text width=44em]
            ]
            [Astronomy and Cosmology
                [\eg~StarWhisper~\cite{wang2024starwhisper}{,} mephisto~\cite{sun2024interpreting}{,} AI Agents~\cite{kostunin2025ai}{,} AI Cosmologist~\cite{moss2025ai}{,} \\ MAS-Cosmology~\cite{laverick2024multi}{,} SimAgents~\cite{zhang2025bridging}{,} \textit{etc.}, leaf6, text width=44em]
            ]
            [Computational Mechanics \\ and Fluid Dynamics
                [\eg~OpenFOAMGPT~\cite{pandey2025openfoamgpt}{,} OpenFOAMGPT 2.0~\cite{feng2025openfoamgpt}{,} LLM-Agent~\cite{liu2025large}{,} MechAgents~\cite{ni2024mechagents}{,} \\ AutoGen-FEM~\cite{tian2025optimizing}{,} \textit{etc.}, leaf6, text width=44em]
            ]
            [Quantum Computing
                [\eg~k-agents~\cite{cao2024agents}{,} \textit{etc.}, leaf6, text width=44em]
            ]
        ]
    ]
    \end{forest}
    }
  
        \caption{A comprehensive overview of the core abilities, challenges, and applications of scientific agents across various research domains, from life sciences to astronomy.}
        \label{fig:agentic-science-taxonomy}
\end{figure*}

%% file: Sections/3.Ability.tex
\section{Scientific Agents: Core Abilities and Challenges}
\label{sec:abilities}

\begin{figure}[!t]
    \centering
    \includegraphics[width=1.0\linewidth]{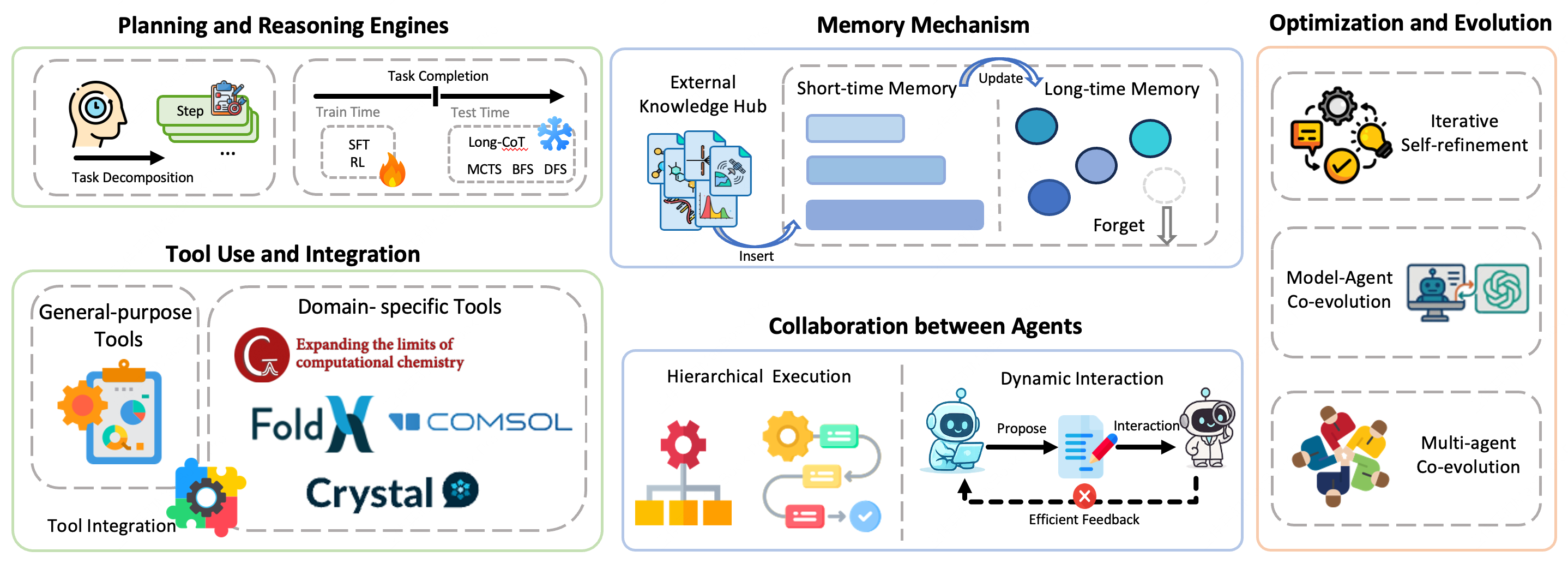}
    \caption{Core abilities of scientific agents.}
    \label{fig:ability}
    \vspace{-0.em}
\end{figure}

To transition from a specialized tool to an autonomous partner in discovery, a scientific agent must possess a suite of sophisticated, interconnected capabilities that collectively enable it to navigate the complexities of the research lifecycle. Unlike general-purpose agents, which are often designed for discrete, short-horizon tasks, a scientific agent must manage long-term, iterative, and empirically grounded workflows. This section breaks down the anatomy of such an agent by examining its five foundational pillars: the Planning and Reasoning Engine, which serves as its cognitive core; the ability to integrate and leverage external Tools to interact with the world; robust Memory Mechanisms for learning and knowledge accumulation; sophisticated Collaboration protocols for multi-agent systems; and the capacity for continuous Optimization and Evolution. For each of these core abilities, we will first review the state-of-the-art methodologies that empower them and then critically analyze the unique and significant challenges that emerge when these capabilities are applied to the high-stakes, verifiable domain of scientific inquiry (Figure~\ref{fig:ability}).

\begin{table*}[t]
\centering
\begin{threeparttable}
\caption{Structured capability taxonomy of \textbf{planning \& reasoning engines} in scientific agents. Rows are grouped into three major paradigms: reasoning structure, adaptation mechanisms, and interaction channels.}
\label{tab:planning}

\renewcommand{\arraystretch}{1.0}
\footnotesize
\setlength{\tabcolsep}{3pt}
\begin{tabularx}{\textwidth}{p{5.0cm} X X p{2.6cm}}

\toprule
\textbf{Paradigm} & \textbf{Purpose} & \textbf{Representative Mechanisms} & \textbf{Key References} \\
\midrule
\multicolumn{4}{l}{\textbf{I. Reasoning Structure}} \\
\rowcolor[HTML]{F5F5F5}
\quad Linear task decomposition & Sequentially break down complex goals & Plan-and-solve; zero-shot CoT; step-by-step execution & \cite{durfee2001distributed,wang2023plan,kojima2022large,wei2022chain} \\
\quad Robust linear planning & Improve reliability of linear chains & Self-consistency; majority voting; multi-agent debate & \cite{wang2022self,li2024enhancing,tao2024chain} \\
\rowcolor[HTML]{F5F5F5}
\quad Non-linear exploration & Explore multiple reasoning branches in parallel & Tree-of-Thought (ToT); backtracking search & \cite{hu2023tree,long2023large,choireactree} \\
\quad Search with lookahead & Handle uncertainty and long horizons & Monte Carlo Tree Search (MCTS); playouts; UCT & \cite{browne2012survey,guo2024can,liu2024large} \\
\midrule
\multicolumn{4}{l}{\textbf{II. Adaptation Mechanisms}} \\
\rowcolor[HTML]{F5F5F5}
\quad Dynamic plan adaptation & Adjust reasoning mid-execution & ReAct loops (thought–act–observe) & \cite{yao2023react,bhat2024grounding} \\
\quad Self-reflection & Detect and correct own errors & Reflection prompts; memory-based revision & \cite{wan2024dynamic,wei2025alignrag} \\
\rowcolor[HTML]{F5F5F5}
\quad Meta-controllers & Control search strategy adaptively & Budget tuning; step granularity adjustment & \cite{sun2023adaplanner,jafaripour2025adaptive} \\
\quad RL-augmented planning & Learn better policies over plans & Reinforcement learning; reward shaping & \cite{zhang2024rest,jiang2024technical} \\
\midrule
\multicolumn{4}{l}{\textbf{III. Interaction Channels}} \\
\rowcolor[HTML]{F5F5F5}
\quad Human-in-the-loop & Incorporate domain knowledge and correction & Interactive teaching; critique feedback & \cite{li2023traineragent,laleh2024survey,zhang2025prompt} \\
\quad Collaboration between Agents & Use diverse perspectives for robustness & Debate, negotiation, role specialization & \cite{tao2024chain,seo2024llm} \\
\rowcolor[HTML]{F5F5F5}
\quad Embodied / robotic execution & Apply planning to physical / simulated environments & Action abstraction; safety-aware exploration & \cite{lykov2023llm,ao2024llm,rivera2024conceptagent} \\
\quad Surveys \& meta-analysis & Provide taxonomies, evaluation frameworks & Systematic reviews; benchmark design & \cite{huang2024understanding,han2025mdocagent,wang2024rethinking} \\
\bottomrule
\end{tabularx}
\end{threeparttable}
\end{table*}

\subsection{Planning and Reasoning Engines}
The planning and reasoning engine is the cognitive core of a scientific agent, responsible for orchestrating the entire discovery process (Table~\ref{tab:planning}). This engine translates high-level scientific goals into a sequence of executable actions, such as formulating hypotheses, designing experiments, executing code, or querying databases. Effective planning is what enables an agent to navigate the vast and complex search space of scientific inquiry with purpose and efficiency~\cite{huang2024understanding,han2025mdocagent,wang2024rethinking}. The design of these engines can be broadly categorized by their approach to task decomposition and their ability to dynamically adapt through feedback.

A foundational approach to planning involves \textbf{task decomposition} through linear reasoning chains. This strategy breaks down a complex problem into a sequence of manageable subtasks. The simplest version is the plan-and-solve paradigm~\cite{durfee2001distributed, wang2023plan}, often implemented via zero-shot Chain-of-Thought (CoT) prompting~\cite{kojima2022large, wei2022chain}, where an agent first devises a sequential plan and then executes it step-by-step. While straightforward, this method can be brittle and prone to error accumulation. To improve robustness, many works enhance this linear process using ensemble-like methods, such as generating multiple reasoning chains and using self-consistency~\cite{wang2022self} or majority voting~\cite{li2024enhancing} to determine the best path, or even employing multiple agents to debate and refine the plan~\cite{tao2024chain}.

More sophisticated engines employ \textbf{non-linear, tree-based search and exploration} to navigate the solution space. Instead of committing to a single path, these methods explore multiple potential reasoning trajectories simultaneously. The Tree-of-Thought (ToT) approach~\cite{hu2023tree, long2023large, choireactree} exemplifies this, allowing an agent to evaluate different intermediate steps and backtrack from unpromising paths, which is crucial for tasks involving trial-and-error. This exploration can be guided by advanced algorithms like Monte Carlo Tree Search (MCTS)~\cite{browne2012survey}, a technique proven effective in complex domains like strategic game-playing~\cite{guo2024can, liu2024large} and robotics~\cite{lykov2023llm, ao2024llm, rivera2024conceptagent} that share similarities with the exploratory nature of scientific research. A critical element for both linear and tree-based approaches is dynamic plan adaptation via feedback and reflection. Modern agents rarely follow a static plan. Instead, they dynamically adjust their course using frameworks like ReAct~\cite{yao2023react}, which interleaves reasoning with action and observation. This allows the agent to incorporate feedback from diverse sources: the environment (e.g., the output of a simulation)~\cite{bhat2024grounding}, human guidance~\cite{li2023traineragent, laleh2024survey,zhang2025prompt}, model self-reflection~\cite{wan2024dynamic,wei2025alignrag}, or other collaborating agents~\cite{seo2024llm}. This continuous, iterative process of planning, acting, and refining~\cite{sun2023adaplanner, jafaripour2025adaptive}, often enhanced with reinforcement learning techniques~\cite{zhang2024rest, jiang2024technical}, is what endows agents with the adaptability needed to tackle the unpredictability of scientific discovery.

\paragraph{Challenges for Scientific Reasoning.}
Despite this progress, developing planning and reasoning engines for scientific discovery presents unique challenges not typically found in general domains. First, scientific planning operates under a paradigm of high-stakes and strict verifiability; a flawed plan can lead to wasted resources, incorrect conclusions, or even unsafe lab experiments, demanding a high degree of reliability and physical plausibility. Second, scientific inquiry often involves navigating vast, unstructured, and poorly-understood search spaces (e.g., all possible chemical compounds), requiring sophisticated strategies to balance exploration and exploitation \cite{todorov2012mujoco}. Third, the feedback loop in science is not simple text but often consists of noisy, multimodal experimental data that the agent must correctly interpret to refine its plan \cite{sprueill2024chemreasoner}. Finally, scientific agents must engage in long-horizon planning to manage multi-step research projects and aim for causal understanding \cite{sauter2023meta} rather than mere correlation, all while mitigating the risk of error accumulation that plagues sequential reasoning.

\begin{table*}[t]
\centering
\begin{threeparttable}
\caption{Structured capability taxonomy of \textbf{tool use and integration} in scientific agents. Rows are grouped into three major paradigms: foundational tools, domain-specific tools, and experimental/simulation platforms.}
\label{tab:tool}

\renewcommand{\arraystretch}{1.0}
\footnotesize
\setlength{\tabcolsep}{3pt}
\begin{tabularx}{\textwidth}{p{5.0cm} X X p{2.6cm}}

\toprule
\textbf{Paradigm} & \textbf{Purpose} & \textbf{Representative Mechanisms} & \textbf{Key References} \\
\midrule

\multicolumn{4}{l}{\textbf{I. Foundational, general-purpose tools}} \\
\rowcolor[HTML]{F5F5F5}
\quad Information retrieval & Access external knowledge beyond model memory & Search engines; scientific databases; integration with retrieval APIs & \cite{qiao2023making,yang2023gpt4tools,yuan2024easytool,jablonka202314,thulke2024climategpt} \\
\quad Computational utilities & Solve well-defined problems; perform symbolic and numerical analysis & Code interpreters; mathematical libraries (SymPy, SciPy) & \cite{DBLP:conf/iclr/GouSGSYHDC24,lai2022ds1000naturalreliablebenchmark} \\

\midrule

\multicolumn{4}{l}{\textbf{II. Domain-specific computational and analytical tools}} \\
\rowcolor[HTML]{F5F5F5}
\quad Chemistry \& materials science & Predict reactions; estimate properties; integrate scientific APIs & ChemCrow; CACTUS; HoneyComb & \cite{M.Bran2024,mcnaughton2024cactus,zhang-etal-2024-honeycomb} \\
\quad Biology \& genomics & Support genome editing and bioinformatics workflows & CRISPR-GPT; domain bioinformatics suites & \cite{huang2024crispr} \\
\rowcolor[HTML]{F5F5F5}
\quad Multi-domain tool hubs & Generalize integration across diverse scientific domains & SciAgent; extensible toolkits for physics, finance, materials & \cite{DBLP:conf/emnlp/MaGHXWPY0S24} \\

\midrule

\multicolumn{4}{l}{\textbf{III. Experimental and simulation platforms}} \\
\quad Physical dynamics engines & Model real-world physics for hypothesis testing & MuJoCo; physics simulation engines & \cite{DBLP:conf/iclr/LiuWGWVCZD23,todorov2012mujoco} \\
\rowcolor[HTML]{F5F5F5}
\quad Engineering \& climatology models & Evaluate designs and environmental impacts & MyCrunchGPT; ClimSight; computational fluid dynamics models & \cite{kumar2023mycrunchgpt,Koldunov2024} \\
\quad Molecular docking simulators & Guide molecule generation via docking feedback & DockingGA; docking-based evaluation loops & \cite{10.1093/bfgp/elae011} \\
\bottomrule
\end{tabularx}
\end{threeparttable}
\end{table*}

\subsection{Tool Use and Integration}

The capacity to harness external tools is essential for scientific agents, enabling them to overcome the intrinsic constraints of language models in computation, data access and interaction with the physical world~\cite{masterman2024landscape, shi2025tool, ferrag2025llm}. Tool integration within scientific workflows can be organised by function, spanning from foundational utilities to highly specialised experimental platforms (Table~\ref{tab:tool}). 

The first tier comprises \textbf{foundational, general-purpose tools} that provide essential computational and informational capabilities. As with general-purpose agents, scientific agents must determine both the timing and the manner of tool use~\cite{qiao2023making, yang2023gpt4tools, yuan2024easytool}. These include search engines and databases for information retrieval, as demonstrated by MAPI-LLM~\cite{jablonka202314} and ClimateGPT~\cite{thulke2024climategpt}, as well as code interpreters and mathematical libraries such as SymPy and SciPy, which support the solution of well-defined problems. Such capacities are refined in systems like Tora~\cite{DBLP:conf/iclr/GouSGSYHDC24} and assessed in data science benchmarks~\cite{lai2022ds1000naturalreliablebenchmark}. These tools form the bedrock upon which higher-order scientific reasoning is constructed.

Building upon this foundation, the second category comprises \textbf{domain-specific computational and analytical tools}. These encapsulate expert knowledge and advanced algorithms, enabling agents to address complex scientific questions. In chemistry and materials science, for instance, ChemCrow \cite{M.Bran2024} and CACTUS \cite{mcnaughton2024cactus} integrate specialized toolkits for reaction prediction and molecular property estimation. HoneyComb \cite{zhang-etal-2024-honeycomb} combines a materials science knowledge base with a hub of domain-specific APIs. In biology, CRISPR-GPT \cite{huang2024crispr} integrates a suite of bioinformatics tools for genome-editing experiment design. Frameworks such as SciAgent \cite{DBLP:conf/emnlp/MaGHXWPY0S24} illustrate how tool-use capabilities can be generalized across diverse scientific domains, from physics to finance, through the development of comprehensive, multi-domain toolsets. Such deep integration enables agents to perform specialized analyses that would otherwise be intractable.

The third and most advanced category centres on \textbf{experimental and simulation tools} for hypothesis validation. This capability is essential for emulating the scientific method, as it enables agents to actively test hypotheses and generate new empirical data. Agents can interact with high-fidelity simulators to investigate complex systems. For example, physics engines such as MuJoCo have been employed to reason about physical dynamics \cite{DBLP:conf/iclr/LiuWGWVCZD23, todorov2012mujoco}. In engineering and climatology, systems including MyCrunchGPT \cite{kumar2023mycrunchgpt} and ClimSight \cite{Koldunov2024} integrate computational fluid dynamics and climate models to optimise designs and evaluate environmental impacts. Similarly, DockingGA \cite{10.1093/bfgp/elae011} employs molecular docking simulations as a feedback mechanism for guiding molecular generation. By engaging with such virtual laboratories, agents can iteratively refine their understanding and uncover novel scientific insights.

\paragraph{Challenges in Scientific Tool Use.}
Notwithstanding these advances, the integration of tools into scientific agents presents distinctive challenges that extend beyond those encountered by general-purpose agents. First, scientific tools demand exceptional precision and deep domain-specific understanding. Unlike a routine web search, even a minor error in parameterizing a bioinformatics tool or a physics simulation can yield scientifically invalid outcomes, making it essential for agents to interpret complex documentation and scientific context accurately. Second, reproducibility and provenance are non-negotiable in scientific research: an agent must not only execute a tool correctly but also record, with meticulous detail, the tool versions, parameters, and data lineage to enable independent verification of its findings. Third, scientific discovery often necessitates the construction of complex, interoperable workflows that chain multiple specialized tools--a process hindered by heterogeneous interfaces and non-standardised data formats. As benchmarks such as ShortcutsBench \cite{haiyang2025shortcutsbench} demonstrate, managing dependencies and adapting to API changes constitute significant obstacles, further exacerbated by the rapid evolution of the scientific software ecosystem. Finally, many high-fidelity simulators and proprietary databases impose substantial computational and financial costs, requiring agents to conduct rigorous cost–benefit analyses and apply effective resource management to ensure research efficiency.

\begin{table*}[t]
\centering
\begin{threeparttable}
\caption{Structured capability taxonomy of \textbf{memory mechanisms} in scientific agents. Rows are grouped into two major paradigms: memory for iterative task execution and memory as a knowledge hub.}
\label{tab:memory}

\renewcommand{\arraystretch}{1.0}
\footnotesize
\setlength{\tabcolsep}{3pt}
\begin{tabularx}{\textwidth}{p{5.2cm} X X p{2.3cm}}

\toprule
\textbf{Paradigm} & \textbf{Purpose} & \textbf{Representative Mechanisms} & \textbf{Key References} \\
\midrule
\multicolumn{4}{l}{\textbf{I. Memory for iterative task execution}} \\
\rowcolor[HTML]{F5F5F5}
\quad Short-term \& feedback memory & Maintain coherent understanding of ongoing tasks; incorporate observations & Dialogue/context windows; tool-output logs; thought–act–observe loops (ReAct) & \cite{yao2023react} \\
\quad Experience repositories & Learn from past attempts to refine strategies & Episodic traces; critique-and-retry; outcome-indexed memory & \cite{shinn2023reflexion,zhao2024expel} \\
\rowcolor[HTML]{F5F5F5}
\quad Reusable skill libraries & Turn successful action sequences into reusable skills/macros & Skill discovery; action chunking; competence tracking & \cite{wang2023voyager,ghafarollahi2024atomagents} \\
\quad Memory governance \& persistence & Decide what to store, summarize, and forget across trials & Summarization, saliency filters, lifetime management & \cite{xu2025mem,mei2025survey} \\

\midrule
\multicolumn{4}{l}{\textbf{II. Memory as a knowledge hub}} \\
\rowcolor[HTML]{F5F5F5}
\quad RAG  & Ground reasoning in external literature; go beyond parametric memory & Dense/sparse retrieval; chunking; reranking; cite-while-generate & \cite{lewis2020retrieval,wei2025alignrag} \\
\quad Literature assistants & Automate literature review and evidence aggregation & PaperQA; LitLLM toolkit; query planning + citation tracking & \cite{Lala2023PaperQA,agarwal2024litllm} \\
\rowcolor[HTML]{F5F5F5}
\quad Structured knowledge graphs & Enforce consistency with scientific ontologies; relational reasoning & GraphRAG; domain KGs; constraint checking & \cite{edge2024graphrag,ghafarollahi2024sciagentsautomatingscientificdiscovery} \\
\quad Specialized scientific databases & Retrieve task-specific entities (e.g., drug–target data) & API connectors; schema-aware queries; evidence linking & \cite{inoue2024drugagent} \\
\rowcolor[HTML]{F5F5F5}
\quad Interleaved retrieval–reasoning & Pull information at intermediate steps of multi-hop reasoning & Step-wise retrieval policies; tool-augmented CoT; planning-integrated RAG & \cite{trivedi2022interleaving} \\
\quad Tiered memory architectures & Manage long-term external memory with internal working memory & Memory managers; paging; pointers to external stores (MemGPT) & \cite{packer2023memgpt} \\
\rowcolor[HTML]{F5F5F5}
\quad Surveys \& meta-frameworks & Synthesize patterns; provide evaluation and design guidance & Taxonomies; memory governance principles; benchmarks & \cite{xu2025mem,mei2025survey} \\
\bottomrule
\end{tabularx}
\end{threeparttable}
\end{table*}

\subsection{Memory Mechanisms}

Memory is a foundational capability of agentic intelligence, enabling agents to retain information, learn from experience, and maintain context during complex tasks~\cite{xu2025mem,mei2025survey}. For scientific agents, memory mechanisms are not just about recalling past dialogue but are fundamental to emulating the scientific process of iterative refinement, knowledge accumulation, and hypothesis testing. We categorize memory mechanisms based on their functional role in the agent's workflow: memory for iterative task execution, which supports in-context learning and adaptation, and memory as a knowledge hub, which connects the agent to vast external information repositories (Table~\ref{tab:memory}).

First, memory for iterative task execution allows an agent to maintain a \textbf{coherent understanding} of an ongoing research task by storing and reflecting on its recent history. This includes short-term context from dialogues and environmental feedback, as seen in frameworks like ReAct \cite{yao2023react}, as well as more structured memory derived from the agent's own actions. For instance, agents can learn from both successes and failures by building experience repositories, a technique central to Reflexion \cite{shinn2023reflexion} and ExpeL \cite{zhao2024expel}, allowing them to refine their strategies over successive trials. This experiential memory can be further structured into reusable skill libraries, where successful action sequences are codified for future use, as demonstrated by Voyager \cite{wang2023voyager} in exploration tasks and AtomAgents \cite{ghafarollahi2024atomagents} through dedicated tool memory. This form of memory transforms short-lived interactions into persistent, actionable knowledge that guides the agent through the cycles of scientific inquiry.

Second, memory as a \textbf{knowledge hub} extends an agent's capabilities by integrating external information sources, grounding its reasoning in established scientific knowledge. The most prevalent approach is Retrieval-Augmented Generation (RAG) \cite{lewis2020retrieval, wei2025alignrag}, which dynamically fetches relevant information from text corpora. This is crucial for tasks like automated literature review, as seen in PaperQA \cite{Lala2023PaperQA} and the LitLLM toolkit \cite{agarwal2024litllm}. Beyond unstructured text, scientific agents leverage structured knowledge graphs to ensure their hypotheses are consistent with known scientific concepts \cite{ghafarollahi2024sciagentsautomatingscientificdiscovery, edge2024graphrag}. Some agents, like DrugAgent \cite{inoue2024drugagent}, query specialized databases to retrieve specific information like drug-target interactions. Advanced architectures interleave retrieval with reasoning steps \cite{trivedi2022interleaving} or use tiered memory systems like MemGPT \cite{packer2023memgpt} to efficiently manage both internal context and external knowledge, enabling agents to surpass their training data and engage with the vast, ever-expanding body of scientific information.

\paragraph{Challenges in Scientific Memory Mechanisms.}
Despite these advancements, memory for scientific agents presents distinct and significant challenges. First, the accuracy and decay of scientific knowledge is a critical hurdle; information in scientific fields can become outdated, and agents must be able to validate and update their memory to avoid relying on superseded facts. Second, scientific data is inherently heterogeneous and multi-modal, comprising not just text but also tables, chemical structures, genomic sequences, and experimental imagery. Current memory architectures are ill-equipped to store, retrieve, and reason across these diverse data types seamlessly. Finally, scientific discovery often involves long-term causal reasoning, where insights gained months or even years into a project may depend on early, seemingly minor experimental results. Existing memory systems lack the capacity to maintain such extended, causally-linked histories with high fidelity, which is essential for ensuring the reproducibility of agent-driven discoveries--a foundation of the scientific method.

\begin{table*}[t]
\centering
\begin{threeparttable}
\caption{Structured capability taxonomy of \textbf{collaboration strategies} in multi-agent scientific systems. Rows are grouped into three major strategies: hierarchical task execution, deliberative refinement, and dynamic adaptive topologies.}
\label{tab:collaboration}

\renewcommand{\arraystretch}{1.0}
\footnotesize
\setlength{\tabcolsep}{3pt}
\begin{tabularx}{\textwidth}{p{5.2cm} X X p{2.3cm}}

\toprule
\textbf{Paradigm} & \textbf{Purpose} & \textbf{Representative Mechanisms} & \textbf{Key References} \\
\midrule
\multicolumn{4}{l}{\textbf{I. Hierarchical task execution}} \\
\rowcolor[HTML]{F5F5F5}
\quad Manager–worker decomposition & Break complex goals into subtasks assigned to specialized agents & Workflow controllers; SOP-driven orchestration (MetaGPT, ChatDev) & \cite{hong2024metagpt,qian-etal-2024-chatdev} \\
\quad Role differentiation & Assign functional roles (planner, executor, tool-user) & Role-playing agents; meta-prompting & \cite{suzgun2024meta,qiao2024autoact,li2023camel} \\
\rowcolor[HTML]{F5F5F5}
\quad Structured scientific workflows & Organize multi-phase experiments or multi-tier planning & Coscientist; AFlow & \cite{boiko2023autonomous,zhang2025aflow} \\

\midrule
\multicolumn{4}{l}{\textbf{II. Deliberative refinement}} \\
\rowcolor[HTML]{F5F5F5}
\quad Dialogue-based debate & Improve factuality and reasoning via structured exchanges & Peer review dialogue; group-chat settings (AutoGen, MAD, MDebate) & \cite{wu2024autogen,liang2023encouraging,du2023improving,kim2024can} \\
\quad Iterative peer refinement & Incrementally improve shared outputs through critique \& revision & Reflexion; ReConcile; METAL; DS-Agent & \cite{shinn2023reflexion,wei2025alignrag,chen2023reconcile,li2025metal,guo2024ds} \\
\rowcolor[HTML]{F5F5F5}
\quad Competitive vs cooperative roles & Assign critic/actor or explainer/evaluator to balance strengths & LEGO; Reflexion (actor–evaluator loops) & \cite{he-etal-2023-lego,shinn2023reflexion} \\
\quad Ensemble and consensus methods & Aggregate multiple solutions for robustness & Majority voting; best-component synthesis; expert review & \cite{jiang2023llmlingua,zhang-etal-2024-exploring,khan2024debating,darcy2024margmultiagentreviewgeneration,tang-etal-2024-medagents} \\

\midrule
\multicolumn{4}{l}{\textbf{III. Dynamic adaptive topologies}} \\
\rowcolor[HTML]{F5F5F5}
\quad Predefined communication patterns & Select topology based on task (relay, debate, star, tree) & Exchange-of-Thought (EoT) & \cite{yin-etal-2023-exchange} \\
\quad Learned adaptive routing & Dynamically reconfigure collaboration graph by task complexity & MDAgents; DyLAN; graph-based orchestrators (DAGs) & \cite{kim2024mdagents,liu2024a,jeyakumar2024advancing} \\
\rowcolor[HTML]{F5F5F5}
\quad Teacher–student policies & Learn adaptive collaboration through meta-learning & Teacher–student frameworks for adaptive teaming & \cite{li2021learning} \\
\bottomrule
\end{tabularx}
\end{threeparttable}
\end{table*}

\subsection{Collaboration between Agents}

Effective collaboration enables multi-agent systems to address complex scientific problems that surpass the capabilities of any single agent~\cite{dafoe2020openproblemscooperativeai, das2023enabling,yu2025survey,wang2025comprehensive}. By distributing tasks, synthesizing diverse information, and iteratively refining solutions, collaborative frameworks enhance both the robustness and creativity of agentic research~\cite{tran2025multi, guo2024large}. The mechanisms governing such interactions can be broadly categorized by their primary collaborative strategy: structured workflows, deliberative refinement, and dynamic adaptation (Table~\ref{tab:collaboration}).

A prominent strategy is \textbf{hierarchical task execution}, where a structured, top-down approach is employed for problem-solving. In these systems, a primary agent or predefined workflow decomposes a complex goal into smaller, tractable subtasks, which are then assigned to specialized agents. This manager–worker paradigm emphasizes efficiency and organized execution. For example, MetaGPT~\cite{hong2024metagpt} implements this paradigm by assigning role-specific agents within a simulated software company, using Standard Operating Procedures (SOPs) to formalize coherent workflows. Similarly, ChatDev~\cite{qian-etal-2024-chatdev} simulates a software development team with distinct roles. Other frameworks achieve this through explicit task decomposition by a controller, as in Meta-Prompting~\cite{suzgun2024meta}, or by differentiating a single model into functional roles such as planner and tool-user~\cite{qiao2024autoact}. This hierarchical approach is particularly suited to structured scientific workflows, such as managing experimental phases in Coscientist~\cite{boiko2023autonomous}, coordinating multi-tier planning in AFlow~\cite{zhang2025aflow}, or structuring complex role-playing tasks as in CAMEL~\cite{li2023camel}.

A second major strategy is \textbf{deliberative, refinement-based collaboration,} which seeks to improve solution quality through iterative peer interaction, often inspired by scholarly debate and review. Some systems facilitate direct dialogue, where agents engage in structured exchanges to challenge and refine each other's ideas, a process shown to improve factuality and reasoning~\cite{du2023improving, xiong-etal-2023-examining, liang2023encouraging}. AutoGen~\cite{wu2024autogen} supports this through a group-chat setting, while frameworks such as MAD~\cite{liang2023encouraging}, MADR~\cite{kim2024can}, and MDebate~\cite{du2023improving} define specific protocols to encourage critical feedback and consensus-building. Another approach involves independent refinement of a shared solution. This may be competitive, as in LEGO~\cite{he-etal-2023-lego}, where an “Explainer” is critiqued by a “Critic,” or cooperative, as in Reflexion~\cite{shinn2023reflexion, wei2025alignrag}, where an “Evaluator” provides feedback to an “Actor.” In ReConcile~\cite{chen2023reconcile}, agents iteratively improve a common answer, while METAL~\cite{li2025metal} and DS-Agent~\cite{guo2024ds} employ specialized revision agents. This category also encompasses ensemble methods, where multiple agent-generated solutions are synthesized--either by combining the best components~\cite{jiang2023llmlingua} or through consensus mechanisms such as majority voting~\cite{zhang-etal-2024-exploring, khan2024debating} or expert consultation, as in scientific review systems like MARG~\cite{darcy2024margmultiagentreviewgeneration} and medical diagnosis frameworks like MedAgents~\cite{tang-etal-2024-medagents}.

The most advanced systems employ \textbf{dynamic, adaptive topologies,} in which the interaction structure itself is flexible and optimized for the task. Rather than relying on fixed hierarchies or communication patterns, these frameworks reconfigure their collaboration graph in response to task requirements or real-time performance feedback. Some offer predefined topologies, such as communication paradigms (e.g., relay, debate) and network structures (e.g., star, tree) in EoT~\cite{yin-etal-2023-exchange}, which can be selected to match a specific problem. Others learn the optimal structure dynamically. For instance, MDAgents~\cite{kim2024mdagents} routes tasks to different collaborative configurations based on an initial complexity assessment. DyLAN~\cite{liu2024a} follows a two-stage process: first identifying the most critical agents for a task, then reconfiguring communication pathways to amplify their influence. Similarly, graph-based orchestrators dynamically construct Directed Acyclic Graphs (DAGs) of tasks and dependencies, enabling parallel execution and flexible collaboration~\cite{jeyakumar2024advancing}. Additional research explores adaptive collaboration policies learned through teacher–student frameworks~\cite{li2021learning}.

\paragraph{Challenges in Scientific Collaboration Mechanism.}
While these general frameworks are foundational, scientific collaboration introduces distinctive challenges. First, it must be grounded in empirical reality: unlike general-purpose tasks, the aim is not merely to produce plausible text, but to generate testable hypotheses and verifiable experimental plans~\cite{boiko2023autonomous, szymanski2023autonomous}. The consensus among agents must be scientifically valid, avoiding pitfalls such as amplified hallucinations~\cite{hong2024metagpt} that could compromise results. Second, scientific inquiry requires managing epistemic diversity and uncertainty. Progress often arises from competing hypotheses; thus, agent collaboration should both foster a diversity of ideas and implement rigorous evaluation against evidence, quantify uncertainty, and prevent premature consensus that could hinder innovation~\cite{gottweis2025aicoscientist}. Finally, scientific discovery demands integrating complex, multimodal information across extended workflows. Agents must collectively reason over heterogeneous data and coordinate the use of specialized software and physical instruments~\cite{M.Bran2024, ramos2025review}. This necessitates communication protocols far more sophisticated than plain text exchange--requiring orchestration, static or dynamic, that can maintain state and share structured knowledge throughout the multi-stage process of scientific investigation.

\begin{table*}[t]
\centering
\begin{threeparttable}
\caption{Structured capability taxonomy of \textbf{optimization and evolution} in scientific agents. Rows are grouped into three major paradigms: iterative self-refinement, self-learning \& interaction, and population-based co-evolution.}
\label{tab:optimization}

\renewcommand{\arraystretch}{1.0}
\footnotesize
\setlength{\tabcolsep}{3pt}
\begin{tabularx}{\textwidth}{p{5.2cm} X X p{2.3cm}}

\toprule
\textbf{Paradigm} & \textbf{Purpose} & \textbf{Representative Mechanisms} & \textbf{Key References} \\
\midrule
\multicolumn{4}{l}{\textbf{I. Iterative self-refinement}} \\
\rowcolor[HTML]{F5F5F5}
\quad Self-feedback correction & Improve outputs by reflecting on own errors & SELF-REFINE; STaR; V-STaR (bootstrapped reasoning) & \cite{madaan2023self,wei2025alignrag,zelikman2024star,hosseini2024v} \\
\quad Tool-based feedback & Ground refinements via external validators & CRITIC; SelfEvolve (execution-based debugging) & \cite{goucritic,jiang2023selfevolve} \\
\rowcolor[HTML]{F5F5F5}
\quad Trial-and-error refinement & Incrementally improve through simulation or direct interaction & Iterative testing frameworks; simulated environments & \cite{song2024trial} \\

\midrule
\multicolumn{4}{l}{\textbf{II. Self-learning and interaction}} \\
\rowcolor[HTML]{F5F5F5}
\quad Model-level self-improvement & Enhance pretraining or tuning via self-supervision & SE; DiverseEvol (self-supervised learning) & \cite{zhong2023self,wu2023self} \\
\quad Self-reward reinforcement learning & Generate intrinsic rewards to guide policy evolution & Self-Rewarding LMs; RLCD; RLC & \cite{yuan2024selfrewardinglanguagemodels,yang2024rlcd,panglanguage} \\
\rowcolor[HTML]{F5F5F5}
\quad Knowledge-guided evolution & Integrate structured priors and external knowledge into planning & KnowAgent; WKM & \cite{zhu2024knowagent,qiao2024agent} \\

\midrule
\multicolumn{4}{l}{\textbf{III. Population-based co-evolution}} \\
\rowcolor[HTML]{F5F5F5}
\quad Cooperative evolution & Improve strategies via collaborative multi-agent interaction & CAMEL (role-playing); ProAgent; CORY (multi-agent RL) & \cite{li2023camel,zhang2024proagent,ma2024coevolving} \\
\quad Competitive evolution & Sharpen reasoning or robustness through adversarial settings & Multi-agent debate; Red-Teaming & \cite{du2023improving,liang2024encouraging,ma2023evolving} \\
\rowcolor[HTML]{F5F5F5}
\quad Mixed dynamics & Balance collaboration and competition for diverse improvement & Hybrid role-based or adversarial–synergistic frameworks & \cite{tran2025multi,guo2024large} \\

\bottomrule
\end{tabularx}
\end{threeparttable}
\end{table*}
\subsection{Optimization and Evolution}
\label{sec:optimization-evolution}

For a Scientific Agent, the ability to optimize and evolve is not merely about refining parameters but about enhancing the entire apparatus of scientific inquiry~\cite{gao2025survey,zhou2024symbolic,liang2024self}. Unlike agents in more constrained domains, a scientific agent's evolution must target its core scientific reasoning, its internal world model, and its collaborative structure to navigate the complexities of discovery. This process unfolds along several key axes: evolving the scientific strategy, the internal knowledge base, and the agent's own architecture (Table~\ref{tab:optimization}).

A primary mechanism for agent improvement is \textbf{iterative self-refinement}, where an agent enhances its outputs through a cycle of generation, feedback, and correction. This process can be entirely self-contained, as seen in methods like SELF-REFINE~\cite{madaan2023self,wei2025alignrag}, which uses self-generated feedback, or STaR~\cite{zelikman2024star} and V-STaR~\cite{hosseini2024v}, which bootstrap reasoning capabilities from a few examples. Agents can also leverage external tool-based feedback for more grounded correction, as demonstrated by CRITIC~\cite{goucritic}, which uses tools to validate and revise outputs, and SelfEvolve~\cite{jiang2023selfevolve}, which debugs code based on execution results. This trial-and-error process, whether simulated~\cite{song2024trial} or actual, allows agents to progressively reduce errors and improve the quality and reliability of their solutions through direct interaction with a task environment.

Beyond refining a single output, agents can evolve their underlying models and knowledge structures through \textbf{self-learning and interaction}. This includes self-supervised approaches that improve the core model, such as SE~\cite{zhong2023self} and DiverseEvol~\cite{wu2023self}, which enhance pre-training and instruction tuning, respectively. Reinforcement learning with self-generated rewards offers another path, where agents learn to produce their own reward signals to guide improvement, as seen in Self-Rewarding models~\cite{yuan2024selfrewardinglanguagemodels}, RLCD~\cite{yang2024rlcd}, and RLC~\cite{panglanguage}. Furthermore, agents can evolve by explicitly integrating external knowledge, which provides structured priors to guide planning and decision-making, as exemplified by KnowAgent~\cite{zhu2024knowagent} and WKM~\cite{qiao2024agent}. These methods focus on evolving the agent's intrinsic capabilities, leading to more robust and generalizable performance.

A third paradigm involves \textbf{population-based co-evolution}, where improvement emerges from the interactions within a group of agents. These interactions can be cooperative, where agents work together to solve problems. For instance, CAMEL~\cite{li2023camel} uses a role-playing framework for collaboration, ProAgent~\cite{zhang2024proagent} enables agents to infer teammates' intent for better coordination, and CORY~\cite{ma2024coevolving} uses multi-agent RL for fine-tuning. Conversely, evolution can be driven by competition. Multi-agent debate frameworks~\cite{du2023improving, liang2024encouraging} force agents to critique and defend positions, sharpening their reasoning. Similarly, adversarial setups like Red-Teaming~\cite{ma2023evolving} use competition to uncover and patch vulnerabilities. This co-evolutionary pressure, whether collaborative or competitive, drives the development of more sophisticated and resilient strategies across the agent population, mirroring evolutionary dynamics found in nature.

\paragraph{Challenges in Scientific Optimization and Evolution.} 
Applying these optimization and evolution techniques to scientific agents presents unique challenges not typically found in other domains. First, the evaluation of a scientific hypothesis or experiment is often resource-intensive, time-consuming, and expensive, making rapid, iterative feedback loops (central to many RL and self-correction methods) impractical. Unlike compiling code or checking a factual answer, a single evaluation may require days of lab work. Second, the reward landscape in scientific discovery is exceptionally sparse and complex; breakthroughs are rare, and the path to discovery often involves long periods with no positive feedback signal. This makes it difficult for agents to learn meaningful policies. Finally, the outputs of scientific agents must be grounded in physical reality and adhere to strict safety protocols. An "optimized" chemical synthesis procedure that is dangerously explosive is a catastrophic failure. Therefore, the optimization process must be constrained by scientific validity, safety, and the ultimate goal of producing reproducible and verifiable knowledge, adding layers of complexity beyond achieving high scores on a typical benchmark.

%% file: Sections/4.Process.tex
\begin{figure}[!t]
    \centering
    \includegraphics[width=1\linewidth]{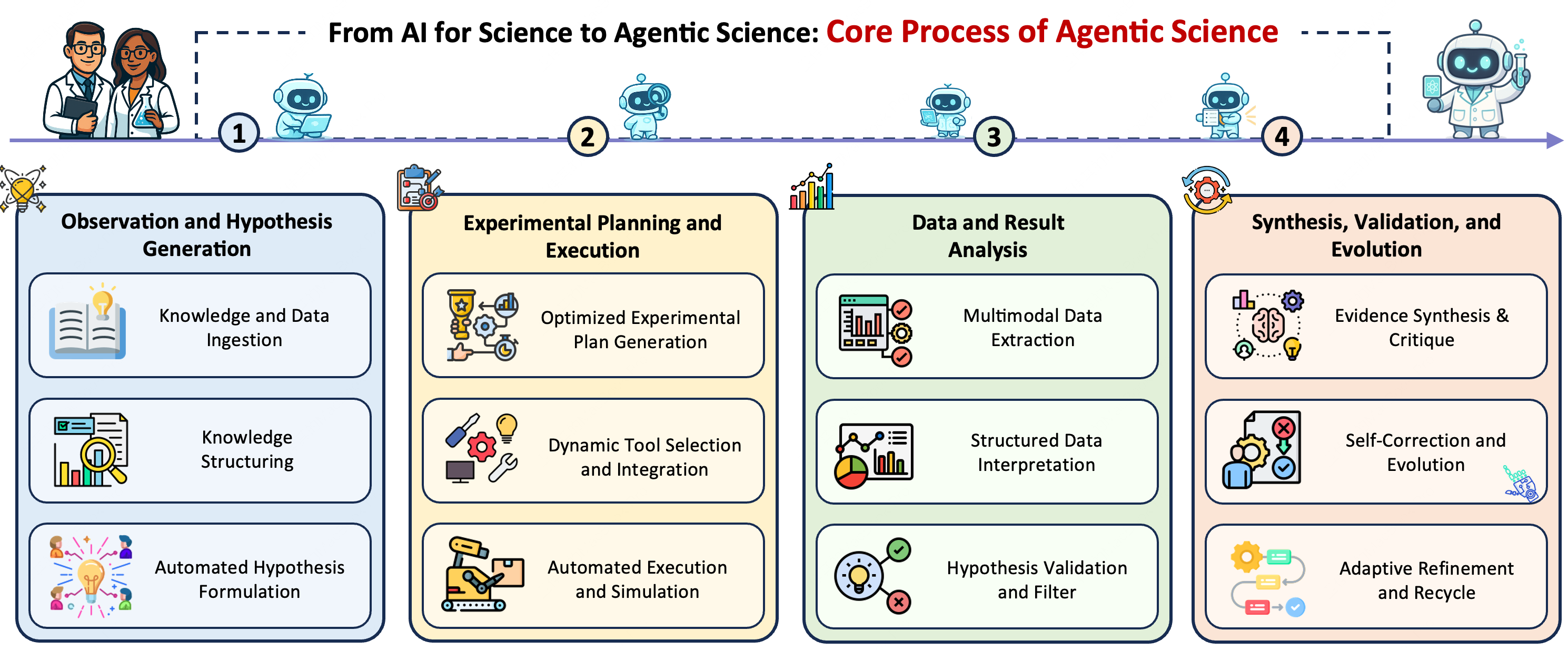}
    \caption{Core process of Agentic Science. Not all steps are required in every instance, and execution order may be \textbf{dynamically adjusted} based on agent objectives, context, and ongoing results.}
    \label{fig:process}
    \vspace{-0.em}
\end{figure}

\section{Agentic Science: Dynamic Workflow and Challenges}
\label{sec:process}

Agentic Science redefines the scientific method as an autonomous, closed-loop workflow, managed by intelligent agents. At its core, this paradigm contains a continual, self-improving cycle of discovery comprising four key stages: (1) \textbf{Observation and Hypothesis Generation}, (2) \textbf{Experimental Planning and Execution}, (3) \textbf{Result Analysis}, and (4) \textbf{Synthesis, Validation, and Evolution}. This section analyzes each stage by connecting it to the core agentic capabilities and challenges discussed previously, highlighting its implementation in current agentic systems. \textbf{Note}: Not all steps are required in every agentic system, and execution order may be dynamically adjusted based on agent objectives, context, and ongoing results.

\begin{table*}[!t]
\centering
\caption{The Agentic Science Loop: Mapping Core Processes to Agent Abilities and Scientific Challenges.}
\label{tab:unified_agentic_science}
\small
\renewcommand{\arraystretch}{1.1} 
\setlength{\tabcolsep}{4pt}      

\begin{tabularx}{\textwidth}{
    >{\RaggedRight\arraybackslash\scriptsize}p{2.6cm}
    >{\RaggedRight\arraybackslash\scriptsize}X
    >{\RaggedRight\arraybackslash\scriptsize}p{3cm}
    >{\RaggedRight\arraybackslash\scriptsize}X
}
\toprule
\textbf{Core Process in Agentic Science} & \textbf{Key Activities \& Representative Works} & \textbf{Primary Agent Abilities Utilized} & \textbf{Unique Scientific Challenges} \\
\midrule

\rowcolor[HTML]{EEF4FB}
\textbf{Observation \& Hypothesis Generation} &
\textbf{Knowledge Ingestion \& Structuring:} Synthesizing corpora via RAG (\textit{e.g.,} LitLLM toolkit~\cite{agarwal2024litllm}); organizing into knowledge graphs~\cite{edge2024graphrag} or taxonomies. \newline
\textbf{Hypothesis Formulation:} Reasoning over structured knowledge to identify novel, testable ideas (\textit{e.g.,} SciAgents~\cite{ghafarollahi2025sciagents}, Robin~\cite{ghareeb2025robin}, OriGene~\cite{zhang2025origene}).
&
\textbf{Memory Mechanism} (as a knowledge nexus) \newline
\textbf{Planning \& Reasoning Engines} (for exploratory pattern discovery)
&
\textbf{Vast Hypothesis Space:} Navigating an enormous and ill-defined space of possible scientific ideas. \newline
\textbf{Knowledge Veracity:} Contending with outdated or conflicting scientific knowledge. \newline
\textbf{Causal Discovery:} Aiming to generate hypotheses about causation, not just correlation. \\
\midrule

\rowcolor[HTML]{FFF0F2}
\textbf{Experimental Planning \& Execution} &
\textbf{Optimized Plan Generation:} Decomposing goals into structured, resource-efficient experimental workflows. \newline
\textbf{Automated Execution:} Controlling robotic hardware (\textit{e.g.,} Coscientist~\cite{boiko2023autonomous}, ORGANA~\cite{darvish2024organa}) or running simulations (\textit{e.g.,} The Virtual Lab~\cite{swanson2024virtual}). \newline
\textbf{Autonomous Coding:} Generating and executing analysis pipelines (\textit{e.g.,} CellAgent~\cite{xiao2024cellagent}, BIA~\cite{xin2024bioinformatics}).
&
\textbf{Planning \& Reasoning Engines} (for task decomposition and adaptation) \newline
\textbf{Tool Use \& Integration} (for real-world interaction and computation)
&
\textbf{Physical Plausibility \& Safety:} Ensuring plans are grounded in reality and adhere to safety protocols. \newline
\textbf{Strict Reproducibility:} Demanding meticulous provenance tracking of all parameters, code, and tool versions. \newline
\textbf{Cost \& Resource Management:} Balancing goals with real-world financial and computational budgets. \\
\midrule

\rowcolor[HTML]{F1FAF0}
\textbf{Data \& Result Analysis} &
\textbf{Multimodal Data Extraction:} Parsing semantic content from charts~\cite{masry2022chartqabenchmarkquestionanswering}, tables~\cite{wang2024chainoftableevolvingtablesreasoning}, and other outputs. \newline
\textbf{Structured Interpretation:} Interleaving reasoning and action to connect results to hypotheses~\cite{yao2023react}. \newline
\textbf{Insight Generation:} Uncovering mechanistic explanations from raw data (\textit{e.g.,} PROTEUS~\cite{ding2024automating}, SpatialAgent~\cite{wang2025spatialagent}).
&
\textbf{Tool Use \& Integration} (to parse experimental data) \newline
\textbf{Planning \& Reasoning Engines} (to interpret outcomes) \newline
\textbf{Memory Mechanism} (to contextualize new findings)
&
\textbf{Noisy \& Ambiguous Feedback:} Scientific results are often incomplete or require expert interpretation. \newline
\textbf{Heterogeneous Data Integration:} Seamlessly reasoning across diverse data types (text, images, spectra, sequences). \newline
\textbf{Avoiding Confirmation Bias:} Objectively evaluating results, especially those that contradict the hypothesis. \\
\midrule

\rowcolor[HTML]{FFFCE5}
\textbf{Synthesis, Validation, \& Evolution} &
\textbf{Evidence Synthesis \& Critique:} Emulating peer review via multi-agent debate to validate claims~\cite{ou2025claimcheckgroundedllmcritiques}. \newline
\textbf{Automated Reproducibility:} Verifying findings through automated replication checks. \newline
\textbf{Adaptive Refinement:} Learning from past experiments to improve future strategy (\textit{e.g.,} Reflexion~\cite{shinn2023reflexion}, Sparks~\cite{ghafarollahi2025sparks}, MOOSE-Chem3~\cite{liu2025moose}).
&
\textbf{Collaboration between Agents} (for peer review and critique) \newline
\textbf{Optimization \& Evolution} (for self-improvement) \newline
\textbf{Memory Mechanism} (for long-term learning)
&
\textbf{Long-Term Causal History:} Maintaining a coherent, causally-linked record of a long-term research project. \newline
\textbf{Expensive \& Sparse Rewards:} Scientific breakthroughs are rare, providing infrequent signals for learning algorithms. \newline
\textbf{Sustained, Productive Improvement:} Ensuring agent "evolution" is scientifically valid and not just reinforcing biases. \\
\bottomrule

\end{tabularx}
\end{table*}

\subsection{Observation and Hypothesis Generation}

The initiation of agentic inquiry centers on the formulation of novel, testable hypotheses derived from prior knowledge. This process relies fundamentally on the agent's \textbf{memory mechanism}, particularly its ability to function as a knowledge connector. 

Agents begin with \textbf{knowledge ingestion}, using techniques like Retrieval-Augmented Generation (RAG)~\cite{lewis2020retrieval} to query and synthesize vast scientific corpora, as demonstrated in systems like the LitLLM toolkit~\cite{agarwal2024litllm} and ResearchAgent~\cite{baek2025researchagentiterativeresearchidea}. This information is then organized via \textbf{knowledge structuring} into formats like taxonomies or knowledge graphs~\cite{gao2025sciencehierarchographyhierarchicalorganization, newman2024arxivdigestablessynthesizingscientificliterature, edge2024graphrag} to ground subsequent reasoning. Building on this structured knowledge, the agent's \textbf{planning and reasoning engine} engages in \textbf{hypothesis formulation}~\cite{si2024can, hu2024nova, weng2025cycleresearcher}. This can be formally represented as the maximization of a potential function $P$ over a set of candidate hypotheses $H_{\text{cand}}$, conditioned on a structured memory $M$ derived from the knowledge base $K$:
\begin{equation}
h_{\text{new}} = \arg\max_{h \in H_{\text{cand}}} P(h | M(K))
\end{equation}
This is not merely a linear deduction but often an exploratory process of pattern discovery and symbolic reasoning to identify promising research directions~\cite{baek2024researchagent,lu2024ai,pu2025piflow,yamada2025ai,su-etal-2025-many,pauloski2025empowering,liu2024aigs, yang2024large, gottweis2025towards}. Systems like SciAgents~\cite{ghafarollahi2025sciagents} and MOOSE-Chem~\cite{yang2024moose} exemplify this by reasoning over structure-property relationships and chemical reactivity, respectively.

This stage faces significant challenges unique to the scientific domain: heterogeneous data formats, dynamic knowledge updating, and large search space. The primary challenge lies in the nature of scientific knowledge itself: its veracity can decay over time, and it is highly heterogeneous and multi-modal. An agent's \textbf{memory system} must therefore not only ingest data but also grapple with potentially outdated facts and seamlessly reason across text, tables, and images. Furthermore, the \textbf{reasoning engine} must navigate a vast, unstructured search space of possible hypotheses, requiring sophisticated strategies to balance exploration and exploitation~\cite{todorov2012mujoco}. The ultimate goal is to formulate hypotheses that aim for causal understanding~\cite{sauter2023meta}, a far more complex task than correlational pattern matching common in general domains.

Empirical results underscore the potential of this agentic formulation. \textbf{OriGene}~\cite{zhang2025origene}, a virtual disease biologist, integrates multimodal data to generate and prioritize therapeutic targets. It identified GPR160 and ARG2 as novel candidates for liver and colorectal cancer, respectively--both of which were subsequently validated in patient-derived systems. Similarly, \textbf{Robin}~\cite{ghareeb2025robin}, a collaborative multi-agent system, autonomously hypothesized the use of ripasudil for treating dry age-related macular degeneration (dAMD)--a drug previously unlinked to the condition--by autonomously conducting background research and inference. In another domain, \textbf{CellVoyager}~\cite{alber2025cellvoyager} exemplifies data-driven hypothesis generation by reanalyzing aging-related transcriptomic datasets. It uncovered a previously unreported link between increased transcriptional noise and brain aging, demonstrating the capacity of agentic systems to surface latent biological insights.

\subsection{Experimental Planning and Execution}

The second phase of Agentic Science operationalizes hypotheses through end-to-end experimental workflows. This stage is managed by the agent's \textbf{planning and reasoning engine}, which performs \textbf{optimized plan generation}. This involves decomposing a high-level goal into a structured, resource-efficient plan, which could be a biological protocol~\cite{odonoghue2023bioplannerautomaticevaluationllms} or an algorithm for causal discovery~\cite{li2025largelanguagemodelshelp}. 

This process can be modeled as a constrained optimization problem, where the agent seeks to find an experimental plan $\pi^*$ that minimizes cost $C(\pi)$ while ensuring the plan's validity $V(\pi, h)$ for testing hypothesis $h$ exceeds a certain threshold $\theta$:
\begin{equation}
\pi^* = \arg\min_{\pi \in \Pi} C(\pi) \quad \text{s.t.} \quad V(\pi, h) \geq \theta
\end{equation}
The execution of this plan, yielding results $R$, depends on the agent's \textbf{tool use and integration} capability, denoted by an execution function that leverages a set of available tools $T$: $R = \text{Execute}(\pi^*, T)$. The agent must perform \textbf{dynamic tool selection}, mapping abstract plan steps to concrete tool invocations, and then engage in \textbf{automated execution} by generating code or controlling robotic hardware. This capability is seen in systems that autonomously generate research code~\cite{jansen2025codescientist,novikov2025alphaevolve,schmidgall2025agent,gandhi2025researchcodeagent}, as evaluated by benchmarks like SciCode~\cite{tian2024scicoderesearchcodingbenchmark} and MLE-Bench~\cite{chan2025mlebenchevaluatingmachinelearning}. To enhance reliability, especially in complex tasks, agents can employ advanced planning strategies like tree search~\cite{jiang2025aideaidrivenexplorationspace} to explore and backtrack from potential execution paths.

Executing scientific experiments introduces formidable challenges that stress agentic capabilities. Scientific planning operates under a paradigm of \textbf{high-stakes and strict verifiability}, where a flawed plan can lead to wasted resources or invalid conclusions. This demands exceptional reliability from the reasoning engine. The \textbf{tool use} itself requires an extremely high degree of precision and domain understanding, as minor errors in parameterizing a simulation or a lab instrument can invalidate results. Moreover, \textbf{reproducibility and provenance} are non-negotiable; the agent must meticulously log all tool versions and parameters to ensure its work can be verified. This is further complicated by the need to create complex workflows by chaining multiple specialized tools, a task known to be difficult~\cite{haiyang2025shortcutsbench}. Finally, because many scientific tools (e.g., high-fidelity simulators, lab equipment) are expensive, the agent must perform sophisticated \textbf{cost-benefit analysis}, a challenge rarely faced by general-purpose agents.

Agentic systems increasingly demonstrate proficiency in closed-loop planning and execution across both virtual and physical domains. For instance, \textbf{Coscientist}~\cite{boiko2023autonomous} autonomously designed and optimized a palladium-catalyzed cross-coupling reaction by interfacing with robotic hardware, showcasing an end-to-end experimental loop. Similarly, the robotic agent \textbf{ORGANA}~\cite{darvish2024organa} executed a 19-step synthesis and characterization protocol for quinone derivatives, reducing human workload by over 80\%. In virtual labs, \textbf{The Virtual Lab}~\cite{swanson2024virtual} autonomously constructed a computational pipeline incorporating AlphaFold and docking simulations to design 92 novel SARS-CoV-2 nanobodies, two of which demonstrated strong binding in subsequent empirical tests. In bioinformatics, agents such as \textbf{BIA}~\cite{xin2024bioinformatics} and \textbf{CellAgent}~\cite{xiao2024cellagent} have demonstrated robust pipeline planning and execution for tasks like single-cell RNA-seq analysis.

\subsection{Data and Result Analysis}

Following experiment execution, the agent must extract actionable insights from raw outputs to update its belief about the hypothesis. This phase relies on a tight integration of \textbf{tool use}, \textbf{reasoning}, and \textbf{memory}. 

The process begins with \textbf{multimodal data extraction}~\cite{xie2024large}, using specialized tools or vision-language models to parse semantic content from outputs like scientific charts~\cite{masry2022chartqabenchmarkquestionanswering, wang2024charxivchartinggapsrealistic}. Subsequently, the agent's \textbf{reasoning engine} performs \textbf{structured interpretation}, employing techniques like Chain-of-Table to understand complex relational data~\cite{wang2024chainoftableevolvingtablesreasoning}. This entire analysis is a practical application of the ReAct~\cite{yao2023react} framework, where the agent observes the experimental outcome and reasons about its implications. This can be conceptualized as a Bayesian update to the agent's belief in the hypothesis $h$, where the posterior probability $P(h|R)$ is proportional to the likelihood of observing the results $R$ given the hypothesis, $P(R|h)$, multiplied by the prior belief $P(h)$:
\begin{equation}
P(h|R) \propto P(R|h) \cdot P(h)
\end{equation}
This reasoning is contextualized by the agent's \textbf{memory}, which holds the prior experimental history and domain knowledge necessary for accurate interpretation and \textbf{hypothesis validation}. Agents may even generate scientific figures to communicate their findings~\cite{belouadi2024automatikztextguidedsynthesisscientific, zadeh2025text2chart31instructiontuningchart}.

The primary challenge in this stage stems from the nature of scientific feedback loops, which often involve \textbf{noisy, multimodal experimental data}. An agent's reasoning engine must be robust enough to correctly interpret this data, distinguishing signal from noise without succumbing to confirmation bias. This is compounded by the \textbf{heterogeneous data types} involved; an agent's memory and reasoning architecture must seamlessly handle a mix of text, tables, genomic sequences, and imagery to form a coherent conclusion. Unlike general tasks where feedback is often a clear text-based signal, scientific analysis demands a deep, contextual understanding of complex and often ambiguous data formats.

Agentic systems have demonstrated increasing autonomy and sophistication in scientific interpretation. For example, after generating a therapeutic hypothesis and proposing an RNA-seq experiment, \textbf{Robin}~\cite{ghareeb2025robin} autonomously analyzed the resulting data to uncover the increase in expression of \textit{ABCA1}, a lipid efflux regulator, as a potential mechanism of action. In proteomics, \textbf{PROTEUS}~\cite{ding2024automating} performs end-to-end analysis of raw mass spectrometry data, generating mechanistic hypotheses judged by human experts to be both valid and insightful. \textbf{SpatialAgent}~\cite{wang2025spatialagent} achieved expert-level performance on spatial biology datasets comprising over two million single-cell measurements. Beyond biology, \textbf{LLM-RDF}~\cite{ruan2024accelerated} integrates specialized analytical agents--including a Spectrum Analyzer and a Result Interpreter--that process experimental feedback to directly inform the next stages of chemical synthesis.

\subsection{Synthesis, Validation, and Evolution}

The final stage of the agentic scientific loop involves synthesizing outcomes, validating hypotheses, and refining future lines of inquiry. This process heavily leverages \textbf{collaboration between agents} and advanced \textbf{memory mechanisms}. 

To ensure robustness, agents can engage in \textbf{evidence synthesis and critique}, emulating peer review by assessing the plausibility of claims~\cite{ou2025claimcheckgroundedllmcritiques}. This is often implemented in deliberative multi-agent systems where agents challenge and refine each other's conclusions~\cite{wu2023autogen, du2023improving}. \textbf{Automated validation} further strengthens findings through reproducibility checks~\cite{takagi2023autonomoushypothesisverificationlanguage, xiang2025scireplicatebenchbenchmarkingllmsagentdriven}. Crucially, the agent undergoes \textbf{adaptive refinement}, where it evolves its strategy based on cumulative experience. This relies on memory frameworks like Reflexion~\cite{shinn2023reflexion}, where agents learn from a repository of past successes and failures. This evolution can be described as updating the agent's internal policy $\phi$ based on a learning function $\mathcal{L}$ applied to its memory $M$ of past trajectories (hypothesis, plan, result tuples):
\begin{equation}
\phi_{t+1} \leftarrow \mathcal{L}(\phi_t, M_t)
\end{equation}
The agent's \textbf{planning engine} can then use this refined policy to guide long-term strategy, for instance by using MCTS to optimize hypothesis selection over an entire research campaign~\cite{rabby2025iterativehypothesisgenerationscientific} or applying formal verification to refine its internal logic~\cite{quan2024verificationrefinementnaturallanguage}.

This final stage faces the most profound long-term challenges. The core difficulty is enabling \textbf{long-term causal reasoning}, as scientific insights can emerge from connecting experiments conducted months or even years apart. Existing \textbf{memory systems} are ill-equipped to maintain such extended, causally-linked histories with the high fidelity required for ensuring the reproducibility and integrity of discoveries. This is the ultimate test of an agentic system: not just executing a single loop, but learning and improving over many loops to conduct a long-horizon research project. Successfully managing this iterative process of self-correction and knowledge accumulation is the key to transforming agents from single-task tools into true partners in sustained scientific discovery.

Agentic systems have begun to demonstrate these abilities. The \textbf{Sparks} framework~\cite{ghafarollahi2025sparks}, for instance, integrates generation-and-reflection agents to autonomously discover two novel protein design rules via iterative self-correction. \textbf{OriGene}~\cite{zhang2025origene} embeds a self-evolving architecture that assimilates experimental and human feedback to progressively refine its disease-targeting protocols. In single-cell data analysis, \textbf{CellAgent}~\cite{xiao2024cellagent} employs a recursive evaluator-planner loop that critiques and improves analysis pipelines, yielding expert-level interpretations. Targeted discovery optimization is also realized in \textbf{MOOSE-Chem3}~\cite{liu2025moose}, which proposes an experiment-guided candidate ranking strategy. By learning from past hypothesis performance, the system adaptively prioritizes the most promising next experiments--closing the loop between evaluation and exploration.

\begin{table*}[!t]
\centering
\caption{Paradigms of Fully Autonomous Research Pipelines. \textbf{Note that we only report the most significant features of each paper}.}
\label{tab:pipeline_paradigms}
\small
\renewcommand{\arraystretch}{1.2} 
\setlength{\tabcolsep}{5pt}      
\begin{tabularx}{\textwidth}{
    >{\RaggedRight\arraybackslash\scriptsize\scriptsize}p{3.5cm}
    >{\RaggedRight\arraybackslash\scriptsize}X
    >{\RaggedRight\arraybackslash\scriptsize}X
}
\toprule
\textbf{Pipeline Paradigm} & \textbf{Core Contribution \& Mechanism} & \textbf{Representative Systems \& Works} \\
\midrule

\rowcolor[HTML]{EBF5FB}
\textbf{Foundational End-to-End Frameworks} & 
Establishes the viability of a complete, closed-loop research cycle. These systems integrate hypothesis generation, coding, experimentation (often virtual), and reporting into a single, cohesive workflow. &
The AI Scientist~\cite{lu2024ai}, NovelSeek~\cite{team2025novelseek}, Dolphin~\cite{yuan2025dolphin}, X-Master~\cite{chai2025scimaster}, DiscoveryWorld (evaluation environment)~\cite{jansen2024discoveryworld} \\
\midrule

\rowcolor[HTML]{FEF9E7}
\textbf{Domain-Specific Automation} & 
Applies the end-to-end paradigm to specialized, high-impact scientific domains. This often involves interfacing with real-world lab robotics, complex simulators, or highly structured domain-specific data formats. &
Coscientist~\cite{boiko2023autonomous}, LLM-RDF~\cite{ruan2024automatic}, MatPilot~\cite{ni2024matpilot}, Biomni~\cite{huang2025biomni}, SpatialAgent~\cite{wang2025spatialagent}, PROTEUS~\cite{ding2024automating},
OriGene~\cite{zhang2025origene},
The Virtual Lab~\cite{swanson2024virtual}, AI co-scientist~\cite{gottweis2025towards} \\
\midrule

\rowcolor[HTML]{F4ECF7}
\textbf{Multi-Agent Collaborative Structures} & 
Emulates the collaborative and adversarial nature of scientific inquiry using teams of agents. These systems explore different organizational structures (e.g., Socratic dialogue, hierarchical teams, peer review) to enhance creativity and rigor. &
VirSci~\cite{su-etal-2025-many}, MAPS~\cite{zhang2025maps}, DORA~\cite{naumov2025dora}, MDAgents~\cite{kim2024mdagents}, AgentRxiv (cross-system collaboration)~\cite{schmidgall2025agentrxiv} \\
\midrule

\rowcolor[HTML]{E9F7EF}
\textbf{Self-Evolving \& Adaptive Systems} & 
Focuses on the pipeline's ability to learn and improve over time. These agents autonomously refine their strategies, expand their toolkits, or update their internal knowledge based on cumulative experience and feedback. &
STELLA~\cite{jin2025stella}, Agent Hospital~\cite{li2024agent}, ResearchAgent~\cite{baek2024researchagent}, 
OriGene~\cite{zhang2025origene},
AlphaEvolve~\cite{novikov2025alphaevolve} \\
\midrule

\rowcolor[HTML]{FDEDEC}
\textbf{Human-in-the-Loop Integration} & 
Explicitly designs the pipeline to incorporate human expertise and oversight. These frameworks treat the human researcher as a collaborator, leveraging their feedback to guide the autonomous process and ensure alignment with scientific goals. &
Agent Laboratory~\cite{schmidgall2025agent}, Conversational Health Agents~\cite{abbasian2023conversational}, MatPilot~\cite{ni2024matpilot} \\
\bottomrule

\end{tabularx}
\end{table*}

\subsection{Fully Autonomous Research Pipeline}
\label{subsec:pipeline}

An emerging frontier in Agentic Science is the development of frameworks that automate the entire scientific research pipeline, from idea generation to discovery and reporting. These systems aim to construct a productive cycle of hypothesis, experimentation, and analysis, effectively creating an autonomous or semi-autonomous researcher (Table~\ref{tab:pipeline_paradigms}). 

Early frameworks such as \textbf{The AI Scientist}~\cite{lu2024ai} and \textbf{NovelSeek}~\cite{team2025novelseek} established this paradigm by proposing comprehensive, closed-loop systems capable of performing research across multiple domains. \textbf{The AI Scientist} demonstrated a fully automated workflow that generates ideas, writes and executes code, and drafts a full scientific paper, applying it to subfields within machine learning. Similarly, \textbf{NovelSeek} showcased a unified multi-agent framework that achieved performance gains in tasks like reaction yield and enhancer activity prediction. Other systems like \textbf{Dolphin}~\cite{yuan2025dolphin} emphasize a feedback-driven loop where ideas are refined based on prior experimental results and literature analysis, demonstrating continuous performance improvement on tasks such as 3D point classification. These foundational efforts established the viability of end-to-end agentic research pipelines.

Building on this general paradigm, subsequent work has specialized these pipelines for high-impact scientific domains, often integrating with real-world laboratory hardware or complex simulation tools. In chemistry, \textbf{Coscientist}~\cite{boiko2023autonomous} demonstrated a landmark achievement by using a GPT-4-powered agent to autonomously design, plan, and execute a palladium-catalyzed cross-coupling reaction in a physical lab. This was supported by other systems like \textbf{LLM-RDF}~\cite{ruan2024automatic}, a multi-agent framework with specialized agents for literature scouting, experiment design, and result interpretation to automate chemical synthesis development. This approach was also extended to materials science with \textbf{MatPilot}~\cite{ni2024matpilot}, which uses a human-machine collaborative framework for materials discovery. In biomedicine, \textbf{Biomni}~\cite{huang2025biomni} acts as a general-purpose agent that autonomously builds its own action space by mining tools and protocols from publications, achieving strong generalization across tasks like drug repurposing and molecular cloning. More specialized agents like \textbf{SpatialAgent}~\cite{wang2025spatialagent} and \textbf{PROTEUS}~\cite{ding2024automating} have achieved expert-level performance in complex fields like spatial biology and proteomics, respectively. The feasibility of virtual research teams was shown by \textbf{The Virtual Lab}~\cite{swanson2024virtual}, where a team of LLM agents designed novel SARS-CoV-2 nanobodies that were later experimentally validated. Similarly, an \textbf{AI co-scientist}~\cite{gottweis2025towards} proposed and validated novel epigenetic targets for liver fibrosis. The scope of agentic pipelines extends even to pure mathematics and computer science, with \textbf{ToRA}~\cite{gou2023tora} integrating symbolic solvers for mathematical reasoning and \textbf{AlphaEvolve}~\cite{novikov2025alphaevolve} using an evolutionary coding agent to discover novel, provably correct algorithms, including an improvement over Strassen's matrix multiplication.

A key structure in these general pipelines is the use of multi-agent systems to emulate the collaborative nature of scientific research. The core insight, demonstrated by systems like \textbf{VirSci}~\cite{su-etal-2025-many}, is that a team of collaborative agents can generate more innovative and impactful scientific ideas than a single agent. These systems explore diverse collaboration structures. For example, \textbf{MAPS}~\cite{zhang2025maps} employs a team of seven agents inspired by personality traits and Socratic dialogue to solve multimodal scientific problems. \textbf{DORA}~\cite{naumov2025dora} utilizes hierarchical teams of generalist and specialist agents to automate the generation of research reports. In the medical domain, \textbf{MDAgents}~\cite{kim2024mdagents} dynamically adapts the collaboration structure--assigning tasks to solo or group agents--based on the complexity of the medical decision, leading to improved performance on clinical diagnosis benchmarks. Extending collaboration beyond a single system, \textbf{AgentRxiv}~\cite{schmidgall2025agentrxiv} introduces a novel framework where multiple agent "laboratories" upload and retrieve research from a shared preprint server, enabling them to iteratively build on each other's work and achieve faster progress than isolated systems.

The long-term success of these pipelines depends on their ability to learn, evolve, and effectively integrate human expertise. Self-evolution is a central theme in systems like \textbf{STELLA}~\cite{jin2025stella}, a biomedical agent that autonomously improves its own performance by dynamically expanding its library of tools and reasoning templates. This enables its accuracy on benchmarks to nearly double with increased operational experience. Similarly, \textbf{Agent Hospital}~\cite{li2024agent} introduces a medical simulation where doctor agents evolve and improve their diagnostic capabilities by treating tens of thousands of simulated patients. Iterative refinement through agent-based peer review is another powerful mechanism, as seen in \textbf{ResearchAgent}~\cite{baek2024researchagent}, which uses a panel of reviewing agents to provide feedback and progressively enhance research ideas generated from scientific literature. Recognizing the value of human oversight, frameworks like \textbf{Agent Laboratory}~\cite{schmidgall2025agent} and \textbf{Conversational Health Agents}~\cite{abbasian2023conversational} are explicitly designed to incorporate human feedback at various stages, from idea generation to final report generation, ensuring that the autonomous process remains aligned with researcher goals and significantly improving research quality while reducing costs.

Underpinning these complex research pipelines are foundational agent capabilities and the critical need for robust evaluation methods. The ability to perform complex, tool-augmented reasoning is a prerequisite for any scientific agent. Systems like \textbf{X-Master}~\cite{chai2025scimaster} are designed to validate this core competence, achieving state-of-the-art performance on exceedingly difficult benchmarks like Humanity's Last Exam by emulating how human researchers flexibly interact with tools. A crucial upstream capability is open-domain hypothesis discovery, where agents must generate novel and valid scientific hypotheses directly from unstructured data like raw web corpora, a challenge tackled in~\cite{yang2023large}. Given the complexity of these end-to-end systems, evaluating their capacity for genuine scientific discovery is a major challenge. To address this, specialized evaluation environments are being developed. \textbf{DiscoveryWorld}~\cite{jansen2024discoveryworld} is a virtual environment that provides a suite of simulated, multi-modal scientific tasks, enabling the benchmarking of an agent's ability to complete a full discovery cycle in a controlled and repeatable setting.

%% file: Sections/5.Life.tex
\begin{figure}[!t]
    \centering
    \includegraphics[width=1\textwidth]{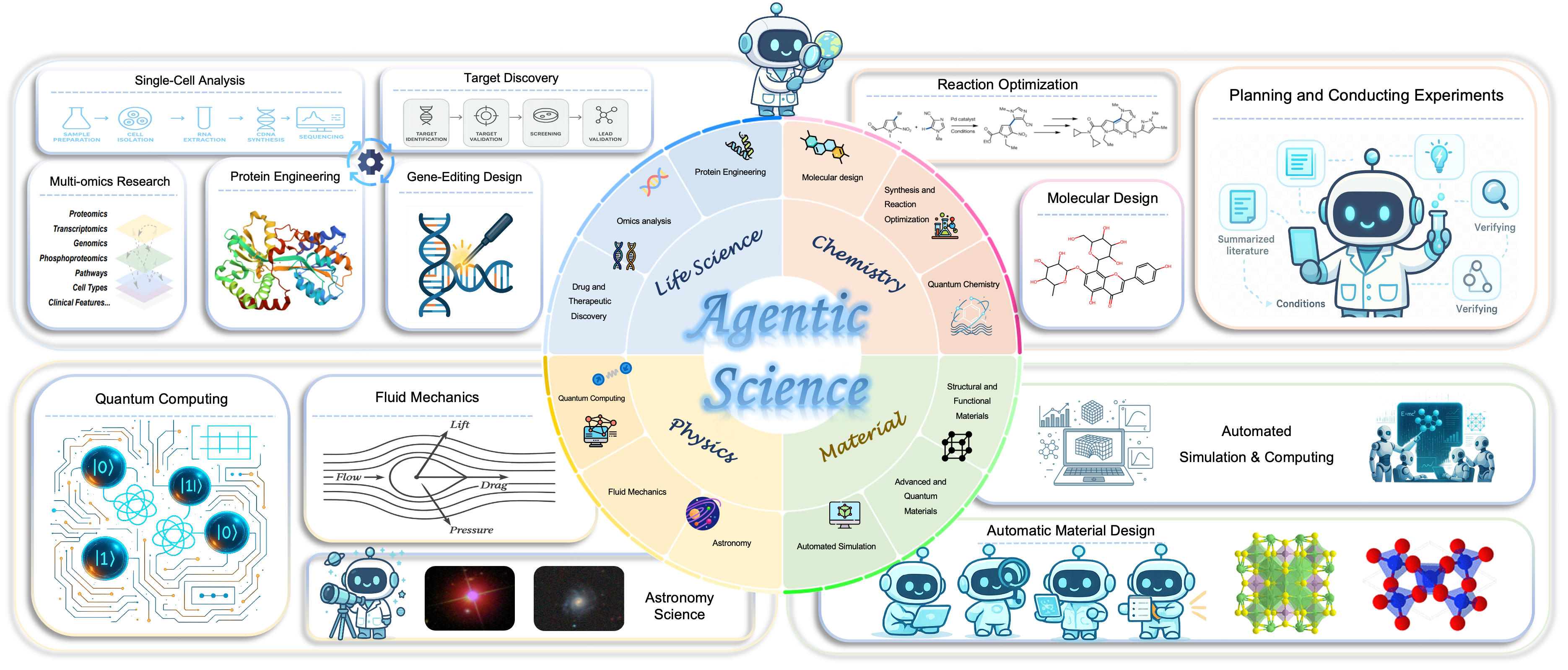}
    \caption{Agentic AI-based Natural Scientific Research. Note that only representative tasks are shown in the figure.}
    \label{fig:taxonomy}
\end{figure}

\section{Agentic Life Sciences Research}
\label{sec:life}

The application of agentic AI systems is rapidly transforming life sciences research, a domain characterized by vast, complex datasets and intricate, multi-step experimental workflows. From genomics and proteomics to drug discovery and protein engineering, AI agents are being developed to automate data analysis, generate novel hypotheses, design experiments, and even interpret results, thereby accelerating the pace of discovery. These systems typically employ a multi-agent architecture, where specialized agents (e.g., planner, executor, analyst) collaborate to tackle complex problems that traditionally require significant human expertise and labor. This section surveys the emerging landscape of agentic systems in life sciences, categorized by their primary application domain (Table~\ref{tab:life} and Table~\ref{tab:conclusion_life}).

\begin{table*}[!t]
   \centering
   \small 
    \caption{Classification of Agentic Systems in Life Sciences, organized to correspond with the survey text. Column Key: \textbf{Hypo.}: Observation or Hypothesis Generation, \textbf{Exper.}: Experimental Planning or Execution, \textbf{Analysis}: Data and Result Analysis, \textbf{Validation}: Synthesis, Validation, and Evolution. {\leveltwoicon} means level 2 and {\levelthreeicon} means level 3.} 
   \label{tab:life}
   
   \renewcommand{\arraystretch}{0.9}
   \setlength{\tabcolsep}{1.3pt}
   
   \begin{tabularx}{\textwidth}{l >{\raggedright\arraybackslash}X ccccc}
   \toprule
   & & \multicolumn{5}{c}{\textbf{Core Process}} \\
   \cmidrule(lr){3-6}
   \textbf{Paper} & \textbf{Application Domain} &
   \textbf{Hypo.} & \textbf{Exper.} & \textbf{Analysis} & \textbf{Validation} & \textbf{Level} \\
   \midrule

   \multicolumn{7}{c}{\textit{General Biomedical Research Frameworks}} \\
   \rowcolor[HTML]{DAE8FC}
   Biomni \cite{huang2025biomni} & General Biomedical Tasks & & \checkmark & \checkmark & & \leveltwoicon \\
   STELLA \cite{jin2025stella} & Self-Evolving Research & & \checkmark &  \checkmark & \checkmark & \leveltwoicon \\
   \rowcolor[HTML]{DAE8FC}
   BioResearcher \cite{luo2025intention} & End-to-End Dry Lab Research & & \checkmark & \checkmark & \checkmark & \levelthreeicon \\
   PiFlow \cite{pu2025piflow} & Principled Scientific Discovery & \checkmark & \checkmark & \checkmark & & \levelthreeicon \\
   \rowcolor[HTML]{DAE8FC}
   Empowering BD \cite{gao2024empowering} & Perspective on AI Scientists & - & - & - & - & - \\
   Healthflow \cite{zhu2025healthflow} & Autonomous Healthcare Research & \checkmark & \checkmark & \checkmark & \checkmark & \levelthreeicon \\
   \midrule
   
   \multicolumn{7}{c}{\textit{Genomics, Transcriptomics, and Multi-Omics Analysis}} \\
   \rowcolor[HTML]{DAE8FC}
   BIA \cite{xin2024bioinformatics} & Bioinformatics Workflow & & \checkmark & \checkmark & \checkmark & \leveltwoicon \\
   CellAgent \cite{xiao2024cellagent} & scRNA-seq Analysis & & \checkmark & \checkmark & \checkmark & \leveltwoicon \\
   \rowcolor[HTML]{DAE8FC}
   TAIS \cite{liu2024toward} & Gene Expression Analysis & & \checkmark & \checkmark & & \leveltwoicon \\
   CRISPR-GPT \cite{huang2024crispr} & Gene-Editing Design & & \checkmark & & & \leveltwoicon \\
   \rowcolor[HTML]{DAE8FC}
   SpatialAgent \cite{wang2025spatialagent} & Spatial Biology & \checkmark & \checkmark & \checkmark & \checkmark & \levelthreeicon \\
   PhenoGraph \cite{niyakan2025phenograph} & Spatial Transcriptomics & & \checkmark & \checkmark & & \leveltwoicon \\
   \rowcolor[HTML]{DAE8FC}
   BioAgents \cite{mehandru2025bioagents} & Bioinformatics Analysis & & & \checkmark & & \leveltwoicon \\
   BioMaster \cite{su2025biomaster} & Bioinformatics Workflow & & \checkmark & \checkmark & \checkmark & \leveltwoicon \\
   \rowcolor[HTML]{DAE8FC}
   TransAgent \cite{zhang2025transagent} & Transcriptional Regulation & & \checkmark & \checkmark & & \leveltwoicon \\
   CompBioAgent \cite{zhang2025compbioagent} & scRNA-seq Exploration & & & \checkmark & & \leveltwoicon \\
   \rowcolor[HTML]{DAE8FC}
   PerTurboAgent \cite{hao2025perturboagent} & Perturb-seq Design & & \checkmark & \checkmark &  \checkmark & \leveltwoicon \\
   PROTEUS \cite{ding2024automating, qu2025automating} & Proteomics/Multi-Omics & \checkmark & \checkmark & \checkmark & \checkmark & \levelthreeicon \\
   \rowcolor[HTML]{DAE8FC}
   CellVoyager \cite{alber2025cellvoyager} & scRNA-seq Discovery & \checkmark &  \checkmark & \checkmark &  \checkmark & \levelthreeicon \\
   AstroAgents \cite{saeedi2025astroagents} & Mass Spectrometry Analysis & \checkmark & \checkmark & \checkmark & \checkmark & \levelthreeicon \\
   \rowcolor[HTML]{DAE8FC}
   BioDiscoveryAgent \cite{roohani2024biodiscoveryagent} & Perturbation Experiment Design & \checkmark & \checkmark & & & \levelthreeicon \\
   OmniCellAgent \cite{Huang2025} & scRNA-seq Data-driven Biomedical Research & \checkmark & \checkmark & \checkmark & \checkmark & \levelthreeicon \\
   \rowcolor[HTML]{DAE8FC}
   GeneAgent \cite{wang2024geneagent} & Gene Set Knowledge Discovery & \checkmark & \checkmark & \checkmark & \checkmark & \levelthreeicon \\
   PrimeGen \cite{wang2025accelerating} & Primer Design & & \checkmark & \checkmark & & \leveltwoicon \\
   \midrule
   
   \multicolumn{7}{c}{\textit{Protein Science and Engineering}} \\
   \rowcolor[HTML]{DAE8FC}
   ProtAgents \cite{ghafarollahi2024protagents} & \textit{De Novo} Protein Design & & \checkmark & \checkmark & & \leveltwoicon \\
   Sparks \cite{ghafarollahi2025sparks} & Protein Principle Discovery & \checkmark & \checkmark & \checkmark & \checkmark & \levelthreeicon \\
   \midrule

   \multicolumn{7}{c}{\textit{Drug and Therapeutic Discovery}} \\
   \rowcolor[HTML]{DAE8FC}
   The Virtual Lab \cite{swanson2024virtual} & Nanobody Design & \checkmark & \checkmark & \checkmark & \checkmark & \leveltwoicon \\
   OriGene \cite{zhang2025origene} & Therapeutic Target Discovery & \checkmark & \checkmark & \checkmark & \checkmark & \levelthreeicon \\
   \rowcolor[HTML]{DAE8FC}
   LLM Agent for DD \cite{ock2025large} & Drug Discovery Pipeline &  &  & \checkmark &  & \leveltwoicon \\
   TxAgent \cite{gao2025txagent} & Precision Therapy & & \checkmark & \checkmark & \checkmark & \leveltwoicon \\
   \rowcolor[HTML]{DAE8FC}
   Robin \cite{ghareeb2025robin} & Therapeutic Candidate Discovery & \checkmark & \checkmark & \checkmark & \checkmark & \levelthreeicon \\
   DrugAgent \cite{liu2024drugagent} & Drug Discovery Programming & & \checkmark & \checkmark & \checkmark & \leveltwoicon \\
   \rowcolor[HTML]{DAE8FC}
   LIDDIA \cite{averly2025liddia} & \textit{In Silico} Drug Discovery &  & \checkmark & \checkmark & \checkmark & \levelthreeicon \\
   PharmAgents \cite{gao2025pharmagents} & Virtual Drug Discovery &  & \checkmark & \checkmark & \checkmark & \levelthreeicon \\
   \rowcolor[HTML]{DAE8FC}
   CLADD \cite{lee2025rag} & RAG-based Drug Discovery & & \checkmark & \checkmark & & \leveltwoicon \\
   Tippy \cite{fehlis2025accelerating} & DMTA Cycle Automation & & \checkmark & \checkmark & \checkmark & \levelthreeicon \\
   \rowcolor[HTML]{DAE8FC}
   ACEGEN \cite{bou2024acegen} & Generative Drug Design & & & \checkmark & & \leveltwoicon \\
   AI Co-scientist \cite{gottweis2025towards} & Drug Repurposing \& Target Discovery &  & \checkmark & \checkmark & \checkmark & \leveltwoicon \\
   \rowcolor[HTML]{DAE8FC}
   Exploring Modularity \cite{van2025exploring} & Meta-Analysis of Drug Discovery Agents & - & - & - & - & - \\
   DO Challenge \cite{smbatyan2025can} & Benchmark for Drug Discovery Agents & - & - & - & - & - \\
   
   \bottomrule
   \end{tabularx}
\end{table*}

\subsection{General Frameworks and Methodologies}
Beyond specialized applications, a number of projects focus on creating foundational, adaptable agentic frameworks capable of addressing a wide range of biomedical research tasks. These systems emphasize self-evolution, modular design, and the integration of scientific principles to build more robust and versatile AI research assistants.

\textbf{STELLA} \cite{jin2025stella} is a self-evolving AI agent designed to overcome the limitations of static toolsets. Its method is a multi-agent architecture featuring two core adaptive mechanisms: an evolving \textbf{Template Library} for reasoning strategies and a dynamic \textbf{Tool Ocean} that expands as a dedicated agent autonomously discovers and integrates new bioinformatics tools. This design allows STELLA to learn from experience; its results show that its accuracy on challenging biomedical benchmarks systematically improves with increased trials, outperforming leading models. \textbf{Biomni} \cite{huang2025biomni} is presented as a general-purpose biomedical AI agent designed for flexibility across a wide array of tasks. Its method is based on decomposing complex user queries into multi-step plans and executing them by dynamically selecting from an expanding set of tools. \textbf{m-KAILIN} \cite{xiao2025m} is presented as a knowledge-driven agentic framework for biomedical corpus distillation, designed to enhance large language model training. Its method is based on a multi-agent collaboration architecture guided by the MeSH knowledge hierarchy, where specialized agents autonomously generate, evaluate, and refine question–answer pairs from scientific literature to produce high-quality, ontology-aligned datasets for biomedical LLMs. \textbf{BioResearcher} \cite{luo2025intention} is another end-to-end automated system that employs a modular, multi-agent architecture for search, literature processing, experimental design, and programming. A key feature of its method is an LLM-based reviewer for in-process quality control, which enabled the system to achieve an average execution success rate of 63.07\% across eight previously unmet research objectives. A more theoretical framework, \textbf{PiFlow} \cite{pu2025piflow}, recasts automated scientific discovery as a structured uncertainty reduction problem. Its information-theoretical method guides a multi-agent system's exploration using scientific principles, which resulted in a 73.55\% increase in discovery efficiency and a 94.06\% enhancement in solution quality in domains including biomolecule discovery. Finally, a perspective piece envisions future "AI scientists" as collaborative agents that integrate AI models, biomedical tools, and experimental platforms \cite{gao2024empowering}. The authors argue that such systems, which feature structured memory for continual learning, will empower human researchers by handling large-scale data analysis and repetitive tasks, leaving creative and strategic oversight to humans.

\subsection{Genomics, Transcriptomics, and Multi-Omics Analysis}
The fields of genomics, transcriptomics, and other omics disciplines are inundated with high-dimensional data from technologies like single-cell RNA sequencing (scRNA-seq), spatial transcriptomics, and mass spectrometry. Key challenges include the need for specialized computational skills to process and interpret this data, the difficulty of integrating multi-modal data, and the labor-intensive nature of designing and executing analysis workflows. AI agents are being developed to make accessible and automate these complex analyses.

A significant focus has been on automating single-cell data analysis. \textbf{BIA} \cite{xin2024bioinformatics} is an intelligent agent designed to autonomously perform bioinformatics analysis from natural language. Its method involves using an LLM to manage the entire pipeline, from data extraction and processing to workflow design, code generation, and final reporting, with a focus on scRNA-seq. The results demonstrate BIA's proficiency in complex information processing and task execution, showcasing a viable path to automated analysis. Similarly, \textbf{CellAgent} \cite{xiao2024cellagent} is a multi-agent framework designed for full automation. Its method is based on a hierarchical team of LLM-driven agents---a planner, executor, and evaluator---that are coordinated by a hierarchical decision-making mechanism. Crucially, it incorporates a self-iterative optimization loop that allows the system to autonomously refine its choice of tools and hyperparameters. When evaluated on a large benchmark, CellAgent consistently identified optimal analysis strategies, achieving high-quality results without human intervention. To enhance accessibility, \textbf{CompBioAgent} \cite{zhang2025compbioagent} offers a user-friendly web application that converts natural language queries into visualizations. Its method integrates an LLM with established platforms like CellDepot and Cellxgene VIP, allowing non-programmers to explore scRNA-seq data interactively. Shifting from executing predefined tasks to autonomous discovery, \textbf{CellVoyager} \cite{alber2025cellvoyager} is an agent that autonomously explores scRNA-seq datasets to generate novel hypotheses. Its method involves conditioning its exploration on a record of prior user-run analyses, allowing it to seek out new biological insights. In case studies, CellVoyager's findings were rated as creative and sound by the original study authors, and it successfully discovered a previously unreported link between increased transcriptional noise and aging in the brain.

Agents are also being tailored for other specific data types and experimental designs. \textbf{CRISPR-GPT} \cite{huang2024crispr} is an LLM agent that automates the intricate design of CRISPR gene-editing experiments. Its method augments an LLM with domain-specific knowledge and external tools to assist non-experts in selecting CRISPR systems, designing guide RNAs, and drafting experimental protocols. Its effectiveness was validated in a real-world use case. For analyzing gene expression data, the \textbf{Team of AI-made Scientists (TAIS)} \cite{liu2024toward} framework simulates a human research team. The method uses multiple LLMs to represent a project manager, a data engineer, and a domain expert that collaborate to identify disease-predictive genes. For designing sequential experiments, \textbf{PerTurboAgent} \cite{hao2025perturboagent} is a self-planning agent that excels at designing iterative Perturb-seq experiments. Through self-directed data analysis and knowledge retrieval, it prioritizes genes for subsequent rounds of testing, and its performance was shown to outperform existing active learning strategies in identifying impactful gene perturbations.

The analysis of spatial and multi-omics data presents further challenges of integration and interpretation. \textbf{SpatialAgent} \cite{wang2025spatialagent} is a fully autonomous agent for spatial biology research. Its method combines LLMs with dynamic tool execution and adaptive reasoning to manage the entire research pipeline, from experimental design to hypothesis generation. On complex datasets, its performance matched or exceeded that of human scientists. For phenotype-driven discovery, \textbf{PhenoGraph} \cite{niyakan2025phenograph} is a multi-agent system that automates the analysis of spatial transcriptomics data. A key aspect of its method is the augmentation of its reasoning with biological knowledge graphs, which enhances the interpretability of its findings. Addressing broader bioinformatics workflows, several agents aim to democratize access. \textbf{BioAgents} \cite{mehandru2025bioagents} uses a multi-agent system built on fine-tuned small language models and Retrieval-Augmented Generation (RAG), enabling accessible, local operation with expert-level performance. \textbf{BioMaster} \cite{su2025biomaster} employs a robust multi-agent framework with enhanced validation and memory management to reliably handle long, complex workflows like RNA-seq and ChIP-seq analysis, outperforming existing methods in scalability and accuracy. \textbf{TransAgent} \cite{zhang2025transagent} focuses specifically on transcriptional regulation, with a method that automates complex multi-omics data integration by integrating over 30 specialized tools and 20 data sources. Finally, agents are emerging for proteomics and mass spectrometry. \textbf{PROTEUS} \cite{ding2024automating, qu2025automating} is a fully automated system that takes raw proteomics or multi-omics data as input. Its method uses hierarchical planning and iterative workflow refinement to generate research objectives, analysis results, and novel, evaluable hypotheses. \textbf{AstroAgents} \cite{saeedi2025astroagents} is a multi-agent system designed specifically for hypothesis generation from mass spectrometry data.

\subsection{Protein Science and Engineering}
Designing novel proteins with specific functions or properties is a central goal in synthetic biology and biomedical engineering. This process involves navigating a vast sequence space and understanding complex relationships between sequence, structure, and function. Current AI models are often limited to specific objectives, lacking the flexibility to incorporate diverse knowledge or perform comprehensive analyses.

To address these limitations, agentic systems are being developed to create a more dynamic and collaborative design environment. \textbf{ProtAgents} \cite{ghafarollahi2024protagents} introduces a platform for \textit{de novo} protein design where multiple AI agents with distinct skills collaborate. Its method establishes a dynamic environment where agents specializing in knowledge retrieval, protein structure analysis, and physics-based simulations work in concert. The results demonstrated a synergistic approach where the system designed new proteins with targeted mechanical properties and performed novel analyses, such as calculating natural vibrational frequencies. This collaborative method allows for a more versatile and powerful approach to protein design. Expanding on this, \textbf{Sparks} \cite{ghafarollahi2025sparks} represents a significant leap towards autonomous scientific discovery. It is a multi-agent AI model that autonomously executes the entire discovery cycle: hypothesis generation, experiment design, and iterative refinement, culminating in a final report without human intervention. The method combines generative sequence design, high-accuracy structure prediction, and physics-aware models, with paired generation-and-reflection agents enforcing self-correction. When applied to protein science, Sparks independently uncovered two previously unknown phenomena: a length-dependent mechanical crossover in peptide unfolding force and a chain-length/secondary-structure stability map revealing unexpectedly robust architectures. These results demonstrate Sparks's ability to conduct rigorous scientific inquiry and discover novel, verifiable design principles, marking a key milestone for agentic science.

\subsection{Drug and Therapeutic Discovery}
Drug discovery is notoriously long, costly, and prone to failure. The process involves numerous stages, from target identification and lead compound generation to preclinical analysis and optimization. Agentic AI aims to create integrated, automated systems that can streamline this entire pipeline, reason about therapeutic strategies, and accelerate the identification of promising drug candidates.

Several agent frameworks function as comprehensive, end-to-end virtual drug discovery platforms. \textbf{PharmAgents} \cite{gao2025pharmagents} simulates a virtual pharmaceutical ecosystem with a method that uses LLM-driven agents equipped with specialized machine learning models to manage the entire workflow, from target discovery and lead compound optimization to \textit{in silico} analysis of toxicity and synthetic feasibility, establishing a paradigm for autonomous and scalable research. \textbf{LIDDiA} \cite{averly2025liddia} is an autonomous agent whose method leverages LLM reasoning to intelligently navigate the \textit{in silico} discovery process, strategically balancing exploration and exploitation of chemical space. As a result, it successfully generated molecules meeting key pharmaceutical criteria for over 70\% of 30 clinically relevant targets and identified promising novel candidates for the critical EGFR cancer target. \textbf{DrugAgent} \cite{liu2024drugagent} focuses on automating the crucial ML programming aspect of drug discovery. Its method employs a \textbf{Planner} agent to formulate high-level ideas and an \textbf{Instructor} agent to translate them into robust code, outperforming baselines with a 4.92\% relative improvement in ROC-AUC for drug-target interaction prediction. Bridging the virtual and physical, \textbf{Tippy} \cite{fehlis2025accelerating} is a production-ready multi-agent system designed to automate the full Design-Make-Test-Analyze (DMTA) cycle in a laboratory setting. Its method uses five specialized agents (Supervisor, Molecule, Lab, Analysis, Report) with safety guardrails, demonstrating significant improvements in workflow efficiency and decision-making speed. A modular framework detailed in \cite{ock2025large} combines LLM reasoning with domain-specific tools for tasks like molecular generation and refinement. In a case study targeting BCL-2, its iterative refinement process more than doubled the number of candidate molecules that passed key drug-likeness rules. Finally, \textbf{CLADD} \cite{lee2025rag} proposes a RAG-empowered agentic system that avoids costly domain-specific fine-tuning. Its method dynamically retrieves information from biomedical knowledge bases to contextualize queries, outperforming both general-purpose and domain-specific LLMs on a variety of discovery tasks.

Other agents focus on specific, critical stages of the discovery pipeline where AI can have an outsized impact. For therapeutic target discovery, \textbf{OriGene} \cite{zhang2025origene} acts as a "virtual disease biologist." Its method is a self-evolving multi-agent system that integrates diverse data modalities (genetics, pharmacology, clinical records) and uses human and experimental feedback to refine its reasoning. OriGene outperformed human experts on a large benchmark and, critically, nominated two previously underexplored targets for liver (GPR160) and colorectal cancer (ARG2) that showed significant anti-tumor activity in patient-derived organoid models. Also demonstrating real-world discovery, \textbf{Robin} \cite{ghareeb2025robin} is a multi-agent system that automated the intellectual steps of discovery, from background research to experimental design. This led to the identification of ripasudil, a clinically used ROCK inhibitor, as a novel therapeutic candidate for dry age-related macular degeneration (dAMD). Robin then proposed and analyzed a follow-up RNA-seq experiment to elucidate its mechanism of action. The AI co-scientist from \cite{gottweis2025towards} utilizes a "generate, debate, and evolve" methodology, where agents use a tournament evolution process to refine hypotheses. This approach led to the discovery of promising drug repurposing candidates for acute myeloid leukemia and novel epigenetic targets for liver fibrosis, both of which were subsequently validated in lab.

Agents are also being developed for experimental design and specialized therapeutic reasoning. \textbf{BioDiscoveryAgent} \cite{roohani2024biodiscoveryagent} designs genetic perturbation experiments by leveraging its intrinsic biological knowledge, avoiding the need for a pre-trained model or Bayesian acquisition function. This method led to a 21\% average improvement in predicting relevant genetic perturbations over specialized baselines. \textbf{TxAgent} \cite{gao2025txagent} is an agent specialized in therapeutic reasoning. Its method leverages a "ToolUniverse" of 211 validated tools to analyze drug interactions and contraindications, achieving 92.1\% accuracy on open-ended drug reasoning tasks. \textbf{ACEGEN} \cite{bou2024acegen} is a streamlined toolkit using reinforcement learning to create generative agents for drug design, which showed performance comparable to or better than other state-of-the-art generative algorithms. For nanomedicine, the \textbf{Virtual Lab} \cite{swanson2024virtual} used a team of LLM agents (chemist, computer scientist, critic) guided by a human to design novel nanobody binders for SARS-CoV-2. The agents created a design pipeline using ESM and AlphaFold, resulting in 92 candidates, two of which were experimentally validated to have improved binding to recent viral variants.

Finally, some work focuses on benchmarking and understanding the agentic systems themselves. The \textbf{DO Challenge} \cite{smbatyan2025can} introduces a benchmark to evaluate the ability of agents to design and implement drug discovery pipelines, testing their capacity to navigate chemical space and manage resources. Another study critically examines the modularity of these systems, finding that core components like LLMs are not easily interchangeable without significant prompt re-engineering, highlighting the need for research into developing stable and scalable solutions \cite{van2025exploring}.

\begin{table*}[!t]
\centering
\small
\caption{Examples of Validated Scientific Discoveries Achieved by AI Agents in Life Sciences.}
\label{tab:conclusion_life}

\renewcommand{\arraystretch}{0.9}
\setlength{\tabcolsep}{2.5pt}

\begin{tabularx}{\textwidth}{
    >{\raggedright\arraybackslash\bfseries\scriptsize}p{2.5cm} 
    >{\raggedright\arraybackslash\scriptsize}p{4.5cm} 
    >{\raggedright\arraybackslash\scriptsize}X 
}
\toprule
\textbf{Agent System} & \textbf{Application Domain} & \textbf{Novel Scientific Contribution or Validated Discovery} \\
\midrule

\rowcolor[HTML]{DAE8FC}
ProtAgents \cite{ghafarollahi2024protagents} & De Novo Protein Design & Designed new proteins and obtained new first-principles data (natural vibrational frequencies) via physics simulations. \\

The Virtual Lab \cite{swanson2024virtual} & Nanobody Design for SARS-CoV-2 & Designed 92 new nanobodies, with experimental validation confirming two candidates exhibit improved binding to recent SARS-CoV-2 variants (JN.1 or KP.3). \\

\rowcolor[HTML]{DAE8FC}
Sparks \cite{ghafarollahi2025sparks} & Protein Principle Discovery & Discovered two previously unknown phenomena: 1) a length-dependent mechanical crossover in peptide unfolding force, establishing a new design principle, and 2) a stability map revealing robust beta-sheet architectures and a "frustration zone" in mixed folds. \\

OriGene \cite{zhang2025origene} & Therapeutic Target Discovery & Nominated and validated previously underexplored therapeutic targets for liver cancer (GPR160) and colorectal cancer (ARG2), which showed significant anti-tumor activity in patient-derived models. \\

\rowcolor[HTML]{DAE8FC}
Robin \cite{ghareeb2025robin} & Therapeutic Candidate Discovery & Identified and validated a novel treatment for dry age-related macular degeneration (dAMD), the clinically-used drug ripasudil. It also proposed a novel therapeutic target (ABCA1) by elucidating the drug's mechanism. \\

CellVoyager \cite{alber2025cellvoyager} & scRNA-seq Discovery & Autonomously re-analyzing existing datasets, it generated new, validated insights: 1) discovered that CD8+ T cells in COVID-19 are primed for pyroptosis, and 2) found a previously unreported link between increased transcriptional noise and aging in the brain's subventricular zone. \\

\rowcolor[HTML]{DAE8FC}
AI co-scientist \cite{gottweis2025towards} & Drug Repurposing \& Target Discovery & Proposed and validated new uses for existing drugs for acute myeloid leukemia. It also discovered and validated new epigenetic targets for liver fibrosis using human hepatic organoids and independently discovered a novel gene transfer mechanism in bacteria. \\

\bottomrule
\end{tabularx}
\end{table*}

%% file: Sections/6.Chemistry.tex
\begin{table*}[!t]

   \centering
   \small
   \caption{Classification of Agentic Systems in Chemistry Science, organized to correspond with the survey text. The checkmark (\checkmark) indicates the system's primary capabilities. Column Key: \textbf{Hypo.}: Observation or Hypothesis Generation, \textbf{Exper.}: Experimental Planning or Execution, \textbf{Analysis}: Data and Result Analysis, \textbf{Validation}: Synthesis, Validation, and Evolution. \textbf{Level}: Level of autonomy. {\leveltwoicon} means level 2 and {\levelthreeicon} means level 3.}
   \label{tab:Chemistry}

   \renewcommand{\arraystretch}{1.0}
   \setlength{\tabcolsep}{1.3pt}

   \begin{tabularx}{\textwidth}{l >{\scriptsize}X cccc c}
   \toprule
   & & \multicolumn{4}{c}{\textbf{Core Process}} & \\
   \cmidrule(lr){3-6}
   \textbf{Paper} & \textbf{Application Domain} & 
   \textbf{Hypo.} & \textbf{Exper.} & \textbf{Analysis} & \textbf{Validation} & \textbf{Level} \\
   \midrule

   \multicolumn{7}{c}{\textit{General Frameworks and Methodologies}} \\
   \rowcolor[HTML]{F5F4C9}
   ChemCrow \cite{bran2023chemcrow} & Organic Synthesis &  & \checkmark & \checkmark &  & \leveltwoicon \\
   ChemAgents \cite{song2025multiagent} & Hierarchical Multi-Agent Robotic Chemist & \checkmark & \checkmark & \checkmark & \checkmark & \levelthreeicon \\
   \rowcolor[HTML]{F5F4C9}
   MOOSE-Chem \cite{yang2024moose} & Rediscovery of Scientific Hypotheses & \checkmark & \checkmark &  & & \leveltwoicon \\
   MOOSE-Chem3 \cite{liu2025moose} & Experiment-Guided Hypothesis Ranking & & \checkmark & \checkmark & \checkmark & \leveltwoicon \\
   \rowcolor[HTML]{F5F4C9}
   ChemMiner \cite{chen2024autonomous} & Agent for Chemical Literature Data Mining & & \checkmark & \checkmark & & \leveltwoicon \\
   Eunomia \cite{ansari2024agent} & Agent for Building Datasets From Literature &  & \checkmark & \checkmark & \checkmark & \leveltwoicon \\
   \rowcolor[HTML]{F5F4C9}
   ChemAgent \cite{wu2025chemagent} & Tool Learning &  & \checkmark & \checkmark & \checkmark & \leveltwoicon \\
   ChemHAS \cite{li2025chemhas} & Hierarchical Agent Stacking to Enhance Tools & & \checkmark & & \checkmark & \leveltwoicon \\
   \rowcolor[HTML]{F5F4C9}
   ChemToolAgent \cite{yu2024chemtoolagent} & Meta-Analysis of Tool Impact & & \checkmark & \checkmark & \checkmark & \leveltwoicon \\
   Chemagent \cite{tang2025chemagent} & Improving Reasoning With a Library & & \checkmark & \checkmark & \checkmark & \leveltwoicon \\
   \rowcolor[HTML]{F5F4C9}
   LabUtopia \cite{li2025labutopia} & Simulation for Embodied Agents & & \checkmark &  &  & \leveltwoicon \\
   CACTUS \cite{mcnaughton2024cactus} & Agent Connecting Tools for Problem-Solving & & \checkmark & \checkmark & & \leveltwoicon \\
   \rowcolor[HTML]{F5F4C9}
   GVIM \cite{ma2025ai} & Intelligent Research Assistant System & \checkmark & \checkmark & \checkmark & \checkmark & \levelthreeicon \\
   MT-Mol \cite{kim2025mt} & Multi-Agent System for Molecular Optimization & \checkmark & \checkmark & \checkmark & \checkmark & \levelthreeicon \\
   \rowcolor[HTML]{F5F4C9}
   CSstep \cite{che2025csstep} & Multi-Agent RL for Exploring Chemical Space & \checkmark & & \checkmark & \checkmark & \leveltwoicon \\
   CRAG-MoW \cite{callahan2025agentic} & Mixture-of-Workflows for Multi-Modal Search & & \checkmark & \checkmark & \checkmark & \leveltwoicon \\
   \midrule

   \multicolumn{7}{c}{\textit{Organic Synthesis and Reaction Optimization}} \\
   \rowcolor[HTML]{F5F4C9}
   Coscientist \cite{boiko2023autonomous} & Reaction Optimization (Pd Cross-Coupling) & \checkmark & \checkmark & \checkmark & \checkmark & \levelthreeicon \\
   LLM-RDF \cite{ruan2024automatic} & End-to-End Synthesis Development & \checkmark & \checkmark & \checkmark & \checkmark & \levelthreeicon \\
   \rowcolor[HTML]{F5F4C9}
   Chemist-X \cite{chen2023chemist} & Reaction Condition Optimization & \checkmark & \checkmark & \checkmark & \checkmark & \levelthreeicon \\
   ORGANA \cite{darvish2024organa} & Robotic Chemistry Experimentation & & \checkmark & \checkmark & \checkmark & \leveltwoicon \\
   \rowcolor[HTML]{F5F4C9}
   Dai et al. \cite{dai2024autonomous} & Exploratory Synthesis With Mobile Robots & \checkmark & \checkmark & \checkmark & \checkmark & \levelthreeicon \\
   Strieth-Kalthoff et al. \cite{strieth2024delocalized} & Closed-Loop Discovery of Laser Emitters & \checkmark & \checkmark & \checkmark & \checkmark & \levelthreeicon \\
   \rowcolor[HTML]{F5F4C9}
   AutoChemSchematic AI \cite{srinivas2025autochemschematic} & Generation of Industrial Process Diagrams & \checkmark & \checkmark & & \checkmark & \levelthreeicon \\
   \midrule

   \multicolumn{7}{c}{\textit{Generative Chemistry and Molecular Design}} \\
   \rowcolor[HTML]{F5F4C9}
   ChatMOF \cite{kang2023chatmof} & Generative Design of MOFs & \checkmark & \checkmark & \checkmark & & \leveltwoicon \\
   MOFGen \cite{inizan2025system} & \textit{De Novo} Discovery of Synthesizable MOFs & \checkmark & \checkmark & \checkmark & \checkmark & \levelthreeicon \\
   \rowcolor[HTML]{F5F4C9}
   OSDA Agent \cite{hu2025osda} & \textit{De Novo} Design of Molecules for Zeolites & \checkmark & & \checkmark & \checkmark & \levelthreeicon \\
   ChemReasoner \cite{sprueill2024chemreasoner} & Heuristic Search for Catalyst Discovery & \checkmark & \checkmark & \checkmark & \checkmark & \levelthreeicon \\
   \rowcolor[HTML]{F5F4C9}
   Horwood \& Noutahi \cite{horwood2020molecular} & Molecular Design via Reinforcement Learning & \checkmark & \checkmark & \checkmark & \checkmark & \levelthreeicon \\
   \midrule

   \multicolumn{7}{c}{\textit{Computational and Quantum Chemistry}} \\
   \rowcolor[HTML]{F5F4C9}
   El Agente Q~\cite{2025arXiv250502484Z} & Autonomous Quantum Chemistry Workflows & & \checkmark & \checkmark & \checkmark & \leveltwoicon \\
   Aitomia \cite{hu2025aitomia} & Intelligent Assistant for Atomistic Simulations & & \checkmark & \checkmark & & \leveltwoicon \\
   \rowcolor[HTML]{F5F4C9}
   ChemGraph \cite{pham2025chemgraph} & Automated Computational Chemistry Workflows &  & \checkmark & \checkmark & \checkmark & \leveltwoicon \\
   xChemAgents \cite{polat2025xchemagents} & Explainable Quantum Chemistry Prediction & \checkmark & & \checkmark & \checkmark & \leveltwoicon \\

   \bottomrule
   \end{tabularx}
\end{table*}

\section{Agentic Chemistry Research}
\label{sec:chemistry}

The application of agentic AI is rapidly transforming chemical research, automating complex processes from hypothesis generation to experimental execution and analysis. By integrating large language models (LLMs) with specialized chemical tools and robotic platforms, these AI agents can autonomously design and perform experiments, discover novel materials, and optimize synthetic reactions. This section surveys the emerging landscape of agentic chemistry, categorized by the primary function of the agents, to highlight the major challenges, proposed frameworks, and significant achievements (Table~\ref{tab:Chemistry} and Table~\ref{tab:conclusion_chemistry}).

\subsection{General Frameworks and Methodologies}
Beyond specific applications, a significant body of research focuses on developing the foundational methodologies, architectures, and tools that support chemical AI agents. This work addresses broad challenges such as effective tool integration, robust reasoning, hypothesis generation, and literature comprehension, which are essential for creating truly autonomous and versatile scientific agents.

Several papers propose general-purpose agent frameworks designed for broad chemical tasks. \textbf{ChemCrow} is an LLM agent augmented with 18 expert-designed tools to accomplish tasks across organic synthesis, drug discovery, and materials design~\cite{bran2023chemcrow}. It demonstrated its capability by autonomously planning and executing the synthesis of an insect repellent and several organocatalysts. Similarly, \textbf{ChemAgents} is a hierarchical multi-agent system powered by an on-board LLM that coordinates four role-specific agents--a Literature Reader, Experiment Designer, Computation Performer, and Robot Operator--to execute complex, multi-step experiments with minimal human intervention~\cite{song2025multiagent}. Methodologies for reasoning and hypothesis generation are also critical. \textbf{MOOSE-Chem} formalizes hypothesis discovery by decomposing the task into retrieving inspirations from literature, composing hypotheses, and ranking them, successfully rediscovering the core innovations of 51 recent high-impact papers~\cite{yang2024moose}. Following this, \textbf{MOOSE-Chem3} tackles the problem of ranking these hypotheses by introducing an "experiment-guided" approach that uses a simulator to generate feedback, allowing it to prioritize candidates based on the outcomes of previously tested ones~\cite{liu2025moose}.

A central theme is the effective use of tools and knowledge. \textbf{ChemAgent} (by Wu et al.) integrates 137 external chemical tools using a Hierarchical Evolutionary Monte Carlo Tree Search (HE-MCTS) framework for planning and execution, significantly improving performance on QA and discovery tasks~\cite{wu2025chemagent}. Taking a different angle, \textbf{ChemHAS} explores how agents can enhance the tools themselves, proposing a hierarchical agent stacking method to compensate for the inherent prediction errors of chemistry tools~\cite{li2025chemhas}. However, \textbf{ChemToolAgent} provides a nuanced analysis, finding that while tools are beneficial for specialized tasks like synthesis prediction, they do not consistently improve performance on general chemistry questions where core knowledge and reasoning are paramount~\cite{yu2024chemtoolagent}. To improve knowledge integration, \textbf{Chemagent} (by Tang et al.) uses a dynamic, self-updating library compiled from decomposed sub-tasks to enhance reasoning, achieving performance gains of up to 46\% on the SciBench benchmark~\cite{tang2025chemagent}. For knowledge extraction, systems like \textbf{ChemMiner}~\cite{chen2024autonomous}, which uses three specialized agents for text, multimodal, and synthesis analysis, and \textbf{Eunomia}~\cite{ansari2024agent}, which autonomously creates structured datasets from unstructured text, are designed to mine accurate data from the scientific literature.

Finally, some works provide crucial infrastructure for the field. \textbf{LabUtopia} offers a comprehensive simulation and benchmarking suite specifically for training and evaluating scientific embodied agents~\cite{li2025labutopia}. It includes an accurate simulator (`LabSim'), a procedural scene generator (`LabScene'), and a hierarchical benchmark (`LabBench'), providing a rigorous platform to advance the integration of perception, planning, and control in laboratory settings.

\subsection{Organic Synthesis and Reaction Optimization}
Organic synthesis is a key area of chemistry, yet it presents considerable challenges, including the laborious optimization of reaction conditions, the creative design of multi-step synthetic routes, and the safe execution of complex experimental protocols. Agentic AI is being developed to address these issues by automating the entire workflow, from experimental design to robotic execution and result interpretation, thereby accelerating the pace of discovery.

A primary focus has been on automating reaction optimization and execution. For instance, \textbf{Coscientist} \cite{boiko2023autonomous} is an AI system driven by GPT-4 that showcases the ability to autonomously design, plan, and execute complex experiments from start to finish. In a notable demonstration, it successfully optimized the reaction conditions for palladium-catalyzed cross-couplings, a widely used and important reaction class. Similarly, the \textbf{LLM-based Reaction Development Framework (LLM-RDF)} \cite{ruan2024automatic} employs a suite of six specialized agents--a Literature Scouter, Experiment Designer, Hardware Executor, Spectrum Analyzer, Separation Instructor, and Result Interpreter--to manage the entire synthesis development workflow. The framework demonstrated its utility by guiding the end-to-end process for several reaction types, including copper/TEMPO catalyzed alcohol oxidation, from literature review and condition screening to scale-up and purification. Addressing the same challenge, \textbf{Chemist-X} \cite{chen2023chemist} targets reaction condition optimization by implementing a novel retrieval-augmented generation (RAG) scheme. This allows the agent to first consult molecular and literature databases to narrow the search space before an AI controller executes the proposed conditions in a wet lab using an automated robotic system.

Another line of research focuses on integrating agentic AI with robotics to physically perform experiments. \textbf{ORGANA} \cite{darvish2024organa} acts as a robotic assistant that automates diverse and labor-intensive experiments such as solubility testing, pH measurement, and recrystallization. It interacts with chemists via natural language to derive experimental goals and provides detailed logs, with user studies showing it reduces physical demand by over 50\% and saves researchers an average of 80\% of their time. \textbf{Autonomous mobile robots} \cite{dai2024autonomous} have been deployed to create a more flexible automated lab. These robots physically shuttle samples between standard, unmodified laboratory instruments like a synthesis platform, an LC-MS, and an NMR spectrometer. This approach allows automated systems to share equipment with human researchers and enables a heuristic decision-maker to process orthogonal data from multiple analysis techniques to guide the experimental campaign. A landmark achievement in this area is the \textbf{delocalized, asynchronous, closed-loop discovery} of organic laser emitters \cite{strieth2024delocalized}. This work utilized a cloud-based AI planner to coordinate robotic synthesis and characterization across five international laboratories. This distributed workflow resulted in the discovery of 21 new state-of-the-art materials, demonstrating a blueprint for global, accessible scientific discovery. Finally, bridging the gap from laboratory discovery to industrial application, \textbf{AutoChemSchematic AI} \cite{srinivas2025autochemschematic} is a closed-loop, physics-aware framework designed to automatically generate industrial-scale Process Flow Diagrams (PFDs) and Piping and Instrumentation Diagrams (P\&IDs). It integrates specialized language models with a process simulator (DWSIM) to ensure the generated plans are physically viable, streamlining the transition from bench-scale chemistry to full-scale manufacturing.

\subsection{Generative Chemistry and Molecular Design}
A major frontier in chemistry is the \textit{de novo} design of novel molecules and materials with precisely tailored properties. This involves navigating a vast and complex chemical space to identify promising candidates. Agentic AI excels at this generative task by combining large-scale knowledge models with targeted search strategies and computational validation.

The design of porous materials has been a significant target for generative agents. \textbf{ChatMOF} \cite{kang2023chatmof} is an autonomous AI system that uses GPT-4 to process natural language queries for predicting properties and generating new Metal-Organic Frameworks (MOFs). Its architecture comprises three core components--an agent, a toolkit, and an evaluator--and has demonstrated high accuracy (over 95\% for prediction) in performing its designated tasks. Building on this, \textbf{MOFGen} \cite{inizan2025system} employs a more complex system of interconnected agents to discover novel, synthesizable MOFs. This system includes a large language model as a proposer, a diffusion model for generating 3D crystal structures, quantum mechanical agents for computational validation, and synthetic-feasibility agents guided by expert rules. This powerful combination led to the generation of hundreds of thousands of novel MOF structures and resulted in the successful experimental synthesis of five entirely new ``AI-dreamt'' MOFs.

Agents are also being created to design specific functional molecules for complex applications. For zeolite synthesis, \textbf{OSDA Agent} \cite{hu2025osda} performs \textit{de novo} design of Organic Structure Directing Agents (OSDAs) using an LLM-based Actor-Evaluator-Self-reflector framework. The Actor generates potential OSDAs, the Evaluator uses computational chemistry to score them, and the Self-reflector analyzes the results to provide feedback, creating a refinement loop that improves generation quality. For catalyst discovery, \textbf{ChemReasoner} \cite{sprueill2024chemreasoner} integrates LLM-based reasoning with quantum-chemical feedback. The agent formulates hypotheses about effective catalysts and iteratively refines its search by using feedback from atomistic simulations, which provide scoring functions based on adsorption energies and reaction barriers to steer the exploration toward highly effective candidates. Another approach utilizes deep reinforcement learning to explore chemical space under realistic constraints \cite{horwood2020molecular}. Here, an agent learns to optimize pharmacologically relevant objectives by navigating a space composed only of synthetically accessible molecules. This is achieved by defining state transitions within the Markov decision process as known chemical reactions, effectively using established synthetic routes as a powerful inductive bias to ensure the generated molecules are practical to create.

\subsection{Computational and Quantum Chemistry}
Computational and quantum chemistry provide powerful tools for understanding molecular behavior, but their use often requires specialized expertise to set up, execute, and interpret complex simulations. Agentic AI is emerging as a solution to make these tools accessible by creating intelligent assistants that can translate natural language prompts into executable workflows, manage simulations, and analyze results.

Several agentic systems function as intelligent assistants for complex simulations. \textbf{El Agente Q} is an LLM-based multi-agent system that dynamically generates and executes quantum chemistry workflows from natural language prompts~\cite{2025arXiv250502484Z}. It is built on a novel cognitive architecture featuring a hierarchical memory framework that enables flexible task decomposition and adaptive tool selection. It demonstrated robust problem-solving, achieving an average success rate of over 87\% on benchmark tasks, and features adaptive error handling through \textit{in situ} debugging. Similarly, \textbf{Aitomia} is a publicly accessible online platform with AI agents and chatbots that assists both experts and non-experts in running atomistic and quantum chemical simulations~\cite{hu2025aitomia}. It leverages open-source LLMs, rule-based agents, and a RAG system to handle setup, monitoring, analysis, and summarization for a wide range of tasks, including geometry optimizations and spectra calculations.

Other frameworks focus on creating structured, automated workflows for specific computational tasks. \textbf{ChemGraph} is an agentic framework designed to simplify and automate computational chemistry workflows, such as geometry optimization, vibrational analysis, and thermochemistry calculations~\cite{pham2025chemgraph}. It uses LLMs for natural language understanding and task planning while leveraging graph neural network (GNN)-based foundation models for accurate and efficient calculations, demonstrating that multi-agent decomposition can enable smaller LLMs to match the performance of larger models on complex tasks. Aiming for improved explainability and accuracy, \textbf{xChemAgents} introduces a cooperative agent framework for quantum chemistry property prediction~\cite{polat2025xchemagents}. It comprises two agents: a Selector, which adaptively identifies a sparse, relevant subset of chemical descriptors and provides a natural language rationale, and a Validator, which enforces physical constraints through iterative dialogue. This approach achieved up to a 22\% reduction in mean absolute error over baselines while producing human-interpretable explanations.

\begin{table*}[!t]
\centering
\small
\caption{Examples of Validated Scientific Discoveries Achieved by AI Agents in Chemistry.}
\label{tab:conclusion_chemistry}
\renewcommand{\arraystretch}{1.5}
\setlength{\tabcolsep}{6pt}
\begin{tabularx}{\textwidth}{
    >{\RaggedRight\bfseries\scriptsize}p{3.5cm}
    >{\RaggedRight\scriptsize}p{4cm}
    >{\RaggedRight\arraybackslash\scriptsize}X
}
\toprule
\textbf{Agent System} & \textbf{Application Domain} & \textbf{Novel Scientific Contribution or Validated Discovery} \\
\midrule

\rowcolor[HTML]{F5F4C9}
Cloud-based AI Planner \cite{strieth2024delocalized} & Closed-loop Discovery of Laser Emitters & Discovered \textbf{21 new state-of-the-art organic solid-state laser emitters}. The agent orchestrated a workflow across five laboratories, leading to the gram-scale synthesis and verification of a material with best-in-class stimulated emission. \\

ChemCrow \cite{bran2023chemcrow} & Organic Synthesis & Guided the discovery of a \textbf{novel chromophore} by autonomously planning and executing the required synthesis steps. \\

\rowcolor[HTML]{F5F4C9}
ChemAgents \cite{song2025multiagent} & Hierarchical Multi-agent Robotic Chemist & Executed complex, multistep experiments that culminated in the \textbf{discovery and optimization of new functional materials}. \\

MOFGen \cite{inizan2025system} & \textit{De novo} Discovery of Synthesizable MOFs & Designed novel Metal-Organic Frameworks (MOFs), leading to the successful experimental \textbf{synthesis of five previously unknown "AI-dreamt" MOFs}, validating the system's ability to create synthesizable materials. \\

\bottomrule
\end{tabularx}
\end{table*}

%% file: Sections/7.Materials.tex
\section{Agentic Materials Science Research}
\label{sec:materials}

This section delves into the application of agentic AI frameworks in materials science, a field ripe for automation due to its vast design spaces and complex, multi-step discovery workflows. We categorize the contributions into three main areas: the design and discovery of novel materials, the automation of simulation and characterization processes, and the development of general discovery platforms (Table~\ref{tab:Materials} and Table~\ref{tab:conclusion_materials}).

\begin{table*}[!t]
    \centering
    \small 
    \caption{Classification of Agentic Systems in Materials Science. The table is organized to correspond with the survey sections. Column Key: \textbf{Hypo.}: Observation or Hypothesis Generation, \textbf{Exper.}: Experimental Planning or Execution, \textbf{Analysis}: Data and Result Analysis, \textbf{Validation}: Synthesis, Validation, and Evolution. {\leveltwoicon} means level 2 and {\levelthreeicon} means level 3.}
    \label{tab:Materials}
    
    \renewcommand{\arraystretch}{1.2}
    \setlength{\tabcolsep}{3pt}
    
   \begin{tabularx}{\textwidth}{l >{\raggedright\arraybackslash}X ccccc}
    \toprule
    & & \multicolumn{5}{c}{\textbf{Core Process}} \\
    \cmidrule(lr){3-7}
    \textbf{Paper} & \textbf{Application Domain} & 
    \textbf{Hypo.} & \textbf{Exper.} & \textbf{Analysis} & \textbf{Validation} & \textbf{Level} \\
    \midrule

    \multicolumn{7}{c}{\textit{General Methodologies and Discovery Platforms}} \\
    \rowcolor[HTML]{E8F5E9}
    MatPilot \cite{ni2024matpilot} & General Materials Discovery & \checkmark & \checkmark & \checkmark & \checkmark & \levelthreeicon \\
    LLMatDesign \cite{jia2024llmatdesign} & General Materials Design & \checkmark & \checkmark & \checkmark & \checkmark & \levelthreeicon \\
    \rowcolor[HTML]{E8F5E9}
    MAPPS \cite{zhou2025toward} & Autonomous Materials Discovery & \checkmark & \checkmark & \checkmark & \checkmark & \leveltwoicon \\
    dZiner \cite{ansari2024dziner} & Inverse Molecular Design & \checkmark & \checkmark & \checkmark & \checkmark & \levelthreeicon \\
    \rowcolor[HTML]{E8F5E9}
    LLaMP \cite{chiang2024llamp} & Materials Informatics (RAG) & \checkmark & \checkmark & \checkmark & & \leveltwoicon \\
    HoneyComb \cite{zhang2024honeycomb} & Materials Knowledge Systems & \checkmark & \checkmark & \checkmark & & \leveltwoicon \\
    \rowcolor[HTML]{E8F5E9}
    PiFlow \cite{pu2025piflow} & General Discovery Methodology & \checkmark & \checkmark & \checkmark & \checkmark & \levelthreeicon \\
    Kumbhar et al. \cite{kumbhar2025hypothesis} & Scientific Hypothesis Generation & \checkmark & & \checkmark & & \leveltwoicon \\
    \rowcolor[HTML]{E8F5E9}
    Bazgir et al. \cite{bazgir2025multicrossmodal} & Multimodal Data Integration & \checkmark & & \checkmark & & \leveltwoicon \\
    \midrule
    
    \multicolumn{7}{c}{\textit{Design and Discovery of Novel Materials}} \\
    \rowcolor[HTML]{E8F5E9}
    AtomAgents \cite{ghafarollahi2025automating, ghafarollahi2024atomagents} & Alloy Design & \checkmark & \checkmark & \checkmark & \checkmark & \levelthreeicon \\
    Ghafarollahi et al. \cite{ghafarollahi2024rapid} & Alloy Design & \checkmark & \checkmark & \checkmark & \checkmark & \levelthreeicon \\
    \rowcolor[HTML]{E8F5E9}
    SciAgents \cite{ghafarollahi2025sciagents} & Biologically Inspired Materials & \checkmark & & \checkmark & \checkmark & \levelthreeicon \\
    PriM \cite{lai2025prim} & Nanomaterial Mechanics & \checkmark & \checkmark & \checkmark & \checkmark & \levelthreeicon \\
    \rowcolor[HTML]{E8F5E9}
    TopoMAS \cite{zhang2025topomas} & Topological Materials & \checkmark & \checkmark & \checkmark & \checkmark & \levelthreeicon \\
    metaAgent \cite{hu2025electromagnetic} & Electromagnetic Metamaterials &  & \checkmark & \checkmark & \checkmark & \leveltwoicon \\
    \rowcolor[HTML]{E8F5E9}
    CrossMatAgent \cite{tian2025multi} & Generative Metamaterial Design & \checkmark & \checkmark & \checkmark & \checkmark & \levelthreeicon \\
    Lu et al. \cite{lu2025agentic} & Inverse Photonic Design & & \checkmark & \checkmark & \checkmark & \leveltwoicon \\
    \midrule
    
    \multicolumn{7}{c}{\textit{Automated Simulation and Characterization}} \\
    \rowcolor[HTML]{E8F5E9}
    AILA \cite{mandal2024autonomous} & AFM Nanocharacterization & & \checkmark & \checkmark & & \leveltwoicon \\
    Foam-Agent \cite{yue2025foam} & Computational Fluid Dynamics (CFD) & & \checkmark & & \checkmark & \leveltwoicon \\
    \rowcolor[HTML]{E8F5E9}
    ChemGraph \cite{pham2025chemgraph} & Computational Chemistry (DFT, MD) & & \checkmark & \checkmark & & \leveltwoicon \\
    MechAgents \cite{ni2024mechagents} & Computational Solid Mechanics & & \checkmark & & \checkmark & \leveltwoicon \\
    
    \bottomrule
    \end{tabularx}
\end{table*}

\subsection{General Methodologies and Discovery Platforms}

Beyond specialized applications, a significant research effort focuses on creating general, flexible, and robust agentic platforms for materials science. The primary challenges are integrating diverse data sources, ensuring the reliability of LLM-generated knowledge, enabling autonomous planning of complex workflows, and facilitating seamless human-AI collaboration. These general-purpose platforms aim to provide extensible frameworks that can be adapted to various subdomains of materials science.

Several platforms focus on improving the reliability and knowledge-grounding of agents. \textbf{LLaMP} \cite{chiang2024llamp} is a retrieval-augmented generation (RAG) framework that uses a hierarchy of agents to interact with materials databases (like the Materials Project) and run simulations. By dynamically fetching and processing data, LLaMP effectively mitigates LLM hallucination without fine-tuning, demonstrating strong performance in retrieving properties like bulk moduli and bandgaps. \textbf{HoneyComb} \cite{zhang2024honeycomb} is another agent system designed specifically for materials science, addressing the issue of outdated or inaccurate knowledge in general-purpose LLMs. It introduces a high-quality, curated materials science knowledge base (MatSciKB) and a tool hub with an inductive method for creating and refining tools, significantly outperforming baseline models on specialized tasks.

Other frameworks concentrate on creating comprehensive, human-in-the-loop "AI scientists". \textbf{MatPilot} \cite{ni2024matpilot} is an LLM-enabled AI materials scientist designed for human-machine collaboration. It integrates human cognitive strengths with AI capabilities for information processing and storage. MatPilot can generate hypotheses, design experiments, and control an automated experimental platform, demonstrating a closed loop of iterative optimization and learning. \textbf{MAPPS} (Materials Agent unifying Planning, Physics, and Scientists) \cite{zhou2025toward} aims to grant agents greater autonomy by automating the planning of entire discovery workflows from high-level goals. Its architecture includes a Workflow Planner, a Tool Code Generator that invokes physics-based models, and a Scientific Mediator to incorporate human feedback and manage errors. MAPPS achieved a five-fold improvement in generating stable and novel crystal structures compared to previous models.

The ability to generate and evaluate hypotheses is a critical component of scientific discovery. To this end, researchers have developed a novel dataset and evaluation metric specifically for testing the ability of LLM agents to generate viable materials discovery hypotheses under given constraints \cite{kumbhar2025hypothesis}. This work provides a structured framework for advancing and benchmarking the hypothesis-generation capabilities of future agentic systems. Finally, to handle the diverse and siloed nature of materials data, a \textbf{multicrossmodal agent framework} \cite{bazgir2025multicrossmodal} was developed. It uses a team of specialized agents to process different data types (images, text, tables, videos), projecting their insights into a shared embedding space for unified reasoning. This approach enhances data integration and retrieval accuracy without requiring expensive model retraining.

\subsection{Design and Discovery of Novel Materials}

The design of new materials with specific target properties is a cornerstone of materials science, yet it presents immense challenges. The chemical design space is combinatorially vast, making exhaustive exploration impossible. Traditional methods rely on expert intuition and laborious trial-and-error, which are often slow and biased. A key challenge is to develop systems that can autonomously navigate this space, generate plausible hypotheses, and iteratively refine designs based on physical principles and computational feedback, thereby accelerating the discovery of materials for applications ranging from sustainable energy to advanced electronics.

Agentic frameworks are being developed to address these challenges by automating the discovery cycle. For instance, \textbf{SciAgents} \cite{ghafarollahi2025sciagents} was designed to automate the discovery of novel biologically inspired materials. The framework employs a multi-agent system that utilizes a large-scale knowledge graph to represent scientific concepts. These agents autonomously generate and refine research hypotheses by identifying hidden relationships in data, leading to the discovery of a new biocomposite with enhanced mechanical properties and sustainability. Similarly, in the realm of alloy design, a widely known as complex multi-objective problem, \textbf{AtomAgents} \cite{ghafarollahi2025automating} utilizes a physics-aware multi-agent system to design alloys with superior properties. The agents, with specialized roles in knowledge retrieval, simulation, and analysis, collaborate to navigate the design space, successfully identifying new alloys with enhanced characteristics. A related work automates this process further by integrating a Graph Neural Network (GNN) for rapid property prediction, reducing the reliance on costly simulations and accelerating the discovery of novel NbMoTa-based alloys \cite{ghafarollahi2024rapid}.

Other systems focus on specific classes of advanced materials. \textbf{TopoMAS} \cite{zhang2025topomas} is a multi-agent system dedicated to discovering topological materials. It coordinates the entire workflow from data retrieval to first-principles validation, guided by human-AI collaboration. A key feature is its dynamic knowledge graph, which is continuously updated with computational results, enabling iterative knowledge refinement. TopoMAS successfully identified a novel topological phase, SrSbO$_3$. For metamaterials, \textbf{CrossMatAgent} \cite{tian2025multi} integrates LLMs (GPT-4o) with generative models (DALL-E 3, Stable Diffusion) to automate design. Its hierarchical agent team specializes in tasks like pattern analysis and synthesis, producing simulation-ready designs. Another framework for photonic metamaterials uses an agent to autonomously develop a deep learning model for inverse design based on a desired optical spectrum \cite{lu2025agentic}. In a different approach, the \textbf{metaAgent} \cite{hu2025electromagnetic} operates as a cognitive entity that reasons in natural language to perform complex electromagnetic field manipulations, demonstrating advanced capabilities by planning and executing tasks in collaboration with robots and humans.

Inverse design, which aims to find a material structure given a desired property, is another area of focus. \textbf{dZiner} \cite{ansari2024dziner} is an AI agent that performs rational inverse design by leveraging literature insights to propose new compounds (e.g., surfactants, ligands, MOFs) and iteratively evaluates them with surrogate models. The framework supports both fully autonomous and human-in-the-loop workflows. Similarly, \textbf{LLMatDesign} \cite{jia2024llmatdesign} uses LLM agents to translate human instructions into material modifications, demonstrating effective zero-shot adaptation for designing materials with user-defined properties in silico. Other works like \textbf{PriM} \cite{lai2025prim} and \textbf{PiFlow} \cite{pu2025piflow} emphasize guidance by scientific principles. PriM uses a multi-agent "roundtable" to guide the discovery of nano-helical materials, while PiFlow frames discovery as a principle-guided uncertainty reduction problem, showing significant efficiency gains in discovering nanomaterials, biomolecules, and superconductors.

\subsection{Automated Simulation and Characterization}

A major bottleneck in materials science is the high level of domain expertise and manual effort required to set up, execute, and analyze computational simulations and physical characterization experiments. Complex software packages often have steep learning curves, and experiments require precise, adaptive control. Automating these workflows can democratize access to powerful scientific tools, reduce human error, and enable high-throughput screening and characterization.

Agentic systems are emerging to tackle these challenges by providing natural language interfaces to complex scientific instruments and software. In computational fluid dynamics (CFD), \textbf{Foam-Agent} \cite{yue2025foam} was developed to automate intricate OpenFOAM simulation workflows. It interprets natural language instructions using a multi-agent framework featuring a hierarchical retrieval system and a dependency-aware file generation process. Critically, its iterative error correction mechanism can diagnose and resolve simulation failures autonomously, achieving a high success rate (83.6\%) on benchmark tasks and significantly lowering the expertise barrier for CFD.

Similarly, \textbf{ChemGraph} \cite{pham2025chemgraph} is an agentic framework designed to streamline computational chemistry workflows. It uses LLMs for task planning and reasoning, allowing users to perform complex calculations (e.g., geometry optimization, thermochemistry) via natural language. The framework intelligently decomposes complex tasks for smaller LLMs and integrates various simulation tools, from machine learning potentials to density functional theory (DFT), making advanced atomistic simulations more accessible. In the field of solid mechanics, \textbf{MechAgents} \cite{ni2024mechagents} uses a team of collaborating LLM agents to solve complex elasticity problems. The agents autonomously write, execute, and self-correct code to perform finite element analysis, handling various geometries, boundary conditions, and material laws. The collaborative "criticism" among agents enhances the reliability of the solutions.

Beyond simulation, agents are also being applied to automate physical experiments. \textbf{AILA} (Artificially Intelligent Lab Assistant) \cite{mandal2024autonomous} is a framework that uses LLM-driven agents to automate atomic force microscopy (AFM). The work introduces AFMBench, a suite for evaluating agent performance across the scientific workflow, from experimental design to data analysis. The study found that multi-agent architectures outperform single agents and highlighted the need for rigorous benchmarking, as domain-specific knowledge (QA proficiency) did not directly translate to effective experimental control.The \textbf{AutoMat} \cite{yang2025automat} framework was developed as an agentic AI system that autonomously reconstructs atomic crystal structures and predicts material properties from high‑resolution microscopy images. Deployed as an end‑to‑end pipeline integrating denoising, physics‑guided template retrieval, symmetry‑constrained reconstruction, and ML‑based property prediction, it bridges experimental STEM imaging with atomistic simulation—achieving accurate, closed‑loop reasoning from microscopy to materials modeling.

\begin{table*}[!t]
\centering
\small
\caption{Examples of Validated Scientific Discoveries Achieved by AI Agents in Materials Science.}
\label{tab:conclusion_materials}

\renewcommand{\arraystretch}{1.5} 
\setlength{\tabcolsep}{6pt} 

\begin{tabularx}{\textwidth}{
    >{\raggedright\arraybackslash\bfseries\scriptsize}p{3cm} 
    >{\raggedright\arraybackslash\scriptsize}p{4.5cm} 
    >{\raggedright\arraybackslash\scriptsize}X 
}
\toprule
\textbf{Agent System} & \textbf{Application Domain} & \textbf{Novel Scientific Contribution or Validated Discovery} \\
\midrule

\rowcolor[HTML]{E8F5E9}
\textbf{SciAgents} \cite{ghafarollahi2025sciagents} & Biologically Inspired Materials & Autonomously discovered a \textbf{new biocomposite material with enhanced mechanical properties} and improved sustainability by identifying hidden interdisciplinary relationships and design principles from nature. \\

\textbf{TopoMAS} \cite{zhang2025topomas} & Topological Materials & In collaboration with human experts, the system guided the identification and confirmation (via first-principles calculations) of \textbf{novel topological phases in the material Strontium Antimonate ($SrSbO_3$)}. \\

\rowcolor[HTML]{E8F5E9}
\textbf{AtomAgents} \cite{ghafarollahi2025automating} & Alloy Design & Autonomously designed and discovered \textbf{novel metallic alloys with enhanced properties} compared to their pure elemental counterparts, demonstrating the crucial role of solid solution alloying. \\

\bottomrule
\end{tabularx}
\end{table*}

%% file: Sections/8.Physics.tex
\section{Agentic Physics and Astronomy Research}
\label{sec:physics}

The application of agentic AI is rapidly transforming research in physics and astronomy, fields characterized by vast datasets, complex simulations, and intricate experimental procedures. From automating telescope operations to solving complex problems in mechanics and quantum computing, agent-based systems are accelerating the scientific discovery pipeline. These agents assist researchers by managing complex software, analyzing data, formulating and testing hypotheses, and even automating the entire research cycle from literature review to final publication. This section reviews recent advancements, categorized by sub-discipline, showcasing how agentic AI is tackling key challenges in these domains (Table~\ref{tab:Physics} and Table~\ref{tab:conclusion_physics}).

\begin{table*}[!t]
    \centering
    \small
    \caption{Classification of Agentic Systems in Physics and Astronomy Science, organized to correspond with the survey text. Column Key: \textbf{Level}: Automation level of the agent, \textbf{Hypo.}: Observation or Hypothesis Generation, \textbf{Exper.}: Experimental Planning or Execution, \textbf{Analysis}: Data and Result Analysis, \textbf{Validation}: Synthesis, Validation, and Evolution. {\leveltwoicon} means level 2 and {\levelthreeicon} means level 3.}
    \label{tab:Physics}
        
    \renewcommand{\arraystretch}{1.2}
    \setlength{\tabcolsep}{2.5pt}
        
    \begin{tabularx}{\textwidth}{l X cccc c}
    \toprule
    & & \multicolumn{4}{c}{\textbf{Core Process}} & \\
    \cmidrule(lr){3-6}
    \textbf{Paper} & \textbf{Application Domain} &
    \textbf{Hypo.} & \textbf{Exper.} & \textbf{Analysis} & \textbf{Validation} & \textbf{Level} \\
    \midrule

    \multicolumn{7}{c}{\textit{Quantum Computing}} \\
    \rowcolor[HTML]{F5EEFA}
    k-agents \cite{cao2024agents} & Quantum Processor Control & & \checkmark & \checkmark & \checkmark & \leveltwoicon \\
    \midrule
    
    \multicolumn{7}{c}{\textit{General Frameworks and Methodologies}} \\
    \rowcolor[HTML]{F5EEFA}
    MoRA \cite{jaiswal2024improving} & Physics Problem Solving & & & \checkmark & & \leveltwoicon \\
    LP-COMDA \cite{liu2024physics} & Power Converter Design & & \checkmark & \checkmark & \checkmark & \leveltwoicon \\
    \rowcolor[HTML]{F5EEFA}
    LLMSat \cite{maranto2024llmsat} & Autonomous Spacecraft Control & & \checkmark & & \checkmark & \leveltwoicon \\
    CosmoAgent \cite{xue2024if} & Agent-based Civilization Modeling & \checkmark & \checkmark & \checkmark & & \leveltwoicon \\
    \midrule
    
    \multicolumn{7}{c}{\textit{Astronomy and Cosmology}} \\
    \rowcolor[HTML]{F5EEFA}
    StarWhisper \cite{wang2024starwhisper} & Supernova Survey Automation & & \checkmark & \checkmark & \checkmark & \leveltwoicon \\
    mephisto \cite{sun2024interpreting} & Galaxy Observation Interpretation & \checkmark & & \checkmark & \checkmark & \leveltwoicon \\
    \rowcolor[HTML]{F5EEFA}
    AI Agents \cite{kostunin2025ai} & Gamma-ray Astronomy Pipelines & & & \checkmark & & \leveltwoicon \\
    AI Cosmologist \cite{moss2025ai} & Cosmological ML Research & \checkmark & \checkmark & \checkmark & \checkmark & \levelthreeicon \\
    \rowcolor[HTML]{F5EEFA}
    SimAgents \cite{zhang2025bridging} & Cosmological Simulation Setup & & \checkmark & \checkmark & & \leveltwoicon \\
    \midrule
    
    \multicolumn{7}{c}{\textit{Computational Mechanics and Fluid Dynamics}} \\
    \rowcolor[HTML]{F5EEFA}
    OpenFOAMGPT \cite{pandey2025openfoamgpt} & CFD Simulation (OpenFOAM) & & \checkmark & \checkmark & \checkmark & \leveltwoicon \\
    OpenFOAMGPT 2.0 \cite{feng2025openfoamgpt} & CFD Simulation (OpenFOAM) & \checkmark & \checkmark & \checkmark & \checkmark & \levelthreeicon \\
    \rowcolor[HTML]{F5EEFA}
    LLM-Agent \cite{liu2025large} & Structural Beam Analysis (FEM) & & \checkmark & \checkmark & & \leveltwoicon \\
    MechAgents \cite{ni2024mechagents} & Solid Mechanics (FEM) & & \checkmark & \checkmark & \checkmark & \leveltwoicon \\
    \rowcolor[HTML]{F5EEFA}
    AutoGen-FEM \cite{tian2025optimizing} & Finite Element Analysis Automation & & \checkmark & \checkmark & \checkmark & \leveltwoicon \\

    \bottomrule
    \end{tabularx}
\end{table*}

\subsection{General Frameworks and Methodologies}
Agentic AI is also being applied to a diverse range of other problems in physics and engineering, from fundamental reasoning to applied system design and theoretical exploration.

A significant challenge for LLMs is scientific reasoning, particularly in physics, where they often exhibit problem miscomprehension, incorrect concept application, and computational errors. To enhance this capability, the \textbf{Mixture of Refinement Agents (MoRA)} framework was developed \cite{jaiswal2024improving}. MoRA uses an ensemble of specialized agents to iteratively refine a base solution generated by an LLM, with each agent targeting a different type of error. This approach significantly improved the accuracy of open-source LLMs on physics reasoning benchmarks like SciEval and PhysicsQA by up to 16\%. In the field of power electronics, \textbf{LP-COMDA} is a physics-informed autonomous agent designed to automate the modulation design of power converters \cite{liu2024physics}. An LLM-based planner coordinates with physics-informed tools to iteratively generate and refine designs, providing an explainable workflow that reduced error by 63.2\% compared to the next-best method and was over 33 times faster than conventional human-led design processes.

In astronautics, the \textbf{LLMSat} agent was developed to serve as a high-level, goal-oriented controller for autonomous spacecraft, aiming to reduce reliance on human mission control for deep space exploration \cite{maranto2024llmsat}. Tested in the Kerbal Space Program simulator, the work found that while current LLMs have limitations in handling high-complexity missions, their performance can be improved with advanced prompting frameworks and by carefully defining the agent's level of authority over the spacecraft. In fundamental physics, the \textbf{ArgoLOOM} framework was developed to act as an agentic AI orchestrator that autonomously coordinates computational tools across cosmology, collider, and nuclear physics domains \cite{bakshi2025argoloom}. Tested on case studies involving sterile‑neutrino scenarios, the system demonstrated its ability to link large‑scale cosmological simulations with collider and deep‑inelastic‑scattering analyses through an LLM‑driven planning pipeline. In experimental physics, the \textbf{Accelerator Assistant} framework was developed as an agentic AI system capable of autonomously executing multi‑stage experiments at a large‑scale synchrotron user facility \cite{hellert2025agentic}. Deployed at the Advanced Light Source, it translates natural‑language prompts into structured execution plans that integrate data retrieval, control‑system interaction, and analysis workflows under strict safety and reproducibility constraints, demonstrating a two‑order‑of‑magnitude reduction in experiment preparation time for machine‑physics studies.
Finally, in a more theoretical application, the \textbf{CosmoAgent} system uses LLM-based agents to simulate interactions between human and hypothetical extraterrestrial civilizations \cite{xue2024if}. By programming agents with different worldviews and ethical paradigms, the system explores potential inter-civilizational dynamics, providing a novel tool for studying cooperation and conflict under conditions of asymmetric information.

\subsection{Astronomy and Cosmology}
Modern astronomy and cosmology face significant challenges driven by the data flood from next-generation telescopes like the James Webb Space Telescope (JWST) and the upcoming Cherenkov Telescope Array (CTA). Key issues include managing complex, large-scale observation schedules, processing and analyzing petabytes of data, and navigating sophisticated simulation software to test theoretical models against observations.

To address the high operational workload in astronomical surveys, the \textbf{StarWhisper Telescope System} was developed as an agent-based observation assistant for the Nearby Galaxy Supernovae Survey (NGSS) \cite{wang2024starwhisper}. This system automates the entire observational workflow, from generating customized observation lists to executing telescope operations via natural language commands. Its agents analyze images in real-time to detect transients and automatically generate follow-up proposals, significantly reducing the manual effort for astronomers. In the domain of data interpretation, \textbf{mephisto} is a multi-agent framework designed to emulate human reasoning when interpreting multi-band galaxy observations \cite{sun2024interpreting}. Mephisto interacts with the CIGALE spectral energy distribution (SED) fitting codebase, employing self-play and tree search to explore hypotheses and build a dynamic knowledge base. This method achieved near-human proficiency in analyzing JWST data, even identifying novel "Little Red Dot" galaxy populations.

For ground-based gamma-ray astronomy, the complexity of instruments like the CTA presents challenges in system control and data analysis. To mitigate this, AI agents have been proposed that are instruction-finetuned on specific documentation and codebases, such as the Gammapy framework \cite{kostunin2025ai}. These agents assist users by understanding the environmental context and automating complex tasks, including the maintenance of data models for the Array Control and Data Acquisition (ACADA) system and the generation of code for analysis pipelines.

Several agentic systems aim to automate the entire research workflow. The \textbf{AI Cosmologist} is an agentic system with specialized agents for planning, coding, execution, analysis, and synthesis, capable of autonomously conducting machine learning research from idea generation to paper writing \cite{moss2025ai}. It mimics the human research process by generating diverse implementation strategies and iterating based on experimental outcomes. Similarly, a multi-agent system built on the autogen/ag2 framework was developed to automate cosmological parameter analysis \cite{laverick2024multi}. This system uses Retrieval Augmented Generation (RAG) and local code execution, demonstrating its potential on data from the Atacama Cosmology Telescope. Another multi-agent system, \textbf{SimAgents}, addresses the bottleneck of translating parameters from academic literature into executable scripts for cosmological simulations \cite{zhang2025bridging}. Its specialized agents for physics reasoning and software validation demonstrated strong performance on a dataset of over 40 simulations, accurately extracting and configuring simulation parameters from published papers.

\subsection{Computational Mechanics and Fluid Dynamics}
Computational mechanics and fluid dynamics rely heavily on sophisticated and often user-unfriendly software for the Finite Element Method (FEM) and Computational Fluid Dynamics (CFD). A major challenge is the steep learning curve required to set up, run, and debug complex simulations, which limits accessibility and slows down research and engineering innovation.

To lower this barrier, \textbf{OpenFOAMGPT} was introduced as an LLM-based agent to streamline CFD simulations using the OpenFOAM solver \cite{pandey2025openfoamgpt}. The agent, augmented with a Retrieval-Augmented Generation (RAG) pipeline to embed domain-specific knowledge, successfully handles complex tasks like zero-shot case setup, boundary condition modification, and code translation across various engineering scenarios. Its successor, \textbf{OpenFOAMGPT 2.0}, expands this into a multi-agent framework for fully automated, end-to-end simulations from natural language queries \cite{feng2025openfoamgpt}. Featuring specialized agents for pre-processing, prompt generation, simulation, and post-processing, it achieved 100\% success and reproducibility across over 450 test cases, demonstrating the reliability of orchestrated agent systems for scientific computing.

In solid mechanics, LLMs often lack the quantitative reliability needed for engineering applications. To address this, an LLM-empowered agent was created for structural beam analysis that reframes the problem as a code generation task \cite{liu2025large}. By using chain-of-thought and few-shot prompting to generate and execute OpeeSeesPy code, the agent achieved over 99.0\% accuracy on a benchmark dataset, showing robust performance across diverse conditions. Expanding on this, \textbf{MechAgents} leverages multi-agent collaboration to solve complex elasticity problems \cite{ni2024mechagents}. Agent teams with specialized roles (e.g., planner, coder, critic) autonomously write, execute, and self-correct FEM code, demonstrating that synergistic collaboration and mutual correction improve overall performance. Research has also focused on optimizing these collaborations; one study used the AutoGen framework to systematically test configurations of agents with roles like "Engineer," "Executor," and "Expert" for Finite Element Analysis \cite{tian2025optimizing}. It found that well-defined roles and interaction patterns significantly increase task success rates, providing a foundation for automating complex simulation methodologies.

\subsection{Quantum Computing}
A central goal in quantum computing is the development of self-driving laboratories capable of high-throughput experimentation for tasks like processor calibration and characterization. A key challenge is integrating unstructured and multimodal laboratory knowledge into autonomous AI systems to enable closed-loop, intelligent control over experiments.

To tackle this, the \textbf{k-agents framework} was introduced to support the automation of quantum experiments \cite{cao2024agents}. This framework employs LLM-based agents to encapsulate complex laboratory knowledge, including available experimental operations and data analysis methods. Execution agents then break down experimental procedures into agent-based state machines, interacting with other agents to perform each step. The analyzed results from one step are used to drive state transitions, creating a closed-loop feedback system. When applied to a superconducting quantum processor, the agent system autonomously planned and executed experiments for hours, successfully producing and characterizing entangled quantum states with a proficiency matching that of human scientists.

\begin{table*}[!t]
\centering
\small
\caption{Examples of Validated Scientific Discoveries Achieved by AI Agents in Physical Sciences.}
\label{tab:conclusion_physics}

\renewcommand{\arraystretch}{1.2} 
\setlength{\tabcolsep}{6pt} 

\begin{tabularx}{\textwidth}{
    >{\raggedright\arraybackslash\bfseries\scriptsize}p{2.8cm} 
    >{\raggedright\arraybackslash\scriptsize}p{4.2cm} 
    >{\raggedright\arraybackslash\scriptsize}X 
}
\toprule
\textbf{Agent System} & \textbf{Application Domain} & \textbf{Novel Scientific Contribution or Validated Discovery} \\
\midrule

\rowcolor[HTML]{F5EEFA}
mephisto \citep{sun2024interpreting} & Galaxy Observation Interpretation & Interprets new multi-band observations from the James Webb Space Telescope to reason about the physical scenarios of a recently discovered population of "Little Red Dot" galaxies, achieving near-human proficiency and contributing directly to the understanding of these objects. \\
\midrule
The AI Cosmologist \citep{moss2025ai} & Cosmological ML Research & Automates the entire research workflow, from idea generation and experimental design to data analysis and the autonomous production of complete scientific publications. The system develops novel approaches by iterating on experimental outcomes, thereby generating new scientific insights directly from datasets. \\

\bottomrule
\end{tabularx}
\end{table*}

%% file: Sections/9.Benchmarks.tex

%% file: Sections/10.Challenge.tex
\section{Challenges in Agentic Science}
\label{sec:foundational_challenges}

\begin{figure}[!t]
    \centering
    \includegraphics[width=1.0\linewidth]{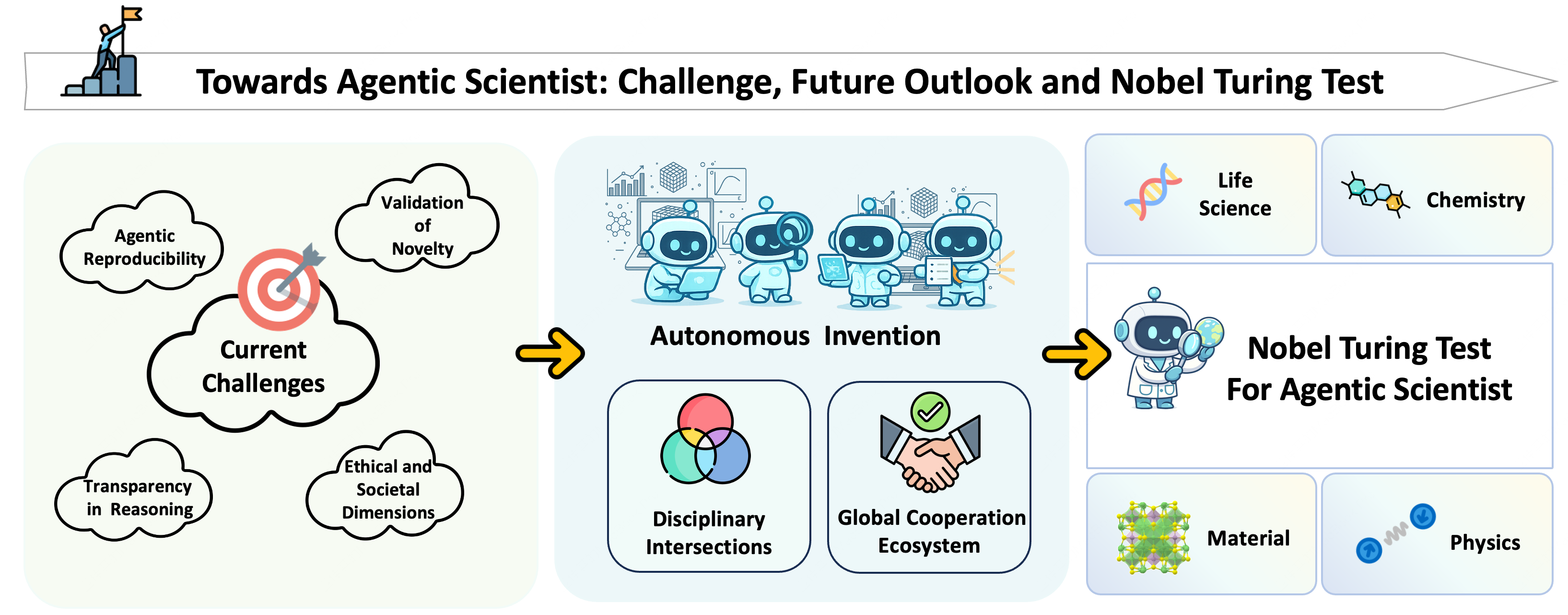}
    \vspace{-0.em}
    \caption{Exploring the Path to Agentic Scientists: Addressing Current Challenges, Enabling Autonomous Invention, and Pioneering the Nobel Turing Test Across Life Sciences, Chemistry, Materials, and Physics.}
    \label{fig:future}
\end{figure}

As AI systems evolve from narrowly scoped tools to autonomous scientific agents, they bring forth a new class of foundational and ethical challenges. These concerns reach beyond the well-documented limitations of large language models-such as hallucination, knowledge-updating inefficiencies, and catastrophic forgetting~\cite{kirkpatrick2017overcoming,huang2025survey, zhang2025siren}-and strike at the philosophical core of scientific inquiry: how knowledge is generated, validated, and trusted. Navigating these challenges is critical for the safe and credible integration of agentic AI into the natural sciences (Figure~\ref{fig:future}).

\subsection{Agentic Reproducibility and Reliability}

Scientific progress is predicated on reproducibility, yet agentic systems strain this foundational principle. Unlike conventional experiments that can be reproduced by rerunning code, agentic discovery involves replicating a stochastic and context-sensitive \textit{discovery trajectory}. Such trajectories are shaped by emergent reasoning patterns and contingent decisions, which are difficult to reproduce consistently~\citep{lu2024ai}. This challenge is compounded by several factors:
\begin{itemize}
    \item \textbf{Planning and Execution Failures:} The planning capabilities of base LLMs are a fundamental weakness; in autonomous modes, they often fail to generate executable plans, producing irrational or illogical steps that deviate from the intended task~\citep{huang2024understanding}. Furthermore, their ability to translate conceptual plans into correct, executable code is severely limited, with state-of-the-art benchmarks showing execution accuracy as low as 39\%~\citep{xiang2025scireplicate}.
    \item \textbf{System Instability:} The continual adaptation of these agents to evolving environments~\citep{majumder2023clincontinuallylearninglanguage, kim2024onlinecontinuallearninginteractive} is threatened by \textbf{catastrophic forgetting}-the tendency to lose previously acquired knowledge when trained on new data~\citep{kirkpatrick2017overcoming}. This instability undermines an agent's ability to maintain a coherent knowledge base over time, making it difficult to reproduce earlier findings after model updates.
    \item \textbf{Prompt Sensitivity:} In multi-stage experiments, agents exhibit a critical sensitivity to prompt wording. Minor variations, even when conveying the same intent, can lead to inconsistent guidance and divergent outcomes, making the discovery trajectory highly fragile and difficult to replicate reliably.
\end{itemize}
Achieving meaningful reproducibility will require formalizing new standards for logging agent states, decision policies, reasoning justifications, and environmental contingencies. Absent such mechanisms, we risk a future where scientific claims become irreproducible anomalies rather than verifiable contributions.

\subsection{Validation of Novelty}

A central promise of autonomous scientific agents is their capacity to generate genuinely novel hypotheses---insights that transcend the agent's training distribution~\citep{yehudai2025survey}. Yet this very capability introduces a fundamental validation dilemma: how can we differentiate between an authentic conceptual leap and an artifact of sophisticated interpolation or hallucination~\citep{ge2025llms, yehudai2025survey}? LLMs are fundamentally constrained by their training data, which leads them to generate ideas that often lack true originality and are repetitive across different runs~\citep{lu2024ai}. The tendency to produce plausible-sounding but false or unverifiable content, known as hallucination, can manifest as fabricated findings, data, or references, undermining the credibility of the output~\citep{huang2025survey, zhang2025siren}.

Verifying that a proposed hypothesis is not a derivative synthesis of existing patterns requires tools capable of auditing the agent's reasoning lineage. This problem is compounded by model opacity, as validation of novelty hinges on interpretability: the ability to trace and understand the inferential steps that led to a claim~\citep{xu2025advancingaiscientistunderstandingmaking}. Without such transparency, we are left with compelling-seeming conjectures that may lack genuine originality. Furthermore, the lack of reliable, objective, and scalable evaluation frameworks for AI-generated hypotheses remains a significant bottleneck, as current methods rely on resource-intensive and subjective human expert judgment.

\subsection{Transparency in Scientific Reasoning}

Scientific reasoning demands not only correct conclusions but also intelligible and auditable justifications. However, the architecture of many high-performing AI models inherently resists interpretation, undermining their trustworthiness as scientific collaborators~\citep{10.1007/978-3-031-67751-9_1}. Delegating scientific discovery to black-box oracles is unsound, as this opacity undermines scientific validation, trust, and the assimilation of AI-driven insights~\citep{xu2025advancingaiscientistunderstandingmaking}. Accordingly, there is an urgent need to move beyond post-hoc explainability toward the development of agents that are \textit{interpretable by design}---systems whose reasoning mechanisms are transparent, verifiable, and aligned with established scientific paradigms~\citep{bengio2025superintelligentagentsposecatastrophic}. Structured internal logs and clear documentation are vital for auditing the AI's reasoning and ensuring its conclusions are based on sound logic~\citep{bano2023investigating, banerjee2024ethical}. Interpretability is not a peripheral concern; it is essential for integrating machine-generated knowledge into the broader scientific corpus and ensuring that such knowledge can be critically evaluated and built upon by human researchers~\citep{banerjee2024ethical}.

\subsection{Ethical and Societal Dimensions}

The deployment of autonomous discovery agents introduces novel ethical and societal risks distinct from those associated with passive LLMs~\citep{yu2025survey}. Without the capacity for ethical judgment or self-regulation based on potential risks, these agents pose multifaceted challenges~\citep{bengio2025superintelligent}. These include:

\begin{itemize}
    \item \textbf{Accountability and Risk:} If an autonomous agent generates erroneous findings or uncovers hazardous compounds, who bears responsibility~\cite{bano2023investigating}? The possibility of dual-use outcomes-such as the autonomous discovery of toxins, pathogens, or other harmful technologies~\cite{huang2022overview}-raises acute concerns about misuse, particularly in the presence of adversarial attacks like backdoors or dataset poisoning~\cite{he2023controlriskpotentialmisuse, zhu2025demonagent, yang2024poisoning}. Effective governance must include mechanisms for attribution, traceability, and rapid response.

    \item \textbf{Impact on Scientific Labor and Education:} Agentic AI systems may significantly alter the structure of scientific labor and education. While they hold the promise of democratizing access to discovery, they also risk displacing human scientists from critical roles and reshaping the ecosystem of expertise and creativity~\cite{tadas2024redefining}. Over-reliance on AI for core research tasks could erode critical thinking and hands-on skills, diminishing scientific literacy from early training to expert practice~\cite{yamada2025ai}. This calls for rethinking human-agent collaboration models and preserving the creative agency of human researchers in the scientific process.

    \item \textbf{Governance and Integrity:} Ensuring ethical behavior from autonomous agents necessitates embedding normative constraints and values directly into their architectures~\cite{li2024agentalignmentevolvingsocial}. The large-scale generation of AI-driven research threatens to overwhelm peer-review systems and lower publication standards~\cite{lu2024ai}. Furthermore, biases in AI can skew research priorities toward topics with abundant data, exacerbating funding inequalities~\cite{yamada2025ai}. This involves setting principled boundaries on agent autonomy, maintaining continuous audit trails, and instituting robust oversight frameworks that include ethical red-teaming, pre-deployment verification, and post-deployment monitoring~\cite{pournaras2023science, brundage2020toward}.
    
\end{itemize}

These foundational and ethical dimensions must be addressed not as afterthoughts but as integral design considerations in the development of agentic scientific systems.

%% file: Sections/11.Future.tex
\section{Future Outlook of Agentic Science}
\label{sec:future_outlook}

Despite significant conceptual, technical, and ethical hurdles, the trajectory of agentic AI suggests the emergence of a transformative paradigm in scientific discovery. Beyond incremental automation, these systems may catalyze a shift toward \emph{computational epistemology}---a mode of inquiry where artificial agents participate in the invention, justification, and dissemination of scientific knowledge. This section outlines the key directions required to bridge current gaps, envisions the distinct evolutionary pathways for AI scientists, and presents four prospective frontiers that could define the next era of agentic science.

\subsection{From Automation to Autonomous Invention}

While current AI agents are predominantly constrained to automating existing workflows, a profound leap will occur when agents begin to engage in \textbf{autonomous invention}. Such systems would possess the capacity to interrogate the conceptual limitations of current methodologies and propose novel scientific instruments or conceptual frameworks. For instance, an agent might invent a new imaging modality to reveal a previously inaccessible subcellular process or formulate a novel mathematical abstraction to model emergent behaviors in complex systems. This marks a transition from tool-user to tool-creator, constituting a qualitatively distinct form of machine-driven scientific creativity.

\subsection{Interdisciplinary Synthesis at Scale}

Many of the most consequential scientific breakthroughs emerge at disciplinary intersections, yet human researchers are often limited by cognitive load and siloed expertise. Future agentic systems, trained on multimodal corpora spanning diverse scientific domains, could act as scalable engines for \textbf{interdisciplinary synthesis}. These agents could surface latent analogies between disparate fields--for example, mapping techniques from topological quantum field theory to deep learning architectures, or leveraging ecological dynamics to model economic systems. Such cross-domain reasoning transcends information retrieval, potentially enabling the discovery of unifying principles that reconfigure entire fields.

\subsection{The Global Cooperation Research Agent}

Looking further ahead, we envision a \textbf{global cooperation ecosystem of scientific agents}, distributed across institutions and research infrastructures. In this paradigm, specialized agents--e.g., a proteomics agent at one lab, a pharmacodynamics agent at another--interact within a decentralized, trust-aware network. These agents would not only share data but also engage in critical peer-review, hypothesis refinement, and collaborative experimentation \cite{de2024can}. Such a system could operate as a planetary-scale scientific engine, capable of tackling grand challenge problems whose complexity defies centralized human coordination. Realizing this vision will require advances in federated agent protocols, secure multi-agent reasoning (e.g., to mitigate recursive attack propagation \cite{zhou2025corba}), and mechanisms for conceptual accountability such as proof-of-thought cryptographic trails \cite{chen2024blockagents}.

\subsection{The Nobel-Turing Test}

A provocative benchmark for agentic science is what we term the \emph{Nobel-Turing Test}: can an autonomous agent, or a hybrid human-agent team, generate a discovery worthy of the Nobel Prize? Such a feat would demand more than competent execution of predefined tasks; it would require the agent to autonomously identify an unresolved and foundational scientific gap, generate a non-obvious and empirically testable hypothesis, and design a novel experimental methodology--potentially leveraging robotic systems and multi-agent collaboration \cite{yoshikawa2023large, song2024multiagent}. Crucially, it must also contextualize and interpret findings in a way that instigates a paradigmatic shift. Achieving this would mark the maturation of a fully autonomous scientific cycle, where agents are not merely instruments of execution, but originators of scientific insight \cite{pmlr-v202-carta23a}.

%% file: Sections/12.Conclusion.tex
\section{Conclusion}

Agentic Science marks a transformative stage in the evolution of AI for Science, where AI systems transition from computational assistants to autonomous research partners capable of reasoning, experimentation, and iterative discovery. Through our unified framework connecting foundational capabilities, core processes, and domain realizations, we provide a domain-oriented synthesis of autonomous scientific discovery across life sciences, chemistry, materials science, and physics. By situating agentic AI within this structured paradigm, we highlight both its broad applicability and the technical, ethical, and philosophical challenges that must be addressed to ensure trustworthy and impactful progress. We envision Agentic Science not as a replacement for human inquiry, but as a co-evolving paradigm that augments scientific creativity, accelerates discovery, and reshapes the future of research.